\documentclass[journal]{IEEEtran} 

\IEEEoverridecommandlockouts                              

\usepackage{adjustbox}    
\usepackage{dblfloatfix} 

\usepackage{subfig}
\usepackage{tikz}
\usepackage{cite}

\usepackage{amsmath}

\let\labelindent\relax
\usepackage{adjustbox}
\usepackage{hyperref}
\usepackage[symbol]{footmisc}

\def\BibTeX{{\rm B\kern-.05em{\sc i\kern-.025em b}\kern-.08em
    T\kern-.1667em\lower.7ex\hbox{E}\kern-.125emX}}
\graphicspath{{figures/evaluation}}
\newlength\Fcolumnseprule
\newcommand{\kstrongest}{$k$-strongest}
\newcommand{\papertitle}{CFEAR Radar odometry}

\newcommand{\eqname}[1]{\tag*{#1}}

\newcommand{\changed}[1]{#1} 
\newcommand{\changedd}[1]{{#1}}

\usepackage{array}
\usepackage{subfig}
\usepackage{makecell}

\usepackage[printonlyused]{acronym}
\usepackage{enumitem}
\usepackage{wrapfig}
\usepackage{amsmath,amsfonts}
\usepackage{bbm} 
\usepackage{textcomp}
\usepackage{url}
\usepackage{verbatim}
\usepackage{graphicx}
\def\BibTeX{{\rm B\kern-.05em{\sc i\kern-.025em b}\kern-.08em
    T\kern-.1667em\lower.7ex\hbox{E}\kern-.125emX}}
    \usepackage{cleveref}
\usepackage{balance}

\acrodef{CFAR}		[CFAR]			{Constant False-Alarm Rate}
\acrodef{FMCW}		[FMCW]			{Frequency-Modulated Continuous Wave}
\newcommand{\zmin}{z_{min}}

\begin{document}
\title{\LARGE \bf Lidar-level localization with radar? The CFEAR approach to accurate, fast and robust large-scale radar odometry in diverse environments}



\author{Daniel Adolfsson$^{1}$, Martin Magnusson$^{1}$, Anas Alhashimi,$^{2,1}$ Achim J. Lilienthal$^{3,1}$, Henrik Andreasson$^{1}$
  \thanks{$^{1}$The authors are with the MRO (Mobile Robotics and Olfaction) lab of the AASS research center at \"Orebro University, Sweden
  \{firstname.lastname\}@oru.se}
  \thanks{$^{2}$Anas Alhashimi is also with the Computer Engineering Department, University of Baghdad, Baghdad, Iraq.}
  \thanks{$^{3}$Achim J. Lilienthal's main affiliation is TU Munich, Germany (chair ``Perception for Intelligent Systems''), achim.j.lilienthal\@tum.de.}
  \thanks{
  This work has been supported by Sweden´s Innovation Agency projects ``Radarize'' and ``TAMMP''.}
  }

\markboth{
accepted for publication in Transactions on Robotics
}{}

\maketitle

\begin{abstract}
This paper presents an accurate, highly efficient, and learning-free method for large-scale odometry estimation using spinning radar, empirically found to generalize well across very diverse environments -- outdoors, from urban to woodland, and indoors in warehouses and mines -- without changing parameters. Our method integrates motion compensation within a sweep with one-to-many scan registration that minimizes distances between nearby oriented surface points and mitigates outliers with a robust loss function. Extending our previous approach CFEAR, we present an in-depth investigation on a wider range of data sets, quantifying the importance of filtering, resolution, registration cost and loss functions, keyframe history, and motion compensation. We present a new solving strategy and configuration that overcomes previous issues with sparsity and bias, and improves our state-of-the-art by 38\%, thus, surprisingly, outperforming radar SLAM and \changedd{approaching} lidar SLAM. \changedd{The most accurate configuration achieves 1.09\% error at 5~Hz on the Oxford benchmark, and the fastest achieves 1.79\% error at 160~Hz.}
\end{abstract}

\section{Introduction}
\IEEEPARstart{E}{stimating} 
 odometry in large-scale environments from exteroceptive sensors is a core challenge for autonomous navigation. Today, research on camera- and lidar-based localization has matured and there exist numerous accurate methods~\cite{8764393}. Unfortunately, camera and lidar are sensitive to \changed{dust\footnote[1]{
We provide a visual comparison that gives an indication on how dust affects camera, lidar, and radar at \url{http://tinyurl.com/RadarVsLidar}} and harsh weather such as rain, fog and snow~\cite{hong-2022-radarslam,1315077,8099786,8575761,8593703,7844000,9304681}}.

 In contrast, radar is largely resilient to these conditions~\cite{burnett2021radar,hong-2022-radarslam,replace_radar_lidar}, and receives increasingly higher attention for various perception tasks~\cite{9760104}. Hence, radar has great potential to enable robust localization in a wide range of scenarios; \changed{from autonomous cars and logistics where uninterrupted autonomy is required, to work machines
 surrounded by fugitive dust when operating on unpaved roads and within underground mines or construction sites.}
 
While spinning radar provides accurate and dense data, it has been considered hard to interpret due to its challenging noise characteristics~\cite{9760104}.
Consequently, previous radar odometry systems put great effort into overcoming sensor-specific challenges such as speckle noise, ghost objects and receiver saturation.
A range of methods utilize learning to remove noise~\cite{barnes_masking_2020,8794014} and predict \changed{keypoints}~\cite{barnes_under_2020,burnett2021radar}, while \changed{others} 
extract features from intensity peaks~\cite{hong2020radarslam,8793990,8460687,ADOLFSSON2022104136}, apply robust data association  ~\cite{hong-2022-radarslam,8460687,8793990,burnett_we_2021,barnes_under_2020},
or detect failures~\cite{8917111,ADOLFSSON2022104136}. 

Until today, the most efficient methods do not integrate local geometry information into the cost function, and instead, minimize a point-to-point metric that is sensitive to noise and sparsity.
\changed{Moreover, limited attention has been put into using multiple sweeps for online incremental odometry~\cite{burnett2021radar}, and no work systematically evaluates the potential benefit.
}



\begin{figure}[]
      {\includegraphics[trim={7.0cm 0cm 3.0cm 0cm},clip,width=0.99\hsize,angle=0]{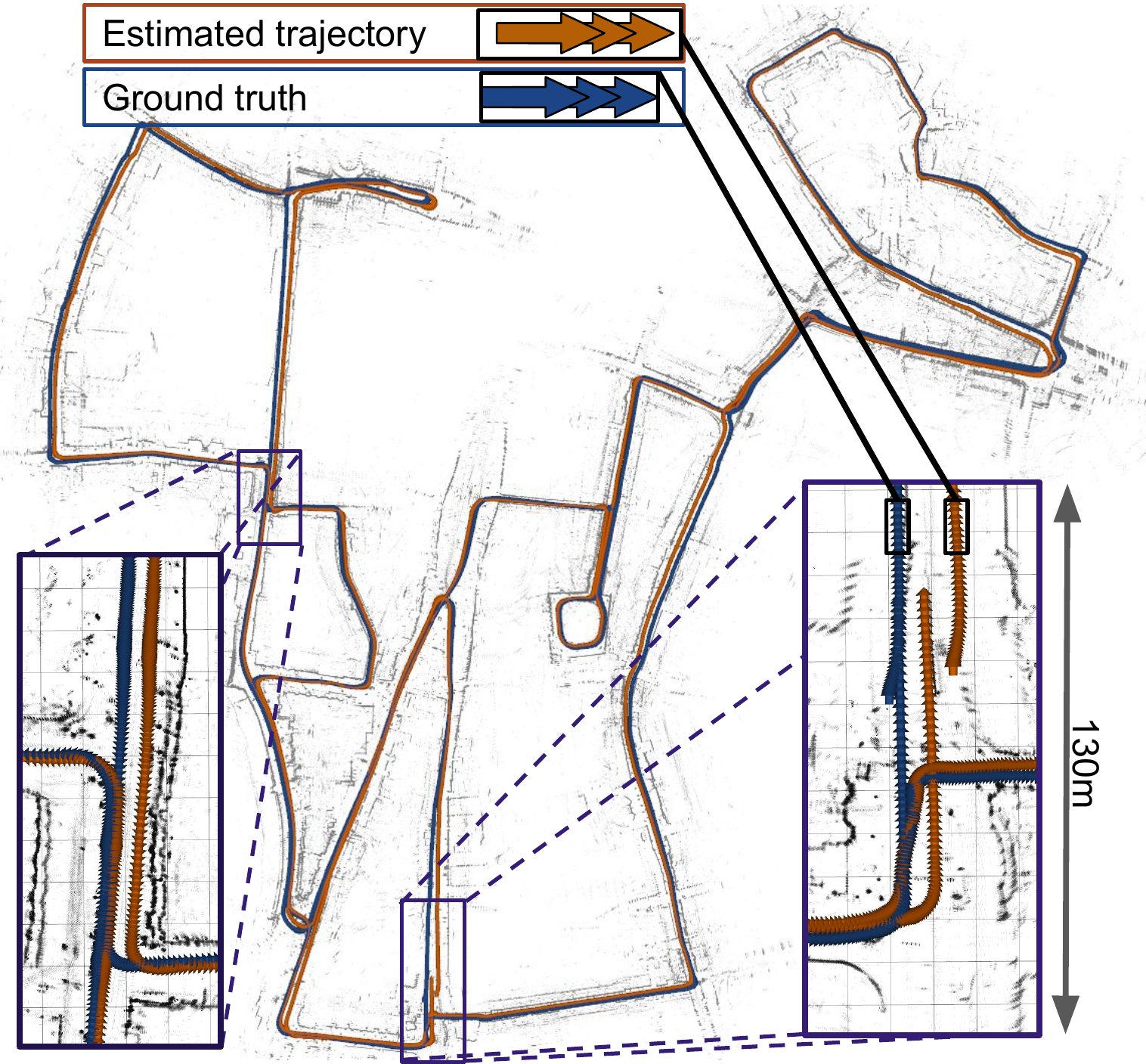}}\hfill\\
    \caption{
    Our efficient and accurate incremental radar odometry (orange) compared to ground truth (blue). Over a distance of $10$~km, we obtain a final error of less than $10$~m in the Oxford Radar RobotCar Dataset. Please note that our pipeline does not include any refinement or loop closure.
    Our code can be found at \textit{\url{https://github.com/dan11003/CFEAR_Radarodometry}}.
    }\label{fig:oxford_best_sequence}
  \vspace{-0.3cm}
\end{figure}



This paper builds upon our previous publications ``Conservative Filtering for Efficient and Accurate Radar odometry'' (CFEAR Radar odometry)~\cite{adolfsson2021cfear,adolfsson2021oriented}. In those articles, we proposed (i) a two-stage learning-free method for computing a sparse set of oriented surface points, by applying a conservative filter and then analyzing local geometries, and  (ii)  an improvement to incremental scan matching that registers the latest scan to multiple scans jointly and minimizes a point-to-line metric. 
In the present work, we provide several improvements and new versions of CFEAR that further increase its performance, one of our most accurately estimated trajectories is visualized in Fig.~\ref{fig:oxford_best_sequence}.
We also substantially extend the validation and discussion of the method and derive insights into its performance, from an evaluation on additional datasets and comparisons to further baseline methods. In particular, we present the following novel contributions:

\begin{enumerate}
    \item A thorough ablation study that quantifies the importance of each component in the CFEAR Radar odometry pipeline, including radar filtering, surface point resolution, registration cost and loss functions, keyframe scan history, intensity weighed surface point estimation, weighted residuals, and motion compensation. 
    \item Based on the ablation study we propose three efficient real-time configurations that run at 44--160~Hz, and one low drift configuration that runs at 5~Hz. We propose a new solving strategy that speeds up our method.
    \item We propose a combination of weighting, key frame history, filtering and a cost function that overcomes previous challenges with sparsity, bias, and overly conservative filtering, hence advancing our previous state-of-the-art by $38\%$, reaching an average of $1.09\%$ translation error on the Oxford Radar RobotCar dataset. This is more accurate than the state of the art in radar-based SLAM~\cite{hong-2022-radarslam} (that additionally includes loop closure) and challenges lidar-based SLAM~\cite{behley2018rss}. 
    \item We carried out and present validation that shows how our learning-free method generalizes across different sensor configurations, spatial scales, and environments, without changing parameters. We validate our method by spatial cross-validation on two separate environments, and by qualitative evaluation on indoor and outdoor datasets. 
    \item Finally, we make our C++ implementation\footnote{Code and dataset: \url{https://github.com/dan11003/CFEAR_Radarodometry}}, our datasets$^2$ and the full evaluation\footnote{Evaluation: \url{https://github.com/dan11003/CFEAR_evaluation}} available to the community.
\end{enumerate}


\section{Related work}
Localization and mapping using radar have been research topics in robotics for many years~\cite{938381,681411}. 
Previous work focuses on mapping~\cite{lu2020smoke} or ego-motion using a single~\cite{9196666,9193901,10.1145/3384419.3430776} or multiple~\cite{9495184} low-cost system-on-a-chip (SOC) radar with Doppler information. Other works operate on automotive Doppler radar for ego-motion estimation~\cite{RAPP2017136,gao2021accurate,kung2021normal,9463737}, SLAM~\cite{7795967,8813841} or localization in a given map~\cite{7535489}. \changedd{Finally, some methods~\cite{vivet2013localization} operate on mechanically spinning Doppler radar such as~\cite{impala}.}
\changedd{
In this work, we focus on estimating incremental 2d odometry from spinning \ac{FMCW} radars with dense returns, without using Doppler measurements.}

Note that our initial feature extraction step reduces dense data to a sparse point cloud. The later stages, which are agnostic to sensor type, could be modified to operate on other sources of sparse range data such as \changedd{ lidar, automotive radar or beamforming radar.}
An example of dense spinning radar data is depicted in Fig.~\ref{fig:polar_and_Cartesian_radar}.
\changed{Recent spinning radars without Doppler measurements} have demonstrated high range, accuracy, and richness and have inspired numerous methods; from odometry estimation and alignment quality assessment to global localization~\cite{barnes_under_2020,saftescu_kidnapped_2020,radar_on_lidar_place,gadd_fool_me_once_place,kradarplusplus,8957240,replace_radar_lidar}, localization in \changedd{previous maps~\cite{radar_on_lidar_place}} and SLAM~\cite{hong2020radarslam,hong-2022-radarslam}.



\subsection{Filtering and feature extraction of spinning radar data}
\label{sec:rw-filter}

Interpreting dense spinning radar data for the task of odometry estimation has been considered challenging and the community has largely explored learning-based approaches. Recent methods have used weak supervision from external ego-motion estimator and spatial sensor coherence~\cite{8794014} to filter data,
or from partial occupancy labels generated by lidar~\cite{weston2019probably} to predict occupancy.
Yin et al.~\cite{yin2020radaronlidar} used a generative adversarial network to transfer radar data into a representation that resembles lidar and can more easily be interpreted.

For odometry estimation, ground truth poses have been used together with semi-supervised learning to mask out  noise~\cite{barnes_masking_2020} and to extract \changed{keypoint} locations, scores and descriptors~\cite{barnes_under_2020}. 
\changed{
More recently, Burnett et al.~\cite{burnett2021radar} showed that unsupervised learning can be achieved by alternating between estimation and back-projection. Their method can consequently be tailored to any environment without an external ground truth system.
}
\changed{
In the evaluation by Barnes et al.~\cite{barnes_masking_2020}, spatial cross-validation (SCV), which spatially separates training and validation data to understand how the odometry generalizes to new environments, 
showed that their initial results suffered from overfitting with lower performance (55\% and 106\% higher translation error for Cart and Dual Cart respectively at a resolution 0.4\,m/pixel) 
in environments outside the training set. These results were achieved when trained from data with limited structural diversity, acquired from 25 repetitions of a 3~km long route.
We are interested in exploring learning-free methods, aiming to find accurate and generalizable models for filtering without the need for a large amount of training data.
}

Traditional learning-free methods for radar filtering, such as \ac{CFAR}, have been challenging to apply in robotics due to unknown context-dependent noise characteristics~\cite{vivet2013localization,8793990,8460687}. Cen et al.~\cite{8793990,8460687} proposed two methods tailored for robotics that extract \changed{keypoints} from peaks in polar space using image intensity and gradients. Similarly, Hong et al.~\cite{hong2020radarslam} extract peaks exceeding one standard deviation above the mean intensity per azimuth. Kung et al.~\cite{kung2021normal} and Mielle et al.~\cite{mielle-2019-comparative} keep all points exceeding a noise threshold. However, a fixed noise floor with no additional restrictions requires prior knowledge of noise level and does not mitigate multipath reflections. Additionally, as shown in our evaluation, a fixed noise floor without bounding the number of detections can have a negative impact on efficiency, and modeling geometrical features such as a surface normal or Gaussian distribution from too complex data can reduce map and tracking quality~\cite{adolfsson2021cfear,8870941}. For that reason, our filter outputs a limited amount of points for computing oriented surface points around the most dominant landmarks in the scene. 
Recently, a combination between \ac{CFAR} and fixed threshold \cite{alhashimi2021bfarbounded} noticeably improved the odometry estimation error, however prior information about the noise level is required.
Similar to most other learning-free methods~\cite{8793990,8460687,kung2021normal,mielle-2019-comparative}, 
our filter assumes that multipath reflections and speckle noise are observed with lower intensity compared to real landmarks.

Marck et al.~\cite{marck2013indoor} suggest keeping only the single strongest reflection per azimuth which we show in our evaluation is insufficient for accurate odometry estimation. \changedd{Our method operates on azimuths in polar form separately (rows in Fig.~\ref{fig:polar_and_Cartesian_radar}) and combines a static threshold with \textit{$k$ strongest} filtering.} 

Following our polar space filtering, we additionally analyze local geometries of the filtered point set in Cartesian space to compute stable oriented surface points as detailed in Section~\ref{sec:surface_point}. We take inspiration from previous lidar-based methods that have exploited local surface orientation for odometry estimation and mapping
\cite{zlot-2014-efficient,behley-2018-suma,todor_ndt} and demonstrate how this is beneficial also for FMCW radar.

\begin{figure*}[htp!]
\vspace{0cm}
    \centering
\includegraphics[width=0.90\hsize]{"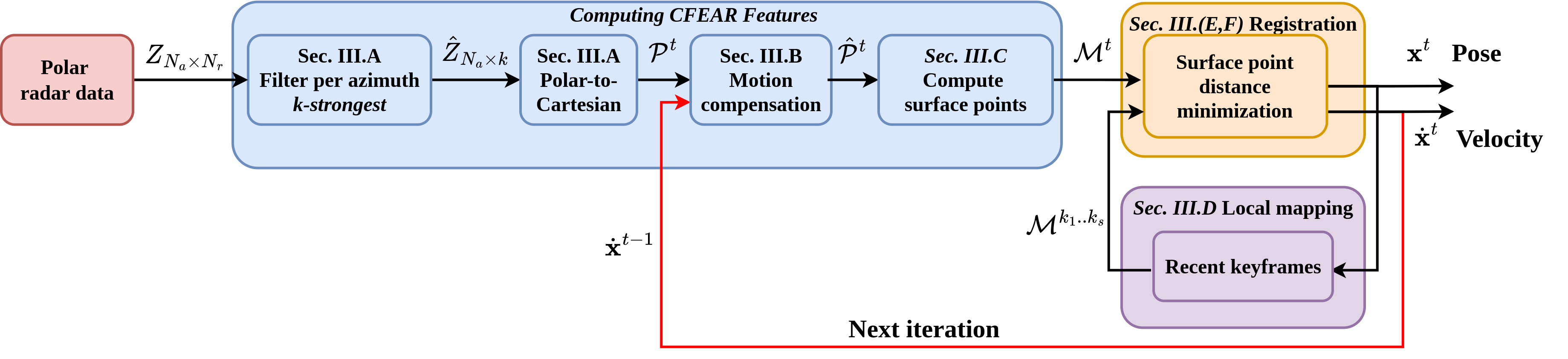"}\vspace{-0.2cm}
    \caption{Overview of \papertitle. Key components are motion compensation and one-to-many scan registration that mitigates outliers via a robust loss function. Red lines indicate that data is passed to the next iteration.}
    \label{fig:overview_method}
    \vspace{-0.4cm}
\end{figure*}

\subsection{Radar odometry}
\label{sec:rw-radar}
Previous work on radar odometry can be categorized into sparse methods, which extract and match keypoints~\cite{barnes_under_2020,hong2020radarslam,hong-2022-radarslam,8460687,8793990,kung2021normal,burnett_we_2021,burnett2021radar}, and dense methods~\cite{Checchin,barnes_masking_2020,9197231,fmbm}, which operate on full radar images and explicitly incorporate information about free space. Barnes et al~\cite{barnes_masking_2020} suggested a brute force approach that samples and selects rotations that maximize the dense correlation between radar scans.
Similar to Checchin et al.~\cite{Checchin}, Park et al.~\cite{9197231} addressed the necessity of sampling by instead maximizing the phase correlation between log-polar images using the Fourier Mellin Transform, to solve for orientation.
Despite this, dense matching suffers from high complexity and does not scale well with sensor range resolution, and needs to perform downsampling prior to estimating alignment. For that reason, we are interested in exploring sparse methods that compute and match sparse yet representative feature sets.

Sparse methods can be categorized further based on data association and matching cost function.
Barnes et al.~\cite{barnes_under_2020} used a dense correspondence search and assigned weights according to keypoint descriptor and score similarity. Cen et al.~\cite{8460687,8793990} and Hong et al.~\cite{hong-2022-radarslam} used consistency graphs to find the largest set of inliers, and accordingly, reject outliers. 
Burnett et al.~\cite{burnett_we_2021} used motion-compensated RANSAC for robust association and matching, and more recently, proposed \textit{Hero}~\cite{burnett2021radar} that learns key-point locations, uncertainties, and descriptors to mitigate outliers and perform matching, using the batch state estimation framework ESGVI~\cite{doi:10.1177/0278364920937608}. 
\changed{
Similar to ICP~\cite{4767965} we use Euclidean nearest neighbor search and resolve data association iteratively. However, our matching module has a few key differences: (i) Each surface point is assigned up to one correspondence per keyframe, as opposed to only handling pairwise correspondences. (ii) Residuals are reshaped via a robust loss to mitigate outliers. (iii) Residuals are weighted according to oriented surface point similarity.
}

When registering the set of correspondences, Kung et al.~\cite{kung2021normal} minimize the point-to-distribution (P2D) distance~\cite{Biber_ndt,martin_ndt}, while Barnes et al.~\cite{barnes_under_2020}, Cen et al.~\cite{8460687,8793990} and Hong et al.~\cite{hong2020radarslam} minimize a weighted squared point-to-point (P2P) distance. 
In our evaluation, we show that traditional scan matching with P2P cost does not perform well with sparse sets in the radar domain, as long as only two consecutive scans are used.
Similar to Kung et al.~\cite{kung2021normal} we integrate geometric uncertainty of landmarks e.g. via a point-to-(line and distribution) (P2L \& P2D) cost function and via residual weights. Going even further, we show that sparsity can be addressed by jointly registering the latest scan to multiple keyframes. Allowing multiple correspondences for a surface point makes matching denser with the additional benefit of gaining robustness from multiple redundant scans which constrain the challenging pose estimation problem. 
\changed{
We compare P2P, P2D and P2L registration cost functions in an extensive ablation study, demonstrating their properties and suitability for different tasks.
}

\changed{
Finally, some methods evaluate odometry quality once registration has been carried out~\cite{barnes_masking_2020,8917111,ADOLFSSON2022104136}. Barnes et al.~\cite{barnes_masking_2020} estimate pose covariance from weights obtained from correlative scan matching. Adolfsson et al.~\cite{ADOLFSSON2022104136} detect failures from an increase in the point cloud entropy owed to misalignment, while Aldera et al.~\cite{8917111} detect failures from patterns in landmark matches using the eigenvalues of the pairwise-compatibility matrix. In this work, we focus on achieving robustness during registration. However, we also present a method for computing the matching uncertainty directly from our objective function (see Sec.~\ref{sec:registration}), based on the work of Bengtsson and Baerveldt~\cite{BENGTSSON200329}, and Censi~\cite{4209579}.

\subsection{Aggregating scans for pose tracking}
Fusing scans into local maps have shown success for lidar-based odometry methods e.g. \changedd{in~\cite{Saarinen644380,loam,ct-icp}}.
Dellenbach et al. accumulate points from lidar scans directly into a voxel grid~\cite{ct-icp}.
Saarinen et al.~\cite{Saarinen644380} fuse points into grid cells using recursive covariance estimation, poses are estimated via scan-to-map registration~\cite{martin_ndt}.
Zhang et al.~\cite{loam} build local maps by accumulating edge and planar points separately and minimizing point-to-plane and point-to-line metrics. 
Each point in the latest scan is assigned a single line or plane correspondence, estimated from two and three nearby edge and planar points respectively.
Zaganidis et al.~\cite{Zaganidis2018} separate scans into disjoint sets based on per-point semantic label for mapping and scan-to-map registration, extending the registration cost function over each semantic class. In our work, we extend the registration cost function to jointly register the latest scan to multiple previous scans; each point is assigned a maximum of one correspondence per keyframe scan in a sliding window. \changedd{Hence, constraints against all keyframes is guaranteed given sufficient overlap and similarity. A detailed discussion follows in Sec.~\ref{sec:method_jointy_register}.}
When using automotive radar, aggregating detections into local maps has been necessary to overcome sparsity for incremental odometry~\cite{8813841}, matching of local maps~\cite{8813841} and localization in a previously corrected map~\cite{7535489}. Kung et al.~\cite{kung2021normal} showed that aggregating multiple scans into a local map is highly favorable for registration and reduces drift.
In spinning radar, local maps have been used for map refinement via bundle adjustment~\cite{hong2020radarslam} and factor graph optimization~\cite{9197231}.
Burnett et al.~\cite{burnett2021radar} presented a sliding window batch odometry estimation framework and found that a larger window up to at least 4 scans improved online odometry. Later, Burnett et al.~\cite{replace_radar_lidar} used local submaps of radar detections aggregated from 3 scans within a teach and repeat localization framework~\cite{replace_radar_lidar}. Despite this, no systematic evaluation has been carried out to investigate the principled benefit of considering additional scans in spinning radar research. In Sec.~\ref{sec:eval_submap_keyframes} we present an evaluation of how extending the window over previous keyframes affects various cost functions in terms of drift, pose accuracy and bias.
}

\section{CFEAR Radar odometry}
\label{sec:method}

\begin{figure}[h!]
    \centering
    \includegraphics[width=\hsize,trim={0cm 0cm 0cm 0cm},clip]{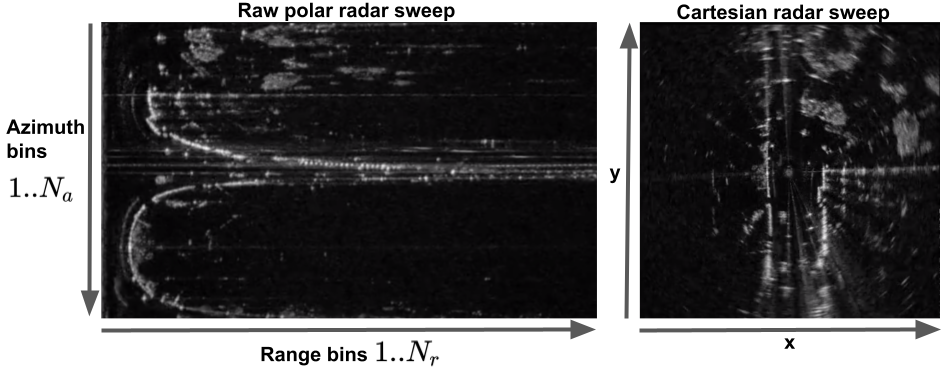}
    \caption{A raw sweep from spinning radar data on polar (left) and its corresponding Cartesian (right) form. The polar data ($Z_{N_a\times N_r}$) has the dimension of $N_a\times N_r$ readings. The conversion to Cartesian space is given by~\eqref{eq:cartesian}.}
    \label{fig:polar_and_Cartesian_radar}
    \vspace{-0.5cm}
\end{figure}

An overview of the CFEAR Radar odometry Pipeline can be seen in Fig.~\ref{fig:overview_method}. Our online incremental pipeline operates on full 360$^\circ$ radar sweeps in polar form as depicted in Fig.~\ref{fig:polar_and_Cartesian_radar}, produced by spinning radars such as the model CTS350-X by Navtech. The sensor data is first filtered to only keep the \textit{$k$ strongest} readings per azimuth to filter noise and produce a lightweight representation for further processing. The filtered point set is then compensated for motion distortion using the estimated velocity from the previous iteration, and accounting for the time difference of measurements. Sparse oriented surface points are then computed 
using a grid-based approach that computes the point distribution around each grid cell. Pose and velocity are finally computed by registering the current scan to a history of previous scans. 
The following sections describe these components in detail.

\subsection{{\kstrongest} filtering}
\label{sec:kstrong}
Given a $360^\circ$ polar intensity radar image $Z_{N_a \times N_r}$ with $N_a$ azimuth and $N_r$ range bins, the proposed filter {\kstrongest} iterates through all azimuth bins separately. For each azimuth bin $a\in\{1..N_a\}$, the $k$ range bins with the highest intensities (among all range bins $Z_{a,1..N_r}$) that exceed the expected noise level $z_{min}$ are selected. The parameter $z_{min}$ serves the purpose of mitigating low-intensity speckle noise and unreliable landmarks, while $k$ provides mitigation of receiver saturation and multi-path reflections.
As depicted in Fig.~\ref{fig:oriented_surface_points}, a small value of $k=12$ makes the filter conservative and efficient, mainly providing points around the primary landmarks where the intensity is highest. This is also the strategy that we used in our previous publication~\cite{adolfsson2021cfear}. However, a higher $k=40$ allows the filter to additionally detect secondary landmarks within the same azimuth bin to produce a more complete and complex representation of the scene.


Finding the $k$ highest values within a vector of length $N_r$ is a well-known selection problem that can be implemented 
\changed{
in e.g.  $O(N_r\times k)$ by iterating through all range bins, and insert elements into a sorted array of size $k$, or in $O(N_r+k)$ by inserting range bins into a hash table with intensities as keys and then selecting the elements (range bins) in the table with the highest key values. We found the first implementation to be faster for $k<40$, which may be due to a majority of returns with intensity lower than $z_{min}$ can be omitted. As a consequence, the average time complexity is lower compared to the worst case. However, the second implementation had a more consistent run-time performance.
}

After filtering, each selected range and azimuth tuple $(d,a), d\in[1..N_r], a\in[1..N_a]$ is transformed to a point in Cartesian space using 
\begin{equation}
\label{eq:cartesian}
    \mathbf{p} =
    \begin{bmatrix} p_x\\p_y
    \end{bmatrix}=
    \begin{bmatrix} d \gamma \text{cos}\left(\theta\right) \\ 
    d \gamma \text{sin}\left(\theta\right)
\end{bmatrix},
\end{equation}
where $\theta=2\pi a/N_a$ and $\gamma$ is the sensor range resolution. The output point cloud is the set of all Cartesian points $\mathcal{P}^t=\{ \mathbf{p}_i \}$

\subsection{Motion compensation}
Given the most recent translation and rotational velocity estimate $\dot{\mathbf{x}}^{t-1}=[v_x \; v_y \; \dot{\theta}]$, the point cloud $\mathcal{P}^t$ is compensated for motion by projecting each point into the time $t$ of the center of the sweep, similar to~\cite{loam} and \cite{hong2021radar}. Each point within a sweep is measured at the time $t+\delta^t$ where $\delta^t \in[-\Delta T/2,\Delta T/2]$ and $\Delta T$ is the duration of a full $360^\circ$ sweep, set to $\Delta T=0.25$~s for the radar used in our experiments. The time offset $\delta^t$ of measurement can then be computed from the index ($a_{\delta^t}$) of the corresponding azimuth bin and the number of azimuth bins per sweep ($N_a$):
\begin{equation}
\delta^t = (a_{\delta^t} -N_a/2)\Delta T/2.
\end{equation}

Assuming zero acceleration since the 
previous sweep
, the translation and rotation distortion terms ($\mathbf{t}_e$ and $\theta_e$) of a point $\mathbf{p}_{\delta^t}$ with time offset $\delta^t$ can be computed from the velocity as 
\begin{equation}
\begin{split}
&\mathbf{x}_e=[\mathbf{t}_e\;\theta_e]^{T}=\delta^t \dot{\bf{x}}^{t-1}=\delta^t[\mathbf{v} \; \dot{\theta}]^{T}=\delta^t[ v_x\;\; v_y \;\; \dot{\theta} ]^{T}.
\end{split}
\end{equation}
The point $\mathbf {p}_{\delta^t}$ can be corrected by applying a rotation and translation according to the inverse of the distortion:
\begin{equation}
\begin{split}
    \mathbf{\hat{p}}_{t} &= \mathbf{R}_{-\theta_e}\mathbf{p}_{\delta^t}-\mathbf{t}_e\\
    &=  \begin{bmatrix}
cos(-\theta_e) & -sin(-\theta_e) \\
sin(-\theta_e) & cos(-\theta_e)
\end{bmatrix}\begin{bmatrix}
p_x\\
p_y
\end{bmatrix}
-\delta^t\begin{bmatrix}
v_x\\
v_y
\end{bmatrix}.
\label{eq:correction}
\end{split}
\end{equation}



\subsection{Compute oriented surface points}
\label{sec:surface_point}
 \begin{figure}
  \begin{center}
    \subfloat[]{\includegraphics[trim={0.0cm 0cm 0cm 0cm},clip,width=0.32\hsize]{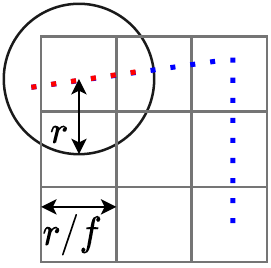}\label{fig:surface_point_grid}}\hfill
     \subfloat[]{\includegraphics[trim={1.3cm 0cm 3cm 0cm},clip,width=0.32\hsize]{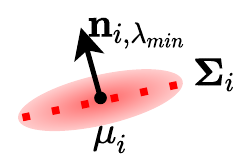}\label{fig:surface_point_cell}}\hfill
     \subfloat[]{\includegraphics[trim={0.0cm 0cm 0cm 0cm},clip,width=0.32\hsize]{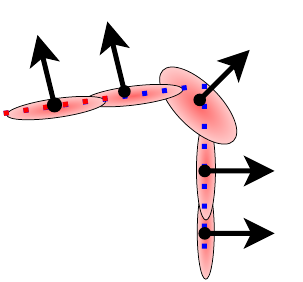}\label{fig:orient_surface_final}}\hfill
    \caption{The procedure of computing oriented surface points. (a) Points are inserted into a grid with cell size $r/f$. For each grid cell with points, the sample mean and covariance are computed from all nearby points within a radius $r$ of the cell centroid. (b) The sample covariance is used to estimate the direction of the surface normal. The final representation is seen in (c).\label{fig:compute_surface_point}}
  \end{center}
  \vspace{-0.5cm}
\end{figure}
\begin{figure}
\begin{centering}
    \subfloat[Point cloud and oriented surface points for k=12.]
      {\includegraphics[trim={0cm 1.5cm 0cm 1.8cm},clip,width=0.99\hsize,angle=0]{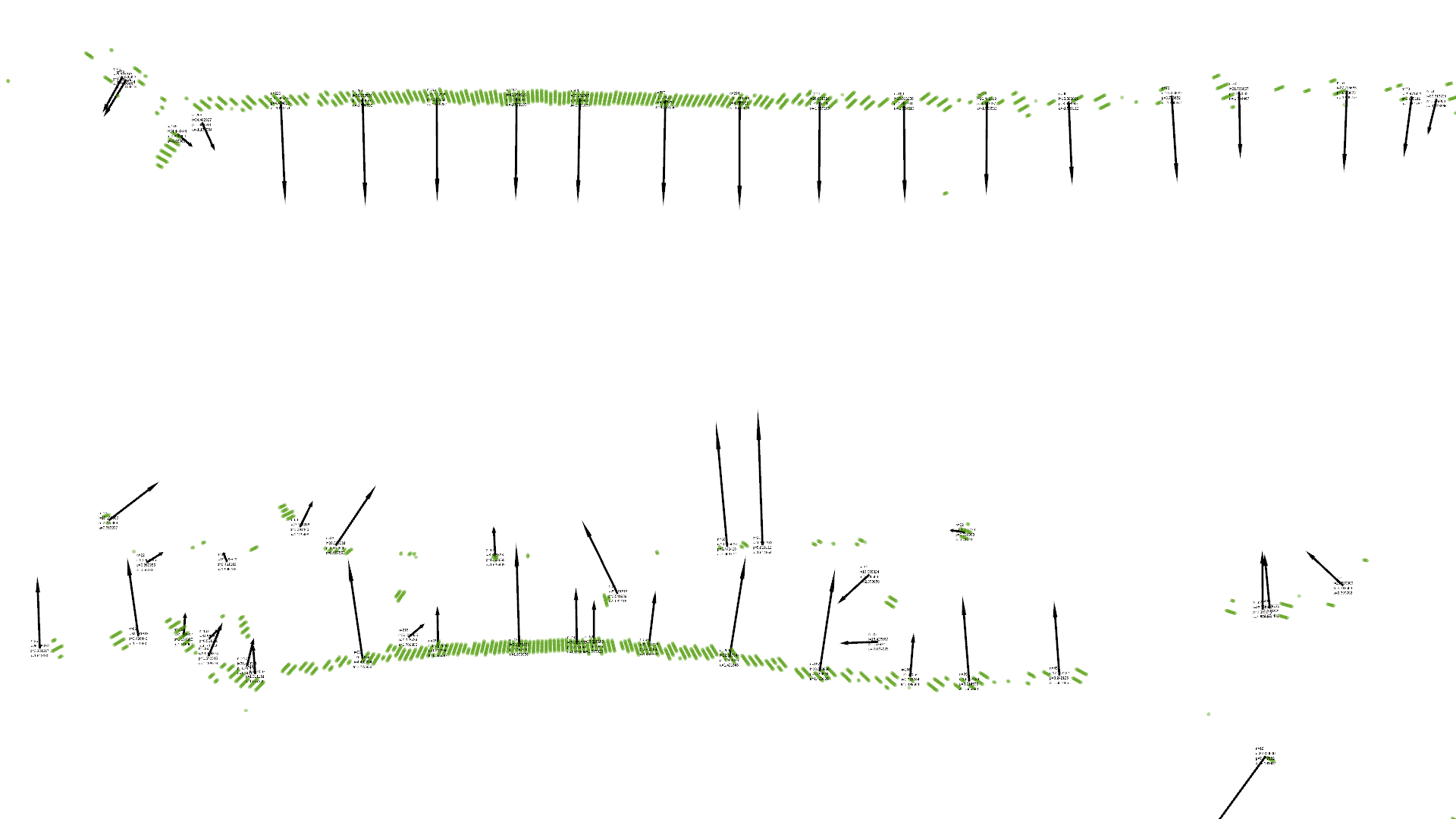}}\label{fig:oriented_k12}\\
    \subfloat[Point cloud and oriented surface points for k=40. Additional landmarks are detected (dashed blue) and more surface points are computed. ]
     {\includegraphics[trim={0cm 1.5cm 0cm 1.3cm},clip,width=0.99\hsize,angle=0]{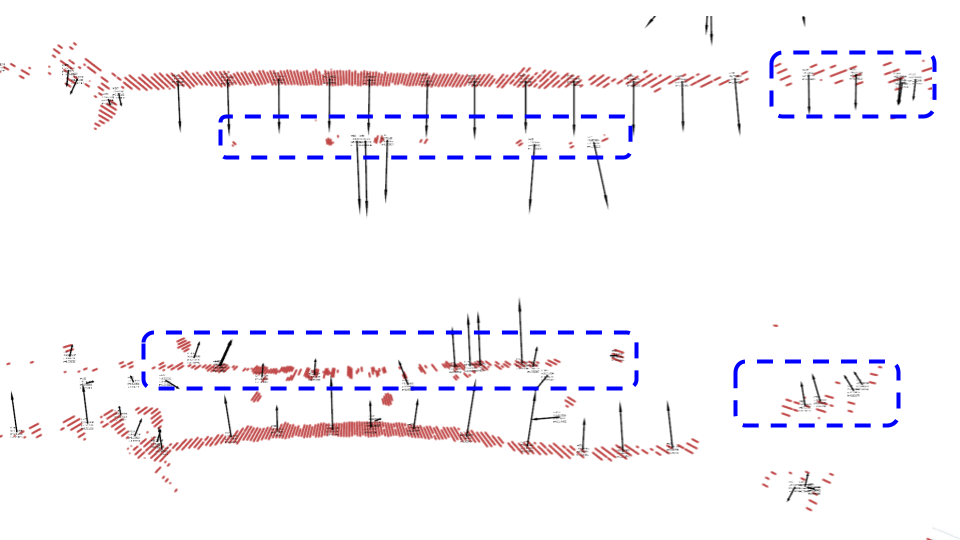}\label{fig:oriented_k50}}
    \caption{
    Point cloud $\mathcal{P}^t$ after \textit{\kstrongest} filter for $k=12$ (green) and $k=40$ (red), and computed oriented surface points (black arrows) scaled by planarity $p$ as in ~\eqref{eq:planarity}.
    A low value of $k$ makes the filter faster and more conservative, producing fewer surface points. A larger $k$ increases the number of detections around landmarks in the scene, resulting in additional surface points.
    }\label{fig:oriented_surface_points}
  \end{centering}
\end{figure}

In this section, we discuss how to create a sparse yet representative point set for efficient and accurate registration. Given the filtered and corrected point set $\mathbf{\hat{\mathcal{P}}}^t$, we aim to create a sparse representation that models geometries in the scene as a set of oriented surface points $\mathcal{M}^t=\{m_i\}$; i.e. pairs of a surface point and a normal $m_i=\{\mathbf{\mu}_i,\mathbf{n}_i\}$. 
\changed{
We aim to compute surface points that are stable in at least one direction (from planar surfaces such as buildings/walls or tree lines). The multitude of surface points with different directions around the sensor is typically sufficient to constrain odometry estimation (as evidenced by our results in Sec.~\ref{sec:evaluation}).
}
First, as depicted in Fig.~\ref{fig:surface_point_grid}, the corrected point cloud $\mathbf{\hat{\mathcal{P}}}^t$ is inserted into a grid with cell size $(r/f)\times (r/f)$, where $f$ controls the target density (size) of the final of the representation $|\mathcal{M}|$. For $f=2$, surface points are computed from disjoint point sets ($r$ fully enclosed in one grid cell).  We choose a lower value $f=1$ such that points can contribute to the computation of multiple surface points.
The radius $r$ corresponds to the expected size of landmarks to model within a surface point.
\changed{
For each cell in the grid that contains points, all $l$ points within a radius $r$ from the cell centroid (represented as a matrix $\mathbf{P}_{2\times l}= [\mathbf{\hat{p}}_k]_{k\in{1..l}} $)
are used to compute the weighted sample mean $\mathrm{\mu}_i$, 
and weighted sample covariance $\mathbf{\Sigma}_i=(\mathbf{P}-[\mu_i]_{\times l})\mathbf{W}(\mathbf{P}-[\mu_i]_{\times l})^\mathbf{T}$.
The diagonal matrix $\mathbf{W}$ can be weighted uniformly ($\mathbf{W}=\mathbf{I}_{l\times l}$). However, weighting based on the reflected intensity: $\mathbf{W}_{j,j}=z_j-z_{min}$ (as discussed by Kung~\cite{kung2021normal}) is slightly beneficial, improving surface point consistency and registration. Note that $\mathbf{W}$ needs to be normalized by its trace $tr(\mathbf{W})$.
} 
By applying eigen-decomposition on the sample covariance $\mathbf{\Sigma}_i$, the surface normal
$\mathbf{n}_i$ can be estimated via the eigenvector that corresponds to the smallest eigenvalue $\lambda_{min}$, see Fig.~\ref{fig:surface_point_cell}. An additional filtering step is carried out by discarding all normals computed from fewer than $6$ points, or when all are from the same azimuth bin and points are located in a straight line (covariance close to singular). This typically occurs when the condition number $\kappa(\mathbf{\Sigma}_i)=\lambda_{max}(\mathbf{\Sigma}_i)/\lambda_{min}(\mathbf{\Sigma}_i) > 10^5$.
Where mean and normal have been successfully computed, the oriented surface point is then obtained as $m_i=\{\mathbf{\mu}_i,\mathbf{n}_i\}$. The full surface point set $\mathcal{M}^t=\{ m_i \}$ is visualized conceptually in Fig.~\ref{fig:orient_surface_final}, and with real data in Fig.~\ref{fig:oriented_surface_points} (on top of the source point cloud $\mathcal{P}^t$). 
The most important parameter $r$ is ideally selected according to the size of the environment and the scale of important landmarks -- a smaller environment requires a smaller radius to compute stable oriented surface points, and finding a radius that works well regardless of spatial scale and environment is potentially challenging. However, as shown in the evaluation we address this issue using the one-to-multiple scan registration technique presented in Sec.~\ref{sec:method_jointy_register}.


\subsection{Local mapping}
Reference scans and poses, commonly referred to as ``keyframes'', are stored $\{\mathbf{\mathcal{M}}^{k_s},\mathbf{T}^{k_s}\}\leftarrow\{\mathbf{\mathcal{M}}^t,\mathbf{T}^t\}$ in a sliding window, with a history (window size) of $s$.
Keyframes are stored at regular intervals when the \changedd{estimated pose} exceeds a distance in rotation or translation from a previous keyframe, hence mitigating drift when the sensor is not moving.

Previous work has shown that fusing keyframes into small local submaps of a few keyframes can be preferred over building larger monolithic maps~\cite{8870941}. However, instead of exploring update schemes for local maps~\cite{behley2018rss,jari_ndt_occupancy,8968140,kung2021normal}, we keep track and use the individual (dense radar) keyframes $k_{1..s}$ separately for the task of registration. This allows us to investigate the principled benefit of embedding additional information for future scan-to-map localization methods.


\subsection{Registration}
\label{sec:registration}

The current pose (2d) is estimated by finding the optimization parameters $\mathbf{x}^t=[ x \; y \; \theta ]^{T}$ that best align the latest scan
$\mathbf{\mathcal{M}}^t$ with the most recent keyframe $\mathbf{\mathcal{M}}^k$ according to: 
\begin{equation}
    \label{eq:reg}
    \underset{\mathbf{x}^t}{\mathrm{arg\: min}} f(\mathbf{\mathcal{M}}^k,\mathbf{\mathcal{M}}^t,\mathbf{x}^t).
\end{equation}
We solve this incrementally by first creating a set of correspondences (between nearby oriented surface points) $\mathcal{C} = \{ m^k_j, m^t_i\}$ and then minimizing the weighted \textit{scan-to-keyframe} cost function between consecutive surface point sets:
\begin{equation}
\label{eq:scan_to_keyframe}
 f_{s2k}(\mathbf{\mathcal{M}}^k,\mathbf{\mathcal{M}}^t,\mathbf{x}^t) = \sum_{\forall \{i,j\}\in\mathbf{\mathcal{C}}} w_{i,j}\mathcal{L}_\delta(g(m^k_j, m^t_i,\mathbf{x}^t)),
\end{equation}
i.e., the sum of squared distances between all correspondences, weighted by a Huber loss $\mathcal{L}_\delta$ function. 
Correspondence to a surface point   $m_i\in\mathbf{\mathcal{M}}^t$ is found by searching for the closest point $m_j\in\mathbf{\mathcal{M}}^k$ within a radius $r$, with the additional criterion that the correspondences must have similar surface normals within a tolerance $\arccos{(\mathbf{n}_j\cdot \mathbf{n}_i)}<\theta_\mathrm{max}$. Here we reuse the grid resolution $r$ from the computation of surface points from Sec.~\ref{sec:surface_point}. The intuition is that for aligned scans, nearby surface points are expected to lie within this distance.


The Huber loss $\mathcal{L}_\delta$, defined in \eqref{eq:huber}, makes the cost less sensitive to outliers~\cite{Huber1992,9013051}. This is done by piecewise reshaping the cost function to increase quadratically for small values and linearly for larger values:
\begin{equation}
    \label{eq:huber}
        \mathcal{L}_\delta(h)= 
\begin{cases}
    \frac{1}{2}h^2, & \text{if } |h|\leq \delta\\
    \delta(|h|-\frac{1}{2}\delta),              & \text{otherwise.}
\end{cases}
\end{equation}
\changed{
   Residuals are then weighted based on surface point similarity. We compute three similarity weights based on: surface point planarity, number of detections used to compute surface points, and directions of surface normals. 
\begin{equation}
\begin{split}
        w^{plan}_{i,j} & = f_{sim}(p_i,p_j)\\
        w^{det}_{i,j}  & = f_{sim}(l_i,l_j)\\
        w^{dir}_{i,j}  & = \max(\mathbf{n}_i \cdot \mathbf{n}_j,0)\\
\end{split}
\label{eq:weights}
\end{equation}
In \eqref{eq:weights}, $l$ is the number of observations used to compute each surface point as described in Sec.~\ref{sec:surface_point}. The planarity $p$ of a surface point is computed by
\begin{equation}
\label{eq:planarity}
p=log(1 + |\lambda_{max}/\lambda_{min}|),    
\end{equation}
and the similarity is computed as
\begin{equation}
f_{sim}(a,b)=2\min(a,b)/(a+b).
\end{equation}
The separate weights are finally combined as
\begin{equation}
\label{eq:combined_weights}
    w_{i,j}   = w^{plan} + w^{det} + w^{dir}.
\end{equation}}

In the literature on point cloud registration, an large number of cost functions have been proposed~\cite{huang2021comprehensive,yokozuka2021litamin2}. In this work, we investigate three common 2d cost metrics: point-to-point (P2P)~\cite{4767965,121791,4543181}, point-to-line  (P2L)~\cite{132043,hoppe_p2l,behley2018rss} and point-to-distribution (P2D)~\cite{Biber_ndt,martin_ndt}:



  \begin{align}
    &g_{P2P}(m^k_j, m^t_i,\mathbf{x}) = ||\mathbf{e}||^2,\label{eqn:p2p}\\
    &g_{P2D}(m^k_j, m^t_i,\mathbf{x}) = \mathbf{e}^T\mathbf{\hat{\Sigma}}_j^{k-1}\mathbf{e},\label{eqn:p2d}\\
    &g_{P2L}(m^k_j, m^t_i,\mathbf{x}) = ||\mathbf{n}^k_j \cdot \mathbf{e}||^2,\label{eqn:p2l}
  \end{align}
using 
\begin{align*}
    &\mathbf{e}=\mathbf{\mu}^k_j - (\mathbf{R}_{\theta}\mathbf{\mu}^t_i+\mathbf{t}_{x,y}),\\
    & \mathbf{\hat{\Sigma}}=(\mathbf{\Sigma}+\lambda\mathbf{I}),
\end{align*}
where $\mathbf{R}_{\theta}$ and $\mathbf{t}_{x,y}$ are the rotation matrix and translation vector created from the optimization parameters, and $\mathbf{n}^k_j$ and $\hat{\mathbf{\Sigma}}_j^{k-1}$ are the surface normal and covariance matrix (dampened by $\lambda=10^{-1}$ to address singularity) of the correspondence surface point. P2D can be understood as a middle-ground between P2P and P2L. Instead of strictly using the residual in the normal direction, the P2P distance is weighted according to surface uncertainty.

Our evaluation shows that P2L is slightly more accurate compared to P2P and P2D between consecutive frames as seen in Fig.~\ref{fig:submap_keyframes}.a. However, we found that P2L has a higher systematic bias and more quickly accumulates drift as seen in Fig.~\ref{fig:submap_keyframes}.(b-d)

\changed{
Applications building on top of the radar odometry estimation may need an estimate of pose uncertainty. Similar to Bentgsson and Baerveldt~\cite{BENGTSSON200329} and Censi~\cite{4209579}, we estimate uncertainty from the Hessian of the objective function once registration has been carried out. Registration covariance is computed as: $C(\mathbf{x}^t)=(\mathbf{J}^T\mathbf{J})^{-1}$ i.e. from the inverse of the Hessian, approximated by the Jacobian of the objective function $J=\partial f_{s2k}/\partial \mathbf{x}^t_i $.} 


\subsection{Jointly registering to multiple keyframes}
\label{sec:method_jointy_register}
While \eqref{eq:scan_to_keyframe} can be solved directly, we propose an improvement that
integrates more information into the registration and provides temporal redundancy.
This is realized by extending the sum in \eqref{eq:scan_to_keyframe} over a history of keyframes $\mathcal{K}=\{k_1..k_s\}$ to compute the \textit{scan-to-multiple-keyframes} cost:
\begin{equation}
\label{eq:scan_to_multikeyframes}
 f_{s2mk}(\mathbf{\mathcal{M}}^{\mathcal{K}},\mathbf{\mathcal{M}}^t,\mathbf{x}^t) =\sum_{k\in\mathcal{K}}\sum_{\forall \{i,j\}\in\mathbf{\mathcal{C}}} w_{i,j}\mathcal{L}_\delta(g(m^k_j, m^t_i,\mathbf{x}^t)).   
\end{equation}
\changed{%
Consequently, each surface point in $\mathcal{M}^t$ is allowed up to one correspondence for each keyframe. We omit to weight the individual keyframes in favor of letting the gradually decreasing overlap (and the number of residuals) reduce the impact of more distant keyframes.
A graphical intuition for the joint registration and the residual weights is given in Fig.~\ref{fig:correspondence_intuition}.
\begin{figure}
\begin{centering}
   \includegraphics[trim={0cm 0cm 0cm 6cm},clip,width=0.99\hsize,angle=0]{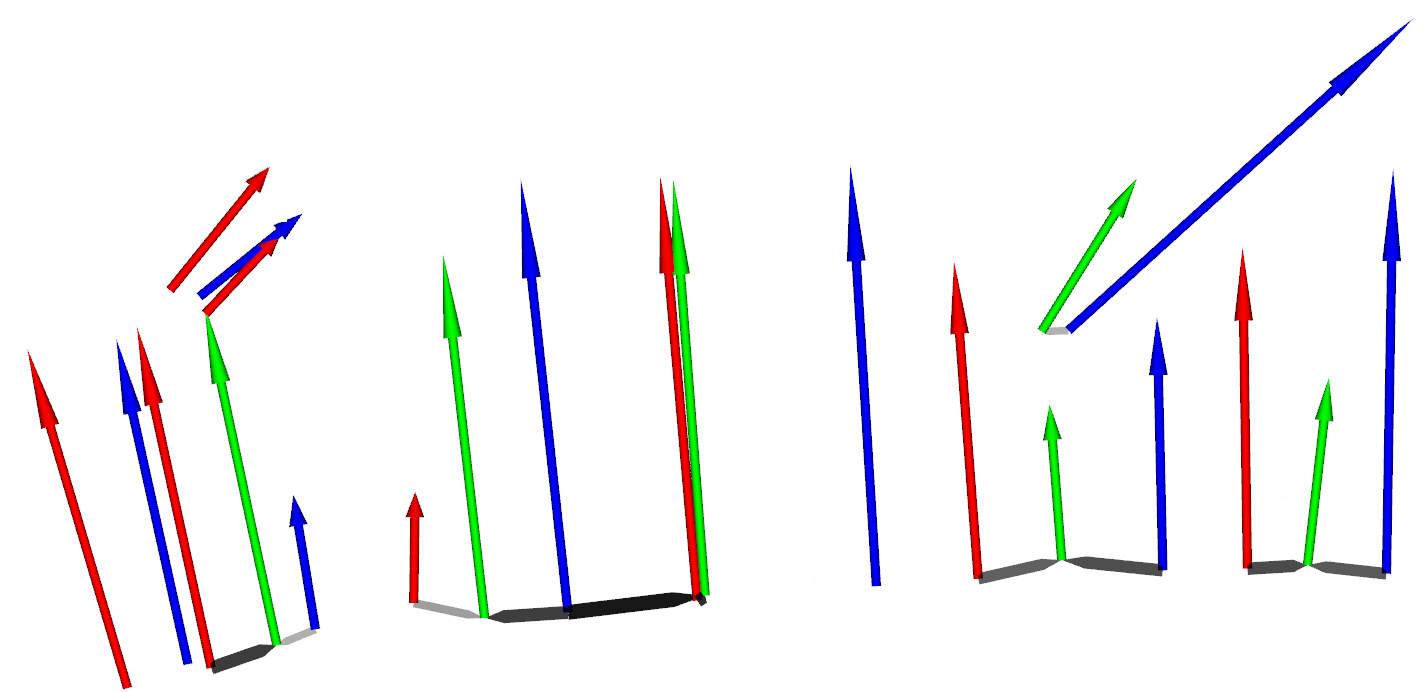}\caption{%
   \changed{
   Surface points from the latest scan (green) and two keyframes (blue \& red), and computed correspondences (grayscale).
   Each green surface point is assigned up to one correspondence per keyframe. Residuals are weighted based on similarity ~\eqref{eq:combined_weights} as indicated by grayscale and width: ranging from thin-white (low weight) to thick-black (high weight). Our matching step alternates between assigning correspondences (seen in this figure), and minimizing~\eqref{eq:scan_to_multikeyframes}. \label{fig:correspondence_intuition}}
   }
  \end{centering}
  \vspace{-0.4cm}
\end{figure}

The formulation in~\eqref{eq:scan_to_multikeyframes} has multiple advantages over~\eqref{eq:scan_to_keyframe}: it overcomes uncertainty from matching sparse or noisy feature sets by additionally constraining the objective function with more correspondences. As a consequence, drift and pose noise are reduced, and tolerance to feature-poor environments is improved as demonstrated in Sec~\ref{sec:kvarntorp_sec}. Moreover, having multiple correspondences at different time points provide temporal redundancy. For example, during sudden occlusions or alternating ground observations (typically caused by vehicle suspension during acceleration). Under such circumstances, traditional consecutive scan matching may fail as the problem is under-constrained. In contrast, our formulation allows correspondences to landmarks found in multiple scans and does not rely on individual scans being complete. In less troublesome scenarios, spurious correspondences from moving objects will have a relatively lower impact compared to stable reappearing landmarks. Hence, the formulation implicitly mitigates scene dynamics without e.g. occupancy grid mapping~\cite{jari_ndt_occupancy,grid_map_wolfram} or dynamic object segmentation~\cite{Pfreundschuh2021DynamicOA}.

A possible risk from considering a longer history of scans is that range measurement bias~\cite{glennie2010static} can increase when matching scans from more distant locations. The measurement bias can lead to sensor movement being systematically underestimated~\cite{8793671}. A more detailed discussion follows in Sec.~\ref{sec:eval_submap_keyframes}.
}%
The time of computing the cost~\eqref{eq:scan_to_multikeyframes} scales roughly linearly with the number of keyframes, and if the set of oriented surface points $\mathcal{M}^t$ is sparse, the scaling factor is low, and extending the cost function with more keyframes is inexpensive.

\subsection{Solving}
\label{sec:solving}
A starting point for the optimization is predicted using the constant velocity model: 
\begin{equation}
\label{eqn:solving}
\mathbf{x}_0^t=\mathbf{x}^{t-1}+\mathbf{\dot{x}}^{t-1} \Delta T.
\end{equation} 
We alternate between finding correspondences $\mathcal{C}$ and minimizing
\eqref{eq:scan_to_multikeyframes}. This iterative process is repeated until either: the cost can be minimized using only a single step, a maximum of 8 iterations has been reached, or the relative decrease is below a limit $\epsilon$. We use the Trust Region Levenberg-Marquardt method with gradients obtained through automatic differentiation using the Ceres non-linear optimization framework~\cite{ceres-solver}.





\section{Evaluation}
\label{sec:evaluation}


In this section, we evaluate our method with an extensive ablation study, a comparison study in a public urban benchmark and a generalization study that investigates how environment type, spatial scale, sensor configuration and parameters affect the odometry pipeline.

Currently, there are four published datasets with ground truth poses which target large-scale spinning radar odometry research: 
Boreas~\cite{boreas_dataset}, Radiate~\cite{sheeny2020radiate},
MulRan~\cite{gskim-2020-mulran} and the extensively utilized Oxford Radar RobotCar dataset~\cite{RadarRobotCarDatasetICRA2020}. In previous publications, most methods for spinning radar odometry have been evaluated on the latter two datasets. In addition to the urban Oxford dataset, we chose MulRan to enable a comparison with varied spatial scale and structural diversity of the environment~\cite{gskim-2020-mulran}, including urban, river, mountain and bridge crossings. In MulRan, the maximum speed ($63km/h$) is slightly higher compared to Oxford ($46km/h$), but the acceleration is typically lower as seen in Fig.~\ref{fig:acceleration_and_velocity}.
Throughout the experiments, we follow the KITTI odometry benchmark~\cite{Geiger2012CVPR} that estimates drift by computing the overall translation error ($\%$) and rotation error (deg/m) averaged over all sub-sequences between $\{100,200, \ldots, 800\}$~m. 
\changed{
Additionally, we compute the mean version of Relative Pose Error (RPE)  between two consecutive poses as presented in Sturm et al.~\cite{6385773} ($\Delta =1$ in their paper).
\begin{equation}
\label{eq:pose_error}
    \mathbf{E}_i:=(\mathbf{Q}^{-1}_i \mathbf{Q}_{i+1})^{-1}(\mathbf{P}^{-1}_i \mathbf{P}_{i+1}),
\end{equation}
\begin{equation}
    \label{eq:rpe}
    \text{RPE}(\mathbf{E}_{1:(n-1)}) =
    \frac{1}{n-1}\sum_{i=1}^{n-1} ||transl(\mathbf{E}_i)||,
\end{equation}
where $transl(\mathbf{E}_i)$ is the translation component of the relative pose error $ \mathbf{E}_i $. $\mathbf{P}_{1..n}\in SE(2)$ and $\mathbf{Q}_{1..n}\in SE(2)$ is the estimated and ground truth trajectory, respectively. In part of our evaluation (Sec.~\ref{sec:eval_submap_keyframes}), the RPE mean is complemented by its bias, calculated from \eqref{eq:rpe} without vector norm in order to retain the direction of each error term. This allows us to measure if movement is being under- or over-estimated in translation, and similarly in rotation.
}
In contrast to odometry drift, which is computed over longer trajectories, RPE is averaged over two consecutive pose estimates (displaced with up to $\sim4$~m), and infrequently occurring odometry failures might not be observed in the RPE while having a large impact on the translation and rotation error (drift). Hence, we distinguish between RPE (pose accuracy) and drift, where the latter should be interpreted as an improved measure of odometry robustness. 

We use the odometry benchmark implementation from~\cite{zhan2019dfvo}. For transparency and to promote comparison, we provide all our data and the scripts to regenerate the figures on our GitHub page\footnote{Evaluation: \url{https://github.com/dan11003/CFEAR_evaluation}}.

The full parameter set of CFEAR together with the four proposed configurations evaluated here is listed in \cref{tab:Parameter}. Configuration CFEAR-1 (single keyframe) and CFEAR-2 (3 keyframes) are similar to our previous work~\cite{adolfsson2021oriented} and \cite{adolfsson2021cfear} respectively, but with an improved solving strategy that increases run-time-performance as proposed in Sec.~\ref{sec:solving}, and with weighted surface point locations and residuals. Other minor adjustments compared to our previous papers are: increased intensity threshold $z_{min}$, lower minimum distance and unlimited max range. The configuration CFEAR-3 proposed in this paper uses a combination of 4 keyframes, less conservative filtering ($k=40$) and the P2P cost function.
These parameters were tuned jointly for MulRan and Oxford, striking a trade-off between speed and accuracy.
\changed{
Additionally, we propose drift optimized \textit{CFEAR-3-s50} where we extend the history to 50 keyframes and adopt the Cauchy loss function.
Finally, we present the configuration ``\textit{baseline odometry}'' ( equivalent to CFEAR-3 with a single keyframe $s=1$). The method can be considered as a traditional scan matching method that minimizes a point-to-point metric between consecutive frames. \textit{baseline odometry} is used in qualitative experiments as a reference to demonstrate the benefit of multiple keyframes, when grid resolution is poorly tuned to the scale of the environment.

}
\subsection{Quantitative ablation study within the Oxford dataset}
\label{sec:ablation_study}
We start by evaluating the components of the odometry pipeline and parameters separately, aiming to understand their relative importance. All experiments in this section are based on the most accurate configuration CFEAR-3 (with $k=12$ instead of $k=40$ to speedup experiments) using the Oxford Radar RobotCar dataset, any changes are explicitly stated. Reported numbers are averaged at least over 8 sequences. Timings statistics are computed for experiments running in parallel on an Intel i7-5930k. No multi-threading is used to speed up execution times.

\begin{table}
\centering
\begin{adjustbox}{width=\hsize}
\begin{tabular}{l|lllll}
& \multicolumn{3}{c}{\textbf{Configurations }}
\\
\textbf{Parameter} & CFEAR-1 & CFEAR-2  & CFEAR-3  & CFEAR-3-s50 & Baseline odom.        \\    
Description & Efficient & Balanced & Low drift & Extreme & Traditional\\
\hline
\kstrongest{}: $k$ & $\mathbf{12}$ & $\mathbf{12}$ & $\mathbf{40}$  & $\mathbf{40}$ & $\mathbf{40}$\\ 
Expected noise: $\zmin$ & $\mathbf{70}$ & $\mathbf{70}$ & $\mathbf{60}$ & $\mathbf{60}$& $\mathbf{60}$ \\  
Resample factor: $f$ & $1$ & $1$ & $1$ & $1$  & $1$\\
resolution: $r$ & $\mathbf{3.5}$ & $\mathbf{3.5}$ & $\mathbf{3}$ & $\mathbf{3}$ & $\mathbf{3}$ \\  
Correspondence: $\theta_\mathrm{max}$ & $30$ & $30$ & $30$ & $30$ & $30$ \\    
Loss: (Huber/Cauchy)   & Huber & Huber & Huber & Cauchy & Huber \\
Loss limit:   $\mathcal{L}_\delta$ & $0.1$ & $0.1$ & $0.1$ & $0.1$  & $0.1$ \\
Keyframe dist: [m]/[deg] & $1.5/5^\circ$ & $1.5/5^\circ$ & $1.5/5^\circ$ & $1.5/5^\circ$ & $1.5/5^\circ$ \\ 
Submap keyframes: $s$ & $\mathbf{1}$ & $\mathbf{3}$ & $\mathbf{4}$ & $\mathbf{50}$  & $\mathbf{1}$ \\  
Residual weight: $w$ & comb.~\eqref{eq:combined_weights} & comb.~\eqref{eq:combined_weights} & comb.~\eqref{eq:combined_weights} & comb.~\eqref{eq:combined_weights} & comb.~\eqref{eq:combined_weights} \\  
Min dist.: $d_{min}$ & $2.5$ & $2.5$ & $2.5$ & $2.5$ & $2.5$ \\
Cost function: $g$ & 
\textbf{P2L} \eqref{eqn:p2l}  &
\textbf{P2L} \eqref{eqn:p2l}  &
\textbf{P2P} \eqref{eqn:p2p}  &
\textbf{P2P} \eqref{eqn:p2p}  & 
\textbf{P2P} \eqref{eqn:p2p}
\end{tabular}
\end{adjustbox}
\caption{Full list of parameters. The configurations \textit{CFEAR-(1-3)} are efficient, being jointly optimized for speed and drift. \textit{CFEAR-3-s50} is optimized for drift only, aiming at investigating the benefit of an extended history. \textit{Baseline odometry} is used as a reference in the qualitative experiments in Sec.~\ref{sec:qualitative_eval}.
}\label{tab:Parameter}
\vspace{-0.3cm}
\end{table}

\subsubsection{\kstrongest}
\begin{figure}
    \subfloat[][Odometry drift wrt. $k$ and $z_{min}$.]{\includegraphics[trim={0.2cm 0cm 0.3cm 0cm},clip,height=0.5\hsize]{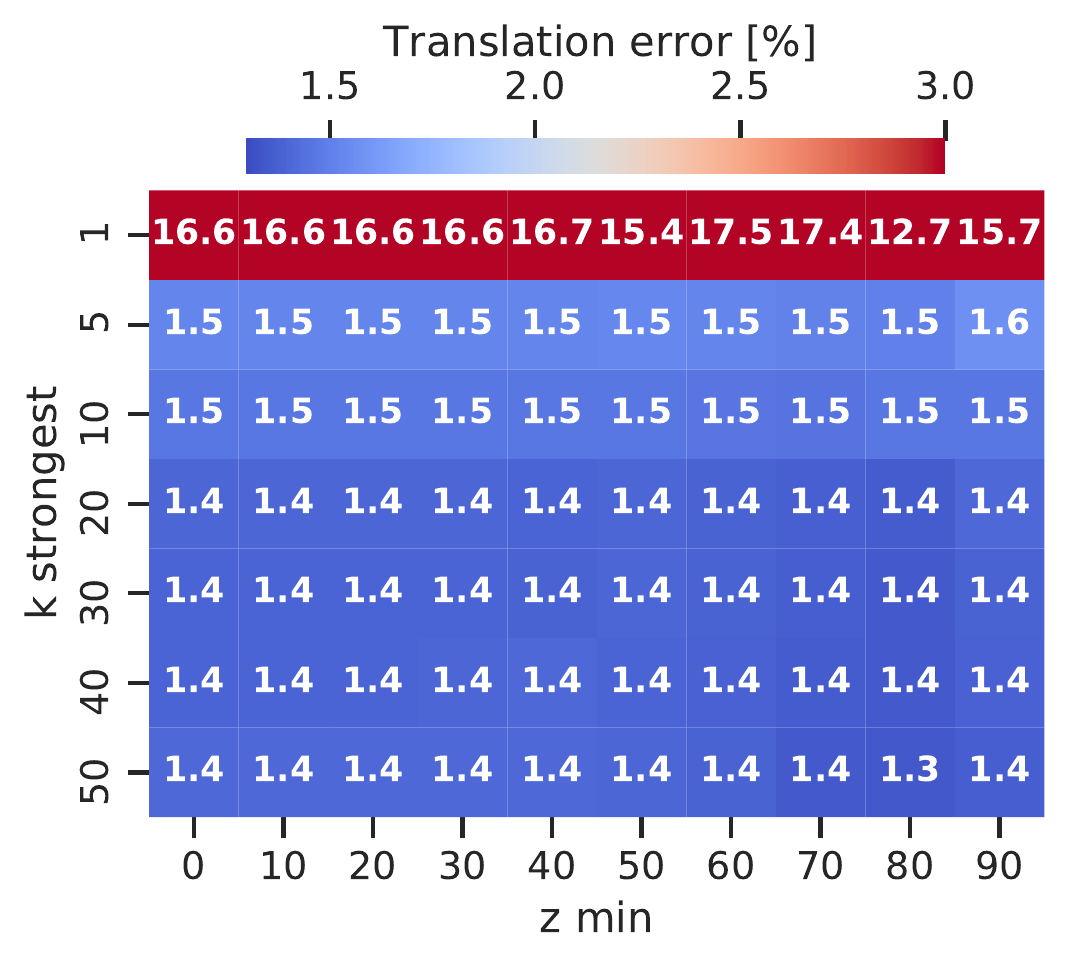}\label{fig:KstrongestAccuracy}}
    \hfill
    \subfloat[][Filtering time wrt. $k$ and $z_{min}$.]{\includegraphics[trim={1.43cm 0cm 0.3cm 0.0cm},clip,height=0.5\hsize]{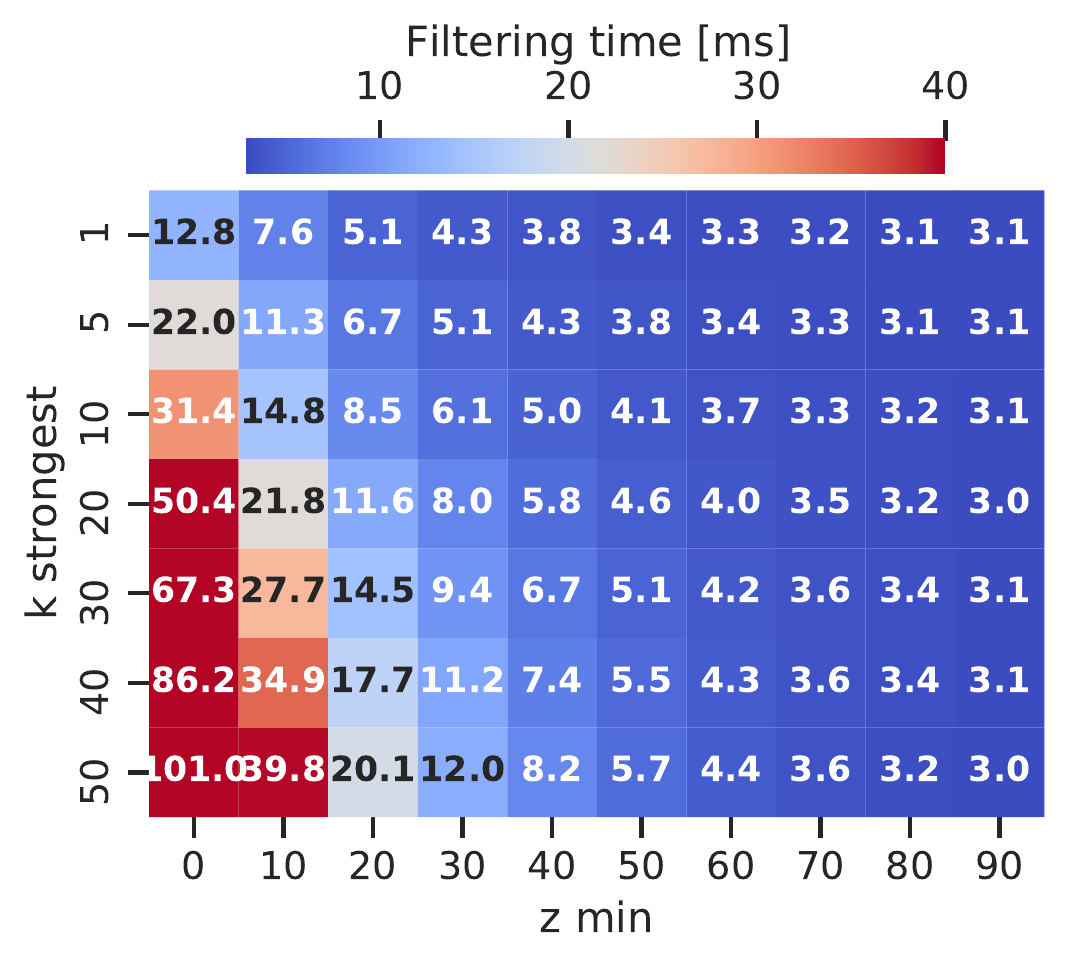}\label{fig:KstrongestTime}}\hfill
    \caption{How \kstrongest \: parameters affect drift and run-time performance of the CFEAR pipeline. CFEAR is largely insensitive to the selection of parameters.}\label{fig:kstrongfilter}
    \vspace{-0.5cm}
    \end{figure}

In the following, we first evaluate the influence of the {\kstrongest} parameters $z_{min}$ and $k$ presented in Sec~\ref{sec:kstrong}.
As depicted in Fig.~\ref{fig:KstrongestAccuracy}, keeping only a few returns $k=5$ is sufficient to obtain results better than the previous state-of-the-art ($1.76\% $ translation error), however, keeping a larger amount of returns per azimuth $\geq 20$ yields even lower drift.
  \begin{figure}
    \subfloat[][Odometry drift wrt. \textit{window size} \\and \textit{false alarm rate}.]
    {\includegraphics[trim={0.3cm 0cm 0.2cm 0cm},clip,height=0.66\hsize]{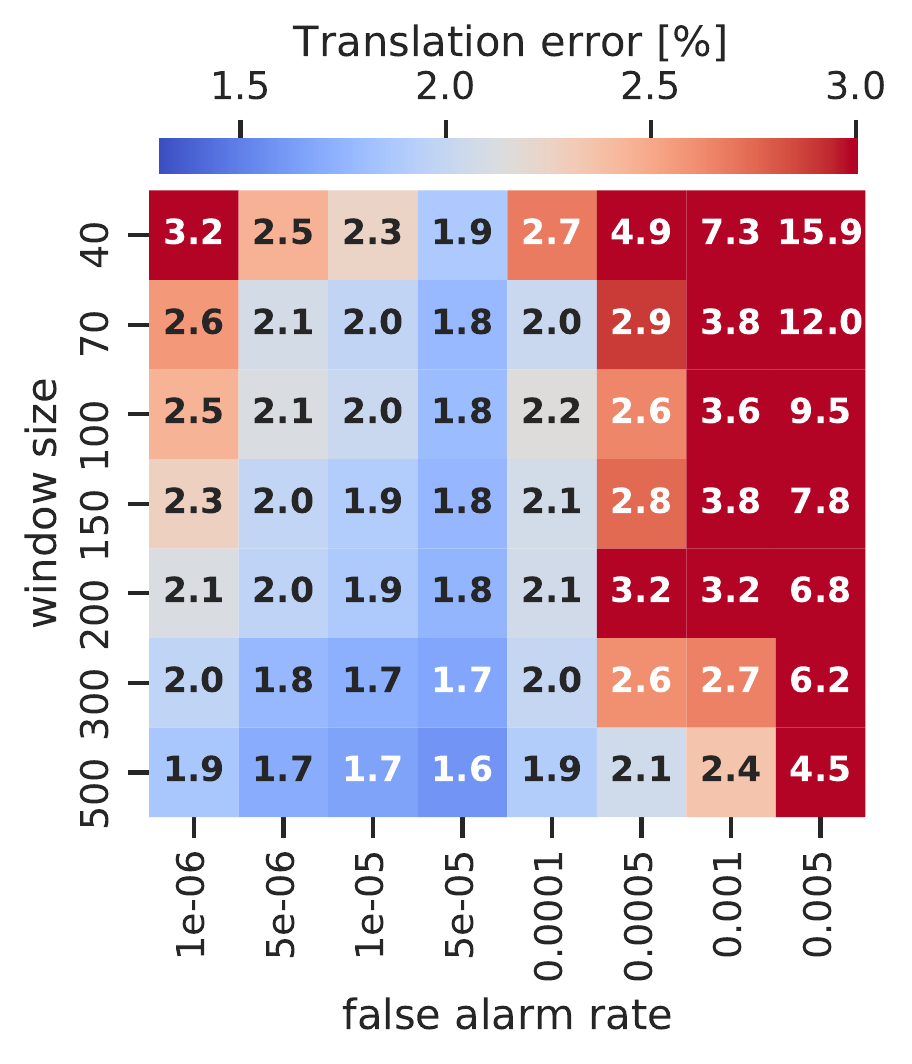}\label{fig:CA-CFAR-drift}}\hfill
    \subfloat[][Filtering time wrt. to \textit{window size} and \textit{false alarm rate}.]{\includegraphics[trim={1.6cm 0cm 0.1cm 0.0cm},clip,height=0.66\hsize]{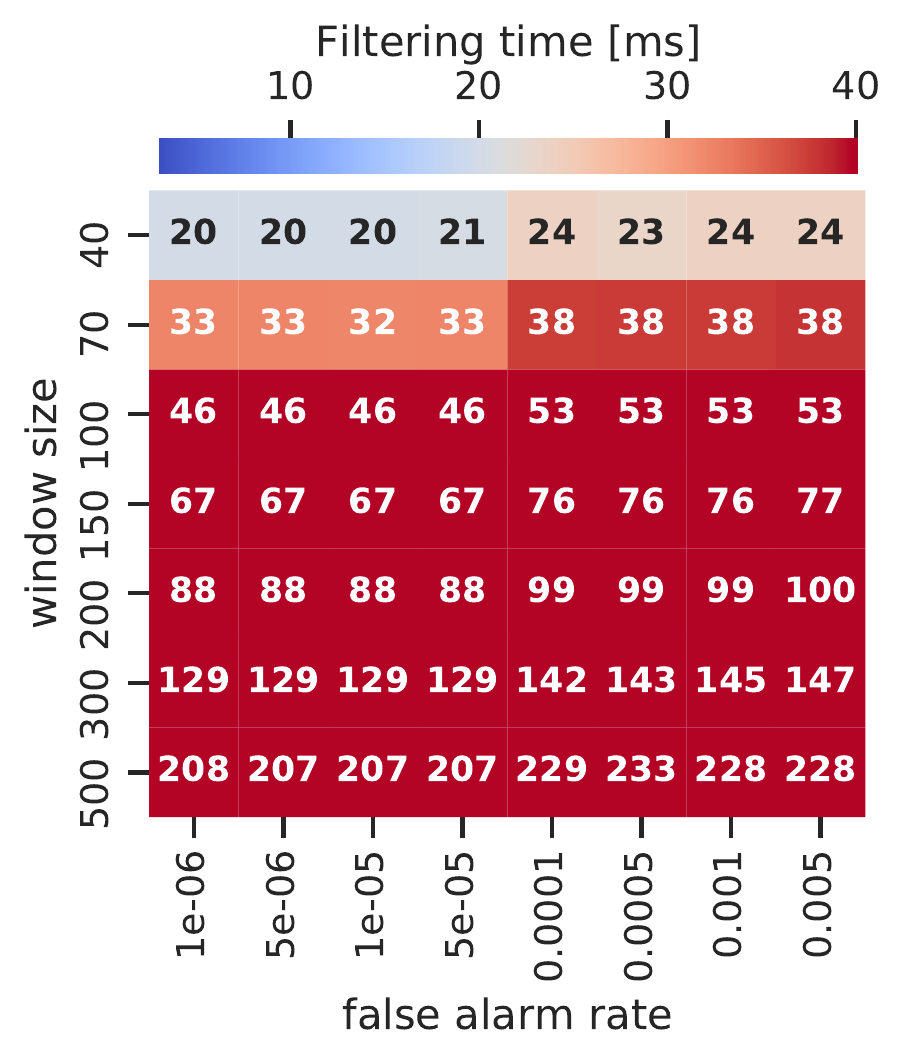}\label{fig:CA-CFAR-time}}\hfill
    \caption{Drift and filtering time when using the radar filter CA-CFAR (instead of \kstrongest{}) within the CFEAR pipeline.
    Careful parameter selection is needed to achieve drift comparable to \kstrongest{}. }\label{fig:CA-CFAR}
    \vspace{-0.4cm}
\end{figure}
This is intuitive as there might be multiple landmarks within a line-of-sight that can be observed within a single beam, and keeping more returns provides more readings around non-primary landmarks that can be used to compute additional surface points to further constrain the registration.

$z_{min}$ can be increased up to $z_{min}=80$ without any noticeable effect on the drift. After that point, the intensity of important landmarks falls below the intensity threshold and the odometry quickly deteriorates for $z_{min}\geq 100$. Hence, choosing $k>5$ and $z_{min}<80$ is favorable to reduce drift. However, as depicted in Fig.~\ref{fig:KstrongestTime}, run-time performance is reduced when choosing $z_{min}<20$ and $k>10$, during which \kstrongest{} evaluate the intensity between an excessive amount of detections.
Conveniently, the parameters can be chosen in a wide range: $50\leq z_{min}\leq 80$ and $10\leq k\leq 50$ in which both drift and run-time performance of the filter speed and odometry are stable.
Within this range, some level of fine-tuning can be done, providing a trade-off between detection quality and speed. The proposed filter settings in Tab.~\ref{tab:Parameter} strike a balance between speed and accuracy over the full odometry pipeline, e.g. additional detections influence run-time performance in the later stages.
\changed{
The filter itself is affected by sensor resolution. A finer resolution may require a higher $k$ to maintain similar detection properties. A coarser sensor resolution requires a lower $k$ for similar mitigation of multipath reflections. In our experiments we observe that the odometry itself is fairly resilient to different sensor resolutions in the range of 4.38--17.5~cm without parameter changes.
}
\changed{
\subsubsection{Comparison to CA-CFAR}
In order to validate the effectiveness of the \kstrongest{} filter, we replaced  \kstrongest{} with the classical radar target detector CA-CFAR~\cite{4102829,335950}. CA-CFAR adaptively estimates the background noise surrounding the \textit{cell under test} by the average intensity within a window, excluding the \textit{cell under test} itself and adjacent \textit{guard cells}.
Detections are provided from cells exceeding the threshold $S=TZ$, where $Z$ is the adaptively estimated noise level, and $T$ is a scaling factor 
calculated from the \textit{false alarm rate} parameter.
We use a standard, non-optimized implementation of CA-CFAR with a fixed number of \textit{guard cells} = 10 (which we found reasonable within the evaluated dataset), and adjust the \textit{false alarm rate} and \textit{window size} similarly to the parameter analysis in the previous paragraph.
To speed up the evaluation, we omit all cells with very low intensity $z_i<20$ as we found that real objects in the scene are observed with higher intensity levels. 
The result is depicted in Fig.~\ref{fig:CA-CFAR}. The lowest drift ($1.6\%$) was achieved with a larger \textit{window size} (lower local adaptivity).
Regardless of settings, the drift remains larger compared to \kstrongest{}. We hypothesize that the relatively lower performance can be attributed to its higher vulnerability to speckle noise and multi-path reflections, which are characteristic of our experimental setup. 
Other versions of CFAR~\cite{alhashimi2021bfarbounded} that specifically account for these effects may be more suitable within these settings.
}
 \begin{figure}[!b]
  \begin{center}
    \subfloat[][Translation error vs key frames ($s$).]{\includegraphics[trim={0.0cm 0cm 0.0cm 0cm},clip,width=0.49\hsize]{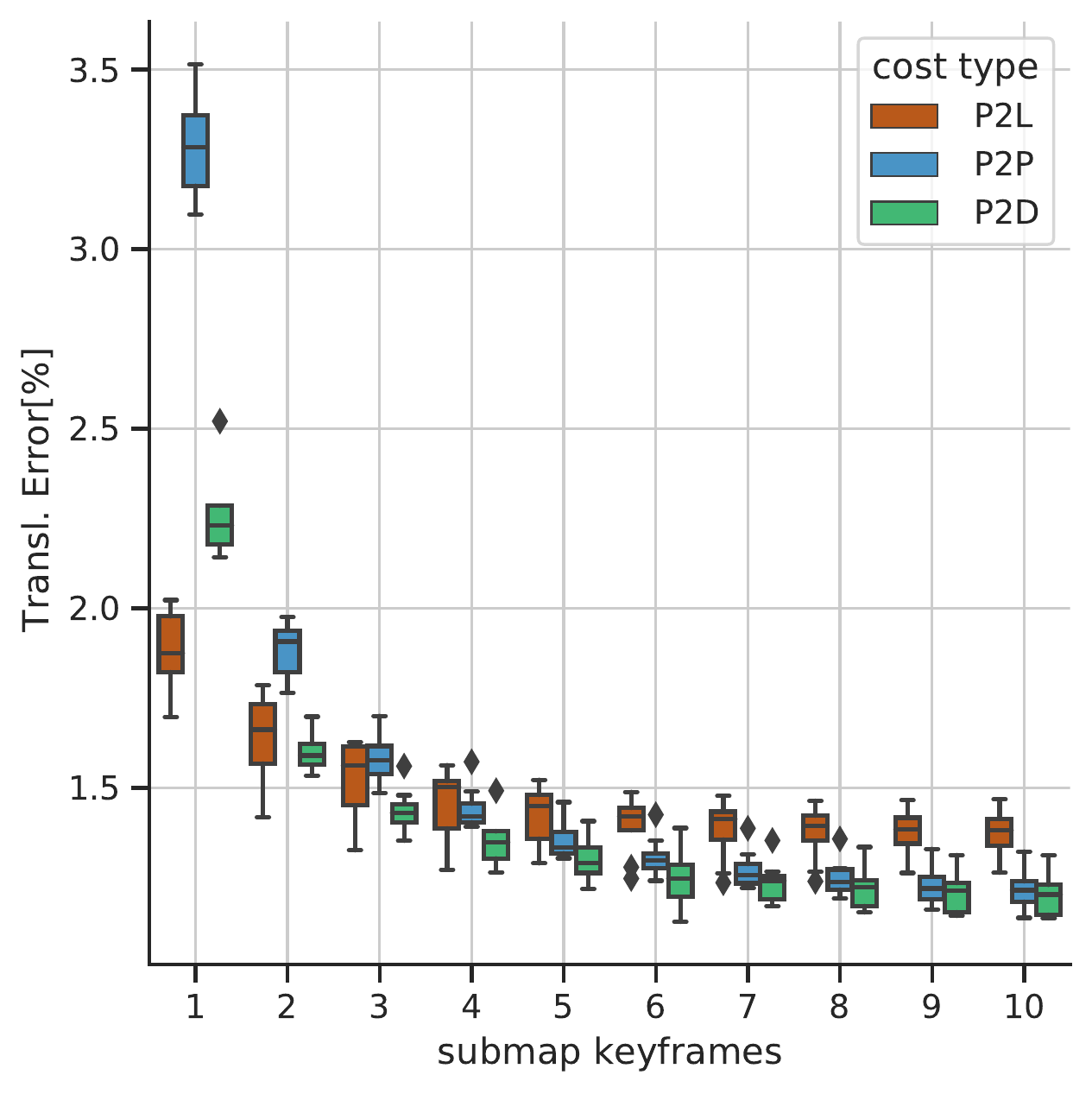}\label{fig:submap_keyframe_transl}}\hfill
     \subfloat[][RPE vs submap keyframes ($s$). ]{\includegraphics[trim={0.0cm 0cm 0.0cm 0cm},clip,width=0.49\hsize]{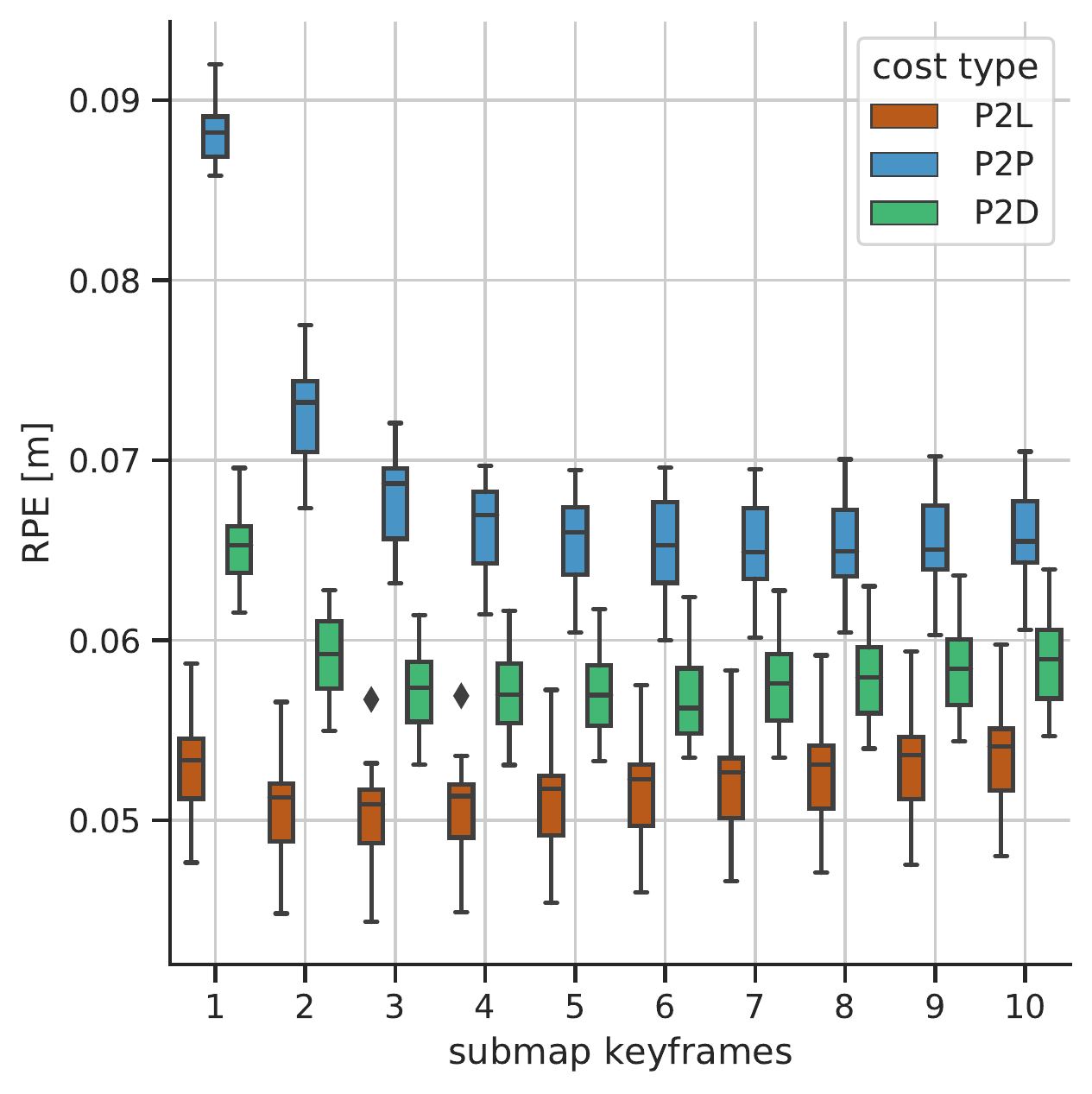}\label{fig:submap_keyframe_RPE}}\\
    \subfloat[][Longitudinal bias vs keyframes. Red dashed line marks zero bias.]{\includegraphics[trim={0.0cm 0cm 0.0cm 0.0cm},clip,width=0.49\hsize]{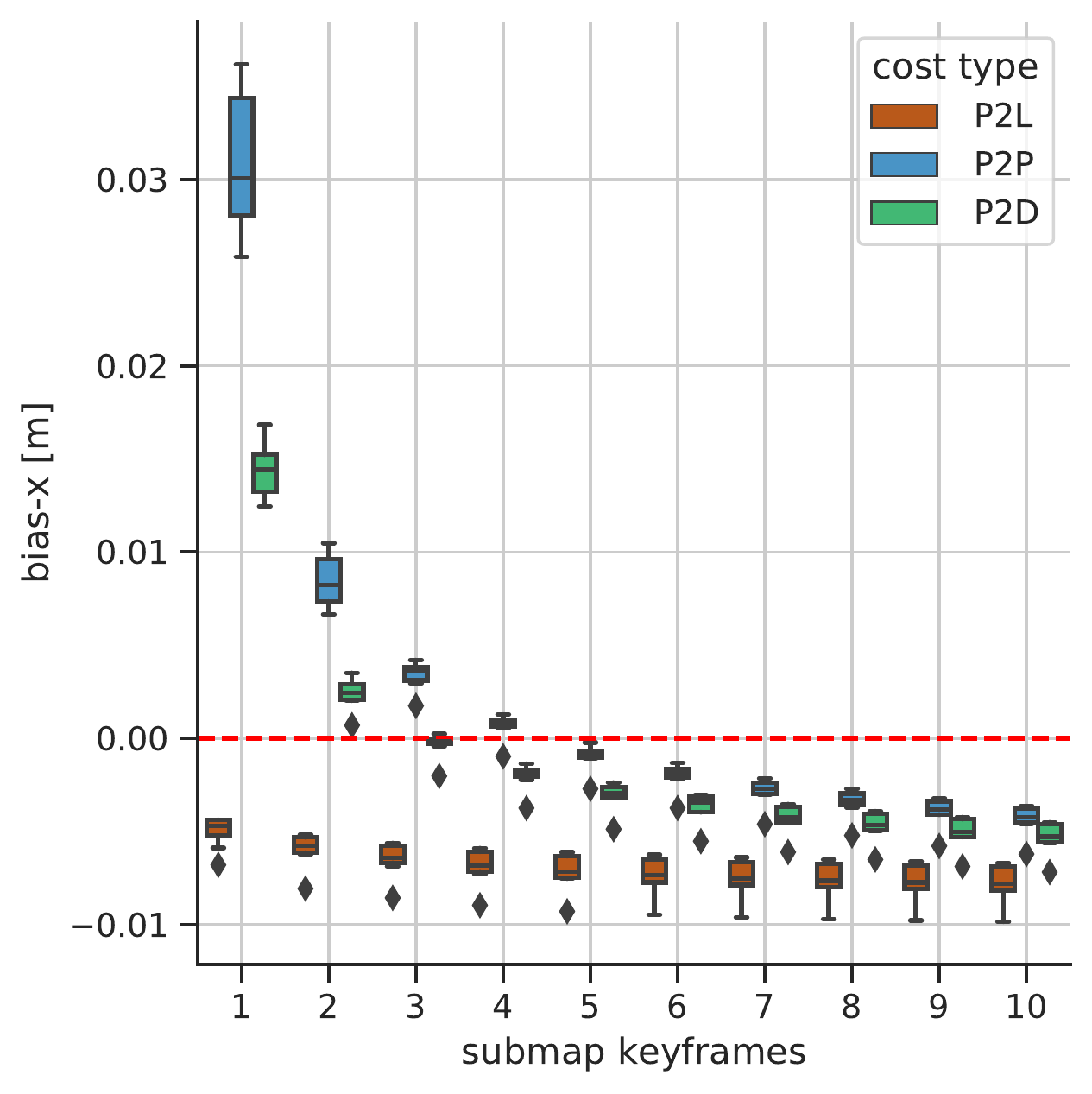}\label{fig:submap_keyframe_bias_x}}\hfill
      \subfloat[][Rotation bias vs keyframes. Red dashed line marks zero bias.]{\includegraphics[trim={0.0cm 0cm 0.0cm 0.0cm},clip,width=0.49\hsize]{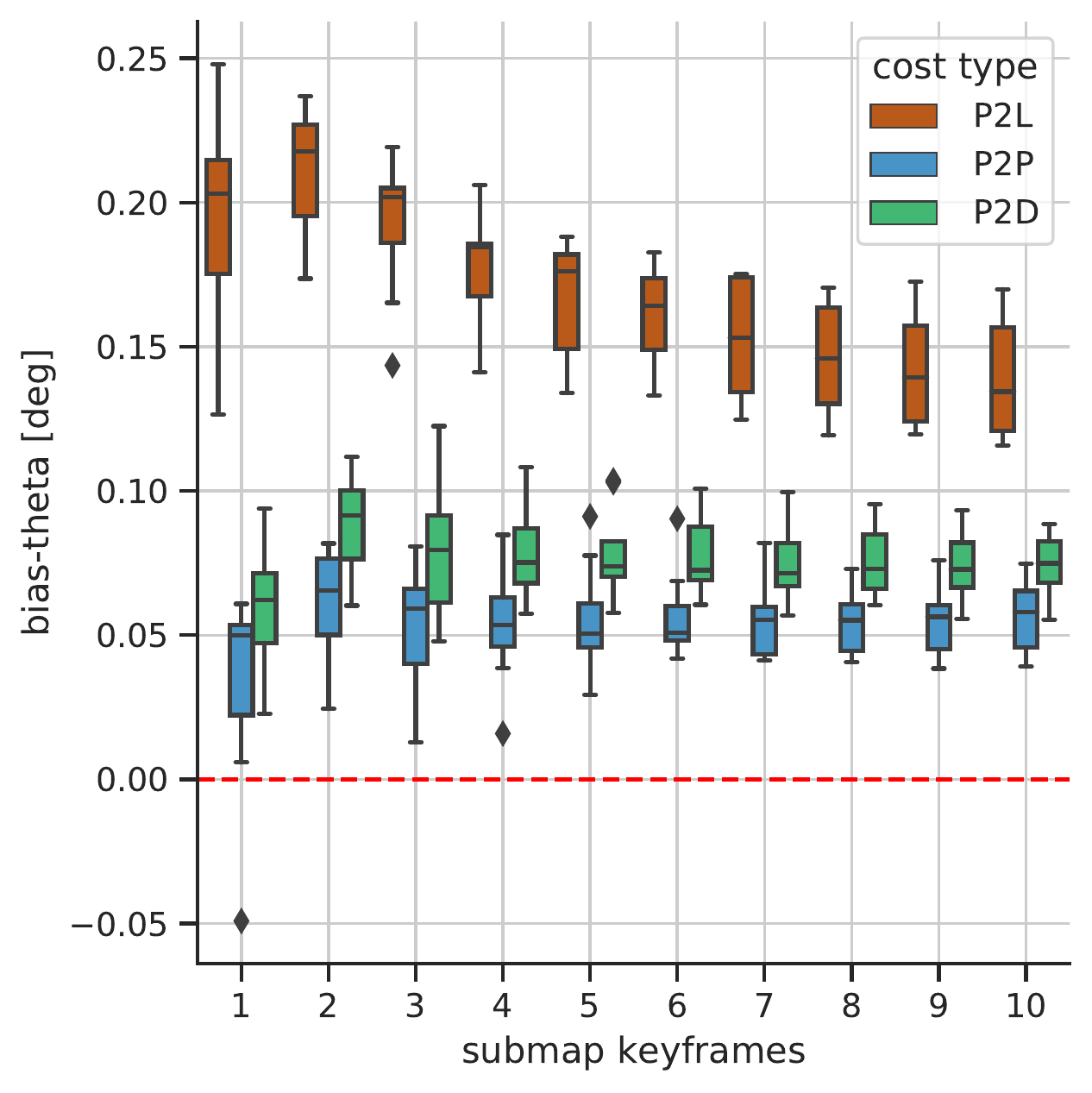}\label{fig:submap_keyframe_bias_rot}}\hfill
    \caption{(a): drift (translation), (b): accuracy (RPE) and (d,e): longitudinal and rotation bias with respect to the number of submap keyframes $(s)$ against which the current scan is registered, for P2L, P2P and P2D. Increasing the number of keyframes allows P2P and P2D to overcome sparsity and longitudinal bias, and to outperform P2L in odometry quality.  }\label{fig:submap_keyframes}
  \end{center}
  \vspace{-0.5cm}
\end{figure}

\subsubsection{Sparsity of surface point set}
By subdividing the number of data points at multiple steps, our pipeline gradually shifts computational focus to landmarks in the scene while reaching a maintainable and lightweight point set. Our preprocessing pipeline reduces the number of data points in the following steps:
First, in the Oxford dataset, that use a sensor configuration of $N_a=400$ azimuth and $N_r=3768$ range bins), \kstrongest{} reduces the original $N_a\times N_r=1.6e6$ intensity readings to a much smaller point set $|\mathcal{P}_t|$ with a maximum size of  ($|\mathcal{P}_t|<N_a\times k$). Note that the set decrease if fewer than $k$ intensities (exceeding $z_{min}$) were found within an azimuth bin, e.g when landmarks are absent. Second, the filtered point set is used to compute an even sparser set of surface points where the output set size is largely influenced by grid resolution parameter $r$.
The number of points remaining after each step is presented in Tab.~\ref{tab:two_step_filter}.

\begin{table}[h!]
\centering
\begin{tabular}{l|lll}
& \multicolumn{3}{c}{\textbf{Datapoints remaining after each step}}\\
\textbf{Config.} & Raw  & \kstrongest $|\mathcal{P}_f|$ & Surface points $|\mathcal{M}|$\\    
\hline
CFEAR-(1\&2)  & $1.5e6$ & $4.75e3\pm 83 (0.3\%)$ & $143\pm 38 (0.01\%)$\\
CFEAR-3  & $1.5e6$ & $1.3e4\pm 1337 (0.09\%)$ & $301\pm 87(0.02\%)$ \\
\end{tabular}

\caption{
\changed{
Absolute quantity and percentage of data points remaining after {\kstrongest} filtering, and after computing oriented surface points.}
}
\label{tab:two_step_filter}
\end{table}


\subsubsection{Submap keyframes}
\label{sec:eval_submap_keyframes}
The advantage of registering the set of surface points in the current scan $\mathcal{M}^t$ jointly towards multiple ($|\mathcal{K}|=s$) keyframes $\mathcal{M}^{\mathcal{K}}$ (rather than a single keyframe) is investigated in Fig.~\ref{fig:submap_keyframes}.

As seen in Fig.~\ref{fig:submap_keyframes}(a,b), when registering against a single submap keyframe, P2L yields the lowest level of drift in translation, and lower RPE compared to P2P and P2D. This is an indication that considering the surface point geometry (i.e. normal) in the cost function makes the scan-to-scan registration less sensitive to sparsity.

Using more keyframes is always advantageous for drift independently of the cost function. Surprisingly, for a history of 4 or higher keyframes, P2P and P2D achieve significantly higher odometry quality compared to P2L. The likely reason is that point-to-point matching works poorly with sparse point sets, which is partly addressed by registering to additional keyframes that essentially increase the density of the reference scan in the registration.
For a larger number of keyframes($\geq4$), we observe that the P2L yields the lowest RPE (pose noise), but at the same time higher drift. Part of the higher drift can be explained by a combination of the higher rotational and longitudinal bias compared to P2P/P2D. The rotational and longitudinal bias is shown in
Fig.~\ref{fig:submap_keyframes}(c,d), zero bias is marked with a red dashed line. 

\changed{
For 3 or more keyframes P2L tends to underestimate the movement in the longitudinal direction to a higher extent compared to P2P and P2D. Similarly, the estimated movement is lower for P2L within the MulRan dataset.
The rotational bias is consistently highest for P2L and lowest for P2P. 
The translation and rotation bias (typically $<1~cm$ and $<0.2deg$) contribute to a minor part of the RPE ($5-7~cm$), however it accumulates into drift over long distances. 
We hypothesize that the bias can be explained by that the estimated directions of surface normals are slightly shifted towards the sensor location. This shift can be explained by that accurately reconstructing planar surfaces, observed at high incident angles with high beam-width sensors such as radars is challenging. In such cases, the measured range to surfaces is underestimated in proportion to the incident angle. The measurement bias propagates into underestimation of ego-motion. This effect has previously been studied with lidar measurements within narrow corridors~\cite{8793671} and we believe that a similar effect is to be expected within our experimental setup (narrow streets and high beam width) and can explain our results.
 
}

In addition to addressing sparsity issues, we believe the drastic improvement from using multiple keyframes can additionally be explained by robustness to dynamics and overlap variations. In such cases, the keyframe history adds redundancy by relying less on individual scans.

 \begin{figure}[h!]
 \vspace{-0.3cm}
  \begin{center}
    \subfloat[ ][RPE vs resolution $r$.]{\includegraphics[trim={.1cm 0.1cm 1.3cm 0.1cm},clip,width=.49\hsize]{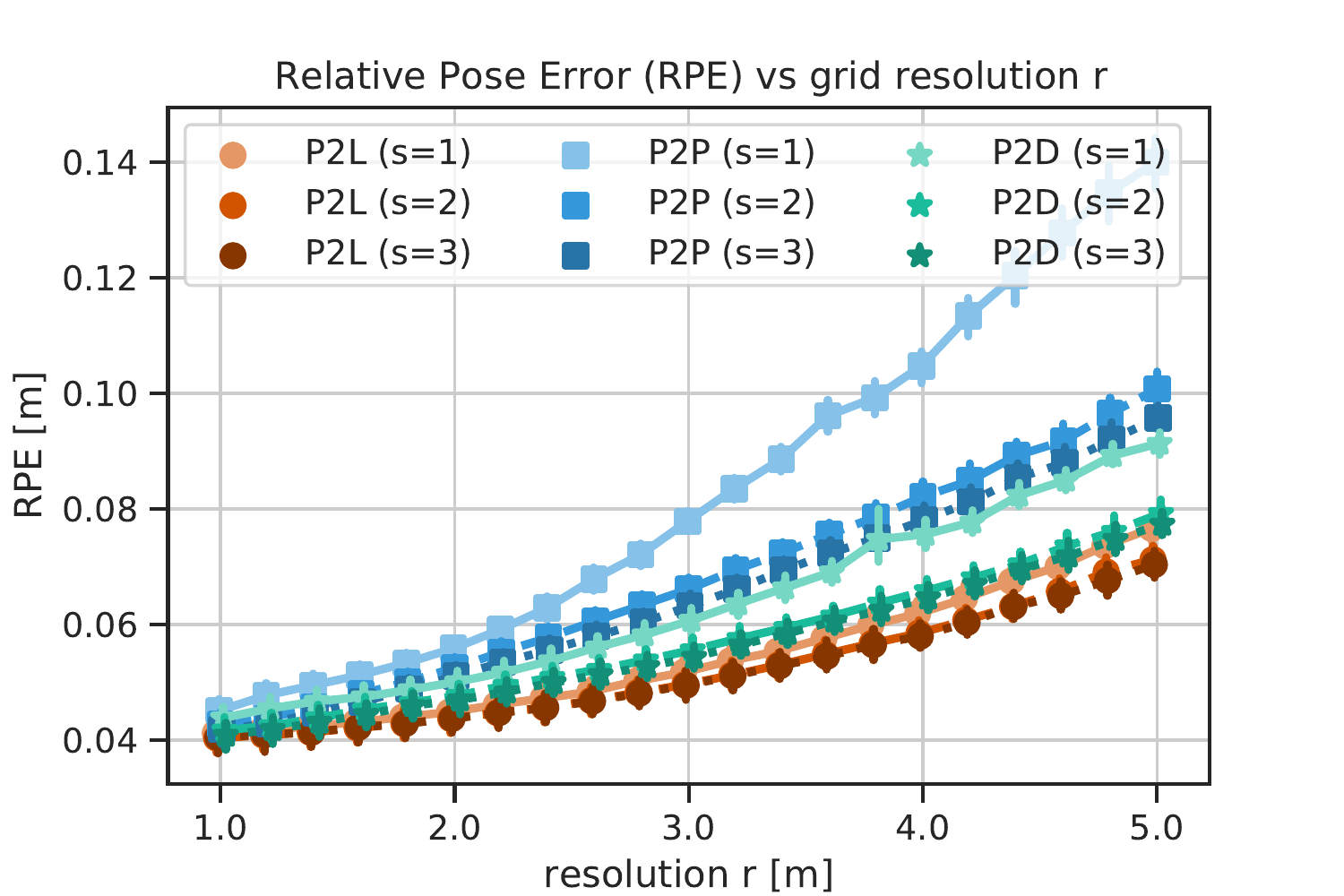}\label{fig:res_RPE}}\hfill
    \subfloat[][Transl. error vs resolution $r$.]{\includegraphics[trim={.3cm 0.1cm 1.3cm 0.1cm},clip,width=.49\hsize]{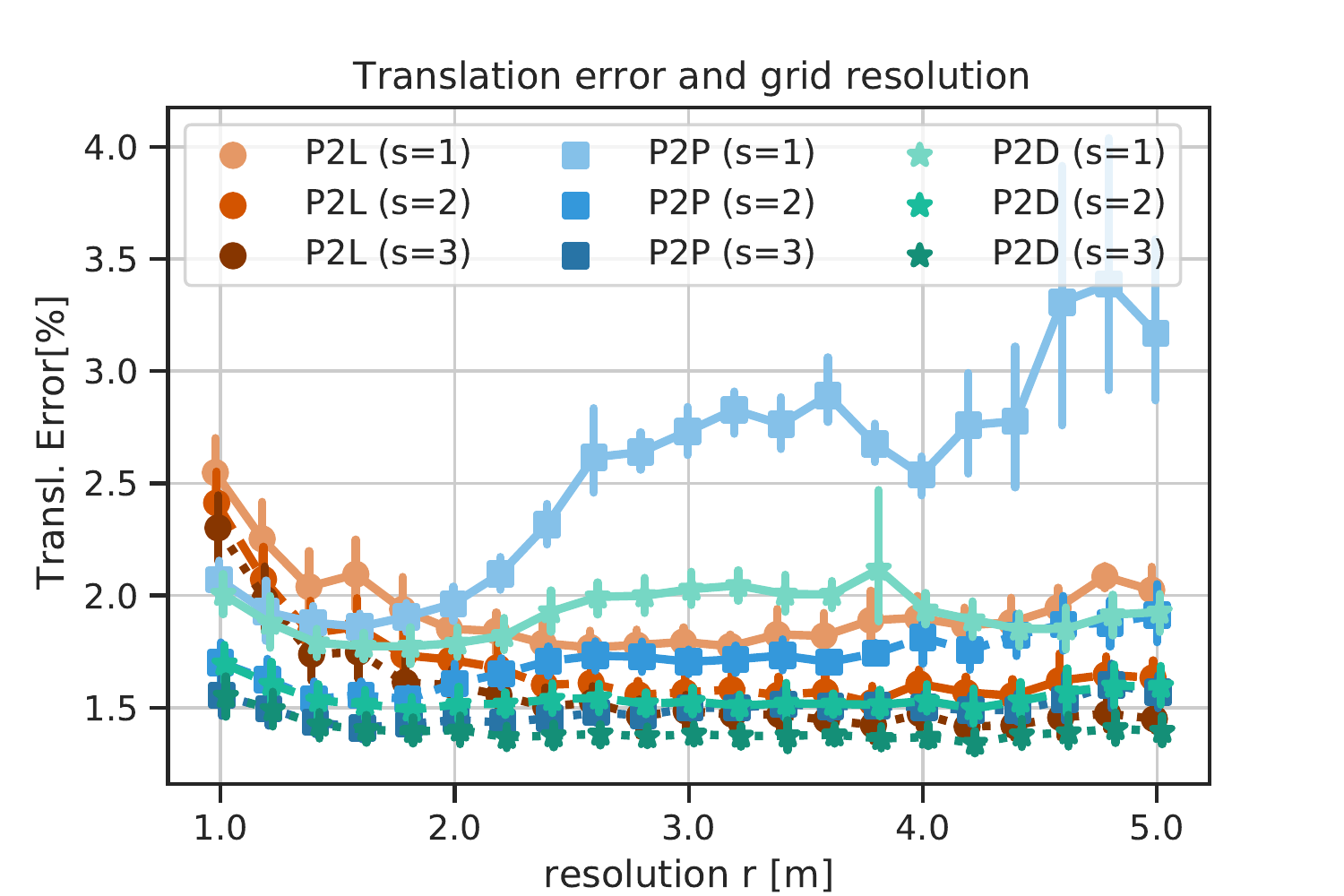}\label{fig:res_transl}}\hfill\\
     \subfloat[][Registration time vs resolution $r$.]{\includegraphics[trim={.3cm 0.1cm 1.3cm 0.1cm},clip,width=.49\hsize]{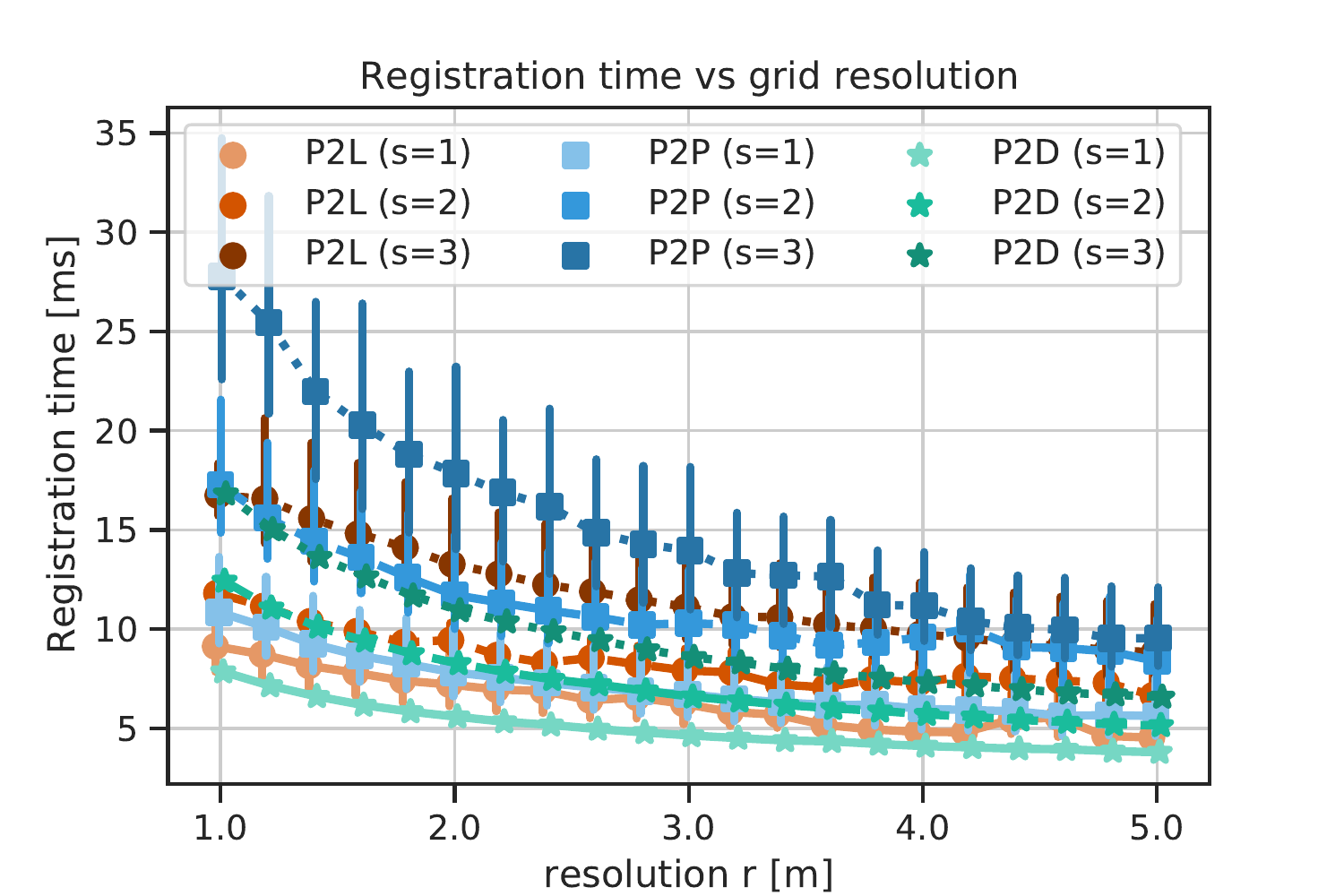}\label{fig:reg_time_res}}\hfill
    \subfloat[][Computational efficiency (Accuracy per time unit) vs resolution $r$. ]{\includegraphics[trim={.1cm 0.1cm 1.2cm 0.1cm},clip,width=.49\hsize]{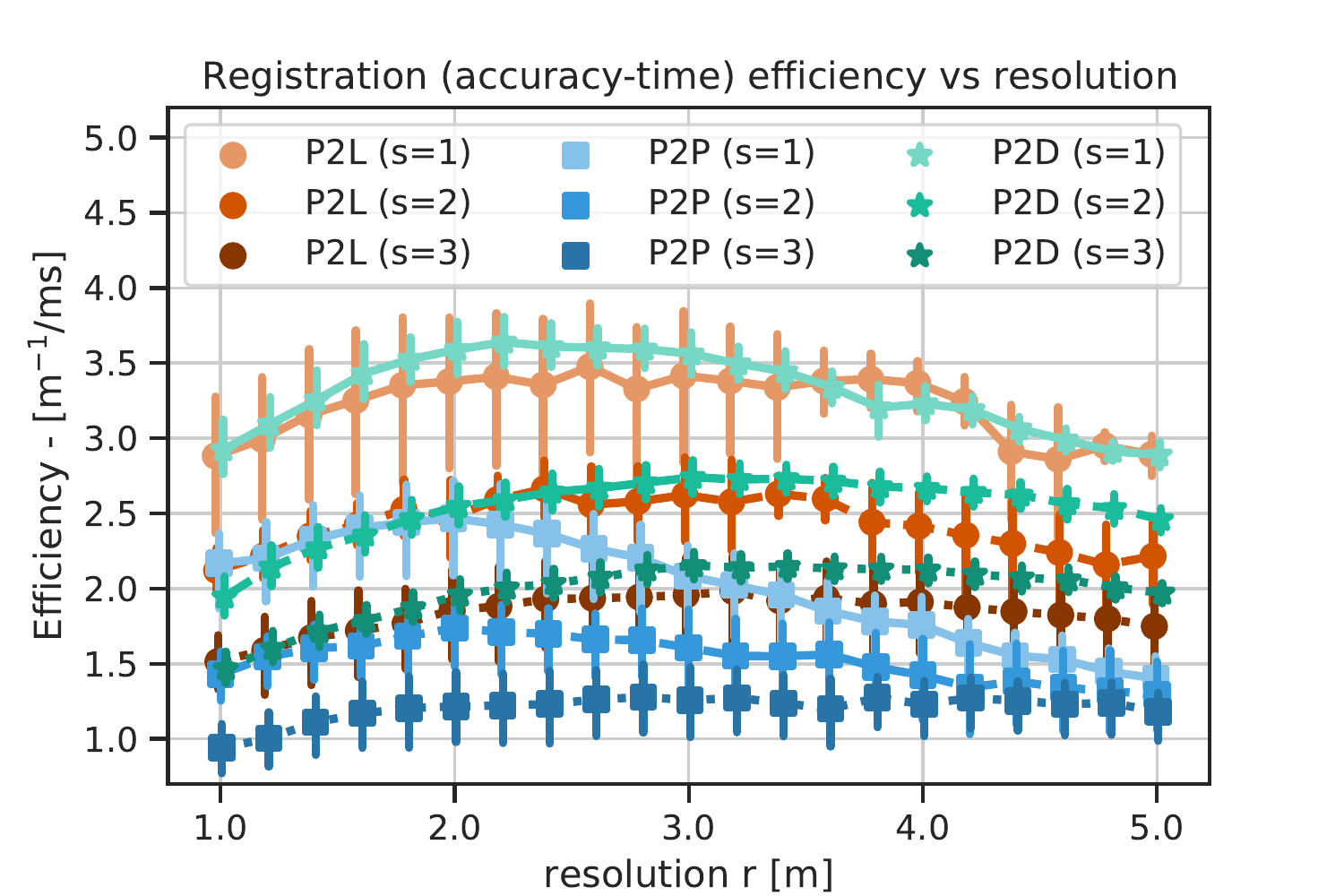}\label{fig:res_efficiency}}
    \caption{Odometry drift~(a), accuracy~(b), computation time~(c) and efficiency~(d) when varying resolution $r$.}\label{fig:res_eval_all}
  \end{center}
  \vspace{-0.3cm}
\end{figure}
\subsubsection{Surface point resolution}

As seen in Fig.~\ref{fig:res_RPE}, a finer grid resolution $r$ reduces RPE for all cost functions, P2L is most accurate in all cases. P2P and P2D are sensitive to a \changedd{coarse} resolution when using a single submap keyframe ($s=1$), however the issue is largely mitigated with additional keyframes. Similarly, additional keyframes substantially improve the odometry robustness which can be seen by the lower mean and variance of the translation drift in Fig.~\ref{fig:res_transl}. While a resolution $\approx3$~m is ideally sized to achieve the lowest translation error in the urban Oxford environment, P2P and P2D effortlessly operate at low drift regardless of resolution. This is interesting as it indicates that the pipeline is expected to achieve low drift in scenes with varied spatial scales without the need for parameter tuning. P2L is somewhat more vulnerable as it relies on more accurate estimates of surface normals.

A coarser resolution produces fewer estimated surface points, and since the registration time largely depends on the number of surface points,
a coarser resolution makes registration faster, as seen in Fig.~\ref{fig:reg_time_res}. Some level of variations is also found between the cost functions.
To additionally understand how to tailor the resolution and cost function to the application, we complement these metrics with the computational efficiency, defined as: RPE obtained per unit of computation time. The computational efficiency is depicted in Fig.~\ref{fig:res_efficiency}. P2L and P2D ($s=1$) achieve the highest efficiency and is therefore suitable when the computational resources are scarce and longitudinal bias is tolerable, e.g. for pose tracking in previously created maps. P2P ($s=3$) is the most computationally expensive method among these.

\begin{figure}[htpb!]
  \begin{center}
  \subfloat[][Translation error with and without motion compensation.]{\includegraphics[trim={0.0cm 0cm 0.0cm 0cm},clip,width=0.48\hsize]{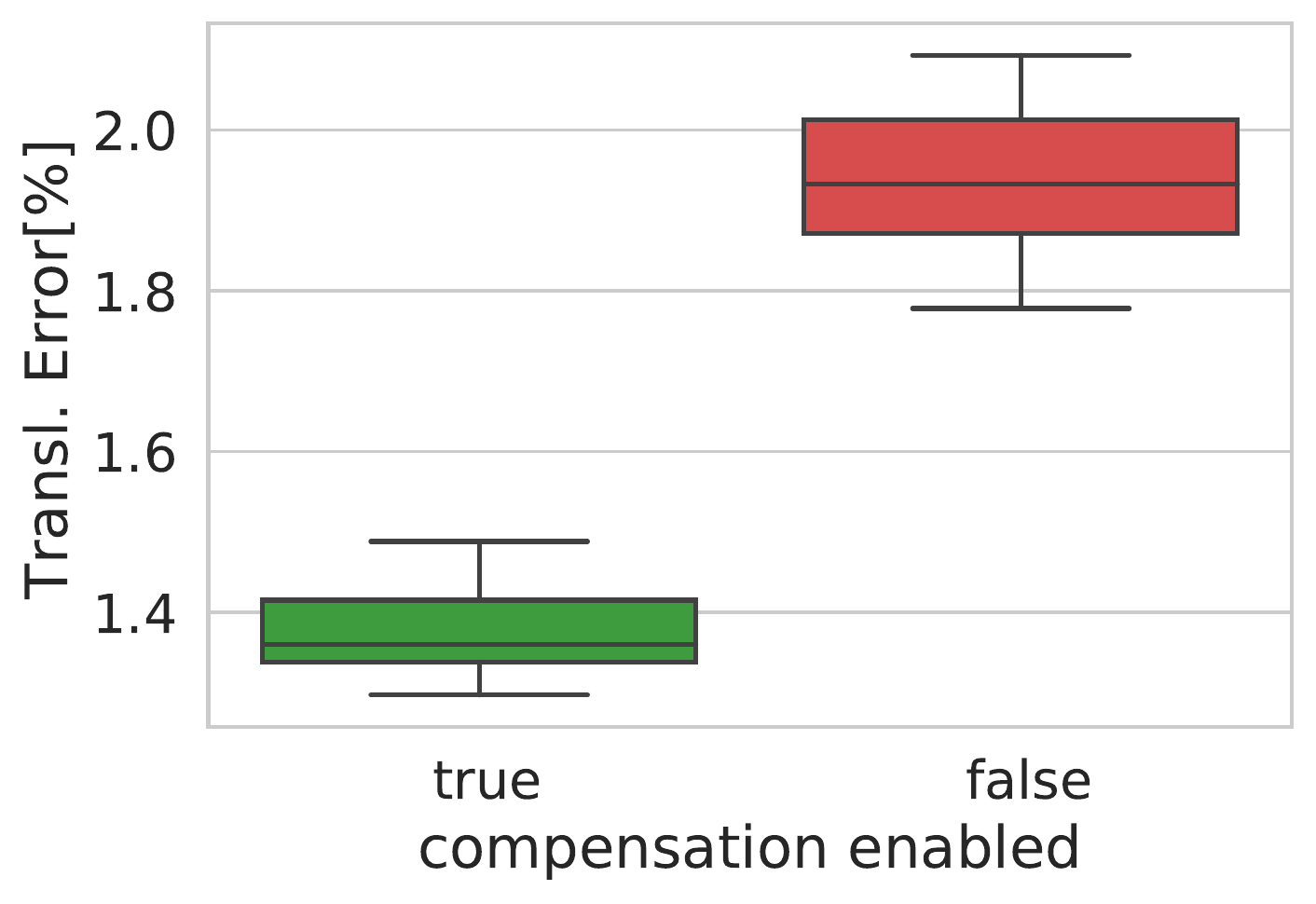}\label{fig:motion_comp_drift}}\hfill
    \subfloat[][RPE with and without motion compensation.]{\includegraphics[trim={0.0cm 0cm 0.0cm 0cm},clip,width=0.48\hsize]{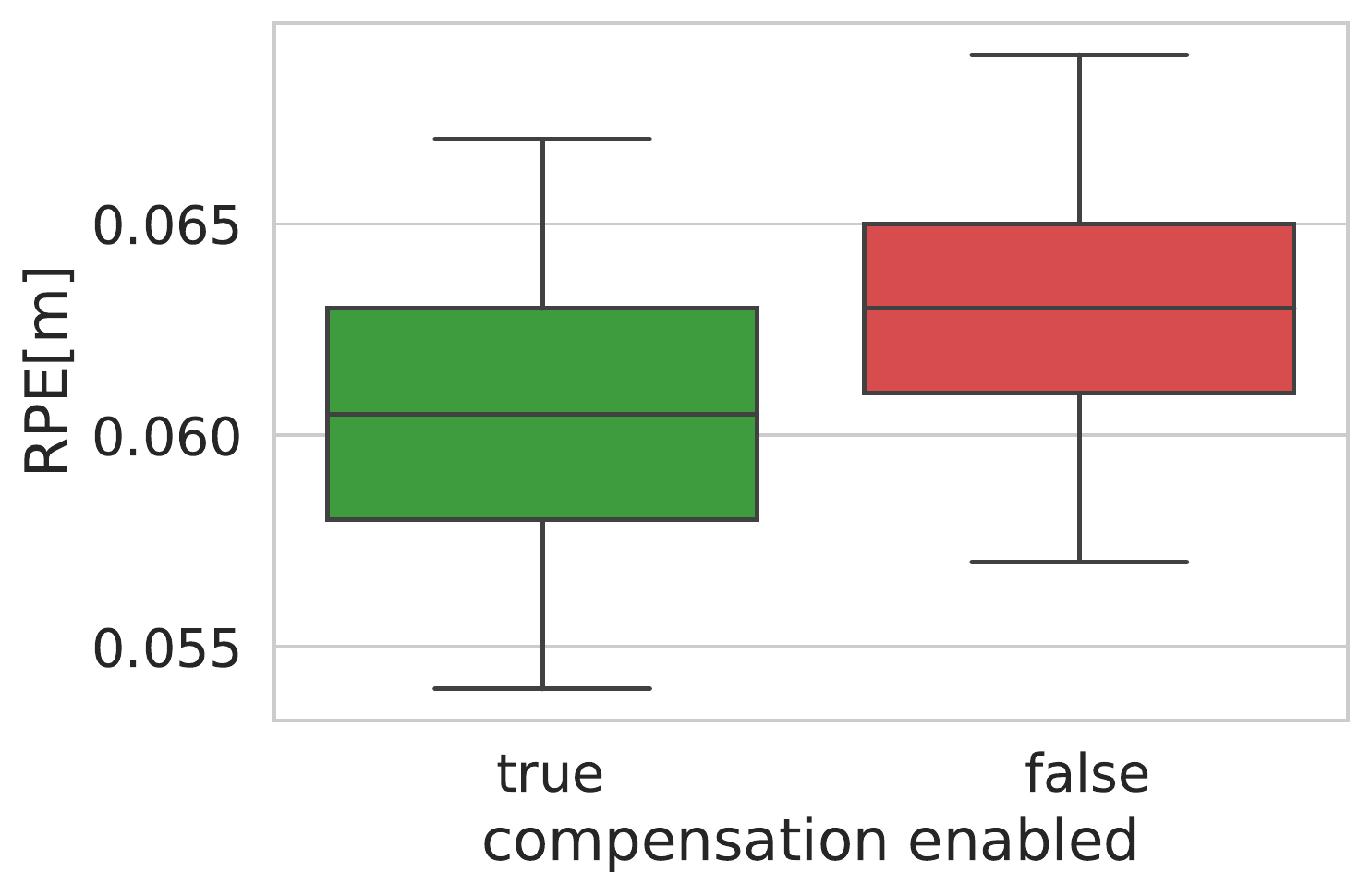}\label{fig:motion_comp_rpe}}\\
     \subfloat[][Translation error vs key frames $s$ with and without compensation.]{\includegraphics[trim={0.0cm 0cm 0.0cm 0cm},clip,width=0.48\hsize]{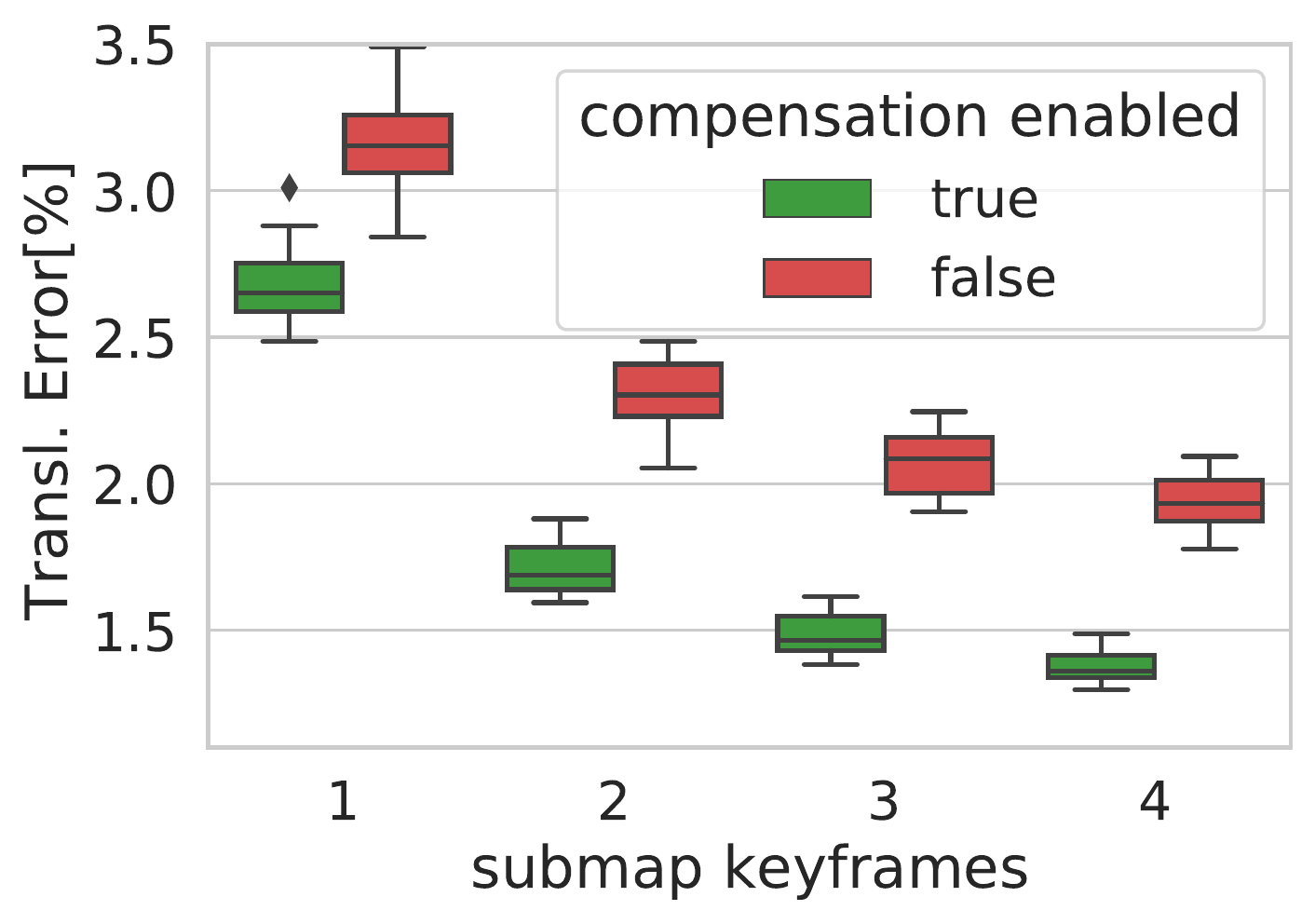}\label{fig:motion_commp_submap_keyframes}}\hfill
    \subfloat[][Translation error and motion compensation per sequence. ]{\includegraphics[trim={0.0cm 0cm 0.0cm 0cm},clip,width=0.48\hsize]{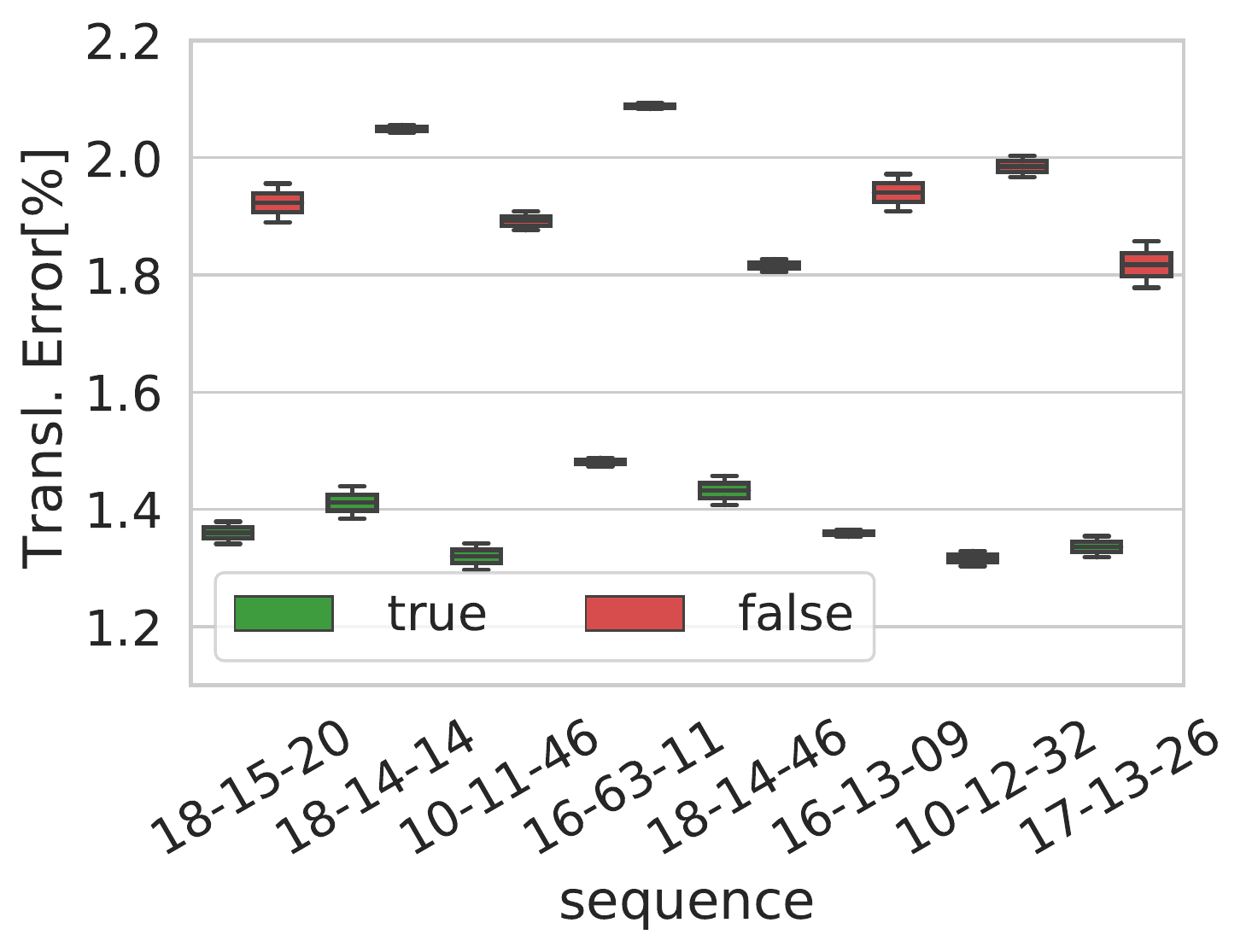}\label{fig:motion_comp_sequence}}\hfill
    \caption{The importance of motion compensation for configuration CFEAR-3. Motion compensation increases robustness which can be observed by the reduced mean and variance of translation error over all evaluated sequences and numbers of submap keyframes. Enabling motion compensation reduces the overall translation error with $29.0\%$. The RPE, however, is only marginally reduced by $3.96\%$. }\label{fig:motion_comp}\hfill
      \end{center}
  \vspace{-0.3cm}
\end{figure}
 
\subsubsection{Motion compensation}
\label{sec:motion_comp_sec}
Following the argument by Burnett et al.~\cite{burnett_we_2021}, that compensating for motion distortion is important for mapping and localization systems, we evaluated the impact on our odometry pipeline.
As seen in Fig.~\ref{fig:motion_comp}(a,b), we found that compensation for motion distortion using the constant velocity model (prior to the registration) dramatically improved odometry with $29\%$ lower error, and lower variance. However, the effect on RPE is limited with a  $3.96\%$ reduced error. In Fig.~\ref{fig:motion_comp_sequence} we break down the drift per sequence to account for their varying difficulty (e.g. different traffic and driving conditions). Motion compensation consistently reduces the mean and variance of the drift, hence making the odometry more accurate and reliable.
%
%
 \begin{figure*}[!t]
 \vspace{-0.5cm}
  \subfloat[][Characteristics of the various loss functions. Squared loss is most sensitive to outliers while Tukey provides the highest level of mitigation.]
  {\includegraphics[trim={0cm 0.0cm 1cm 0.1cm},clip,width=.3\hsize]{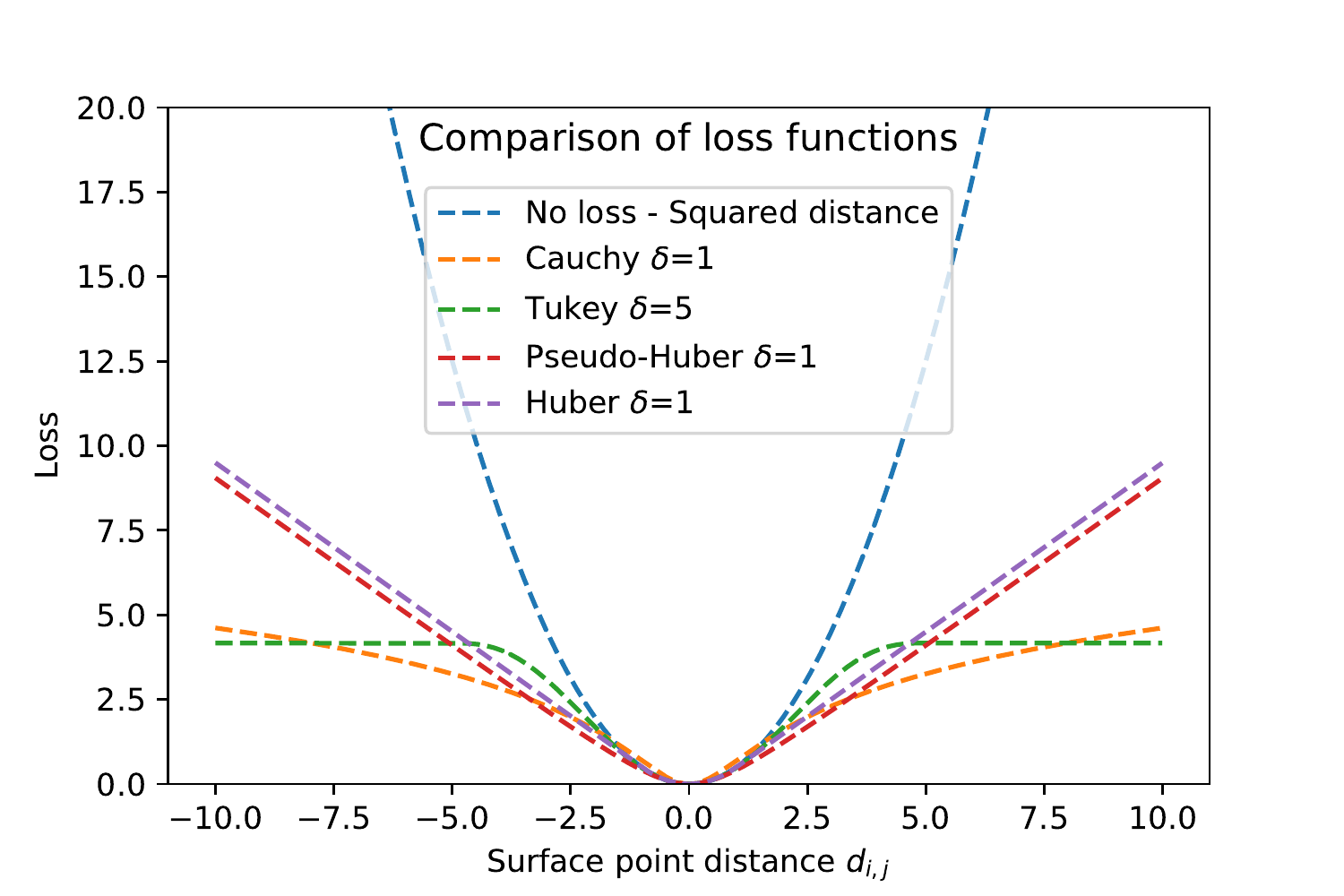}\label{fig:loss_example}}\hfill
 \subfloat[
 RPE w.r.t. loss limit $\delta$. For $\delta=0.1$ the improvement compared to a squared loss is: \textit{Huber}: ($-$32.7\%), \textit{Pseudo Huber}: ($-$31.1\%), \textit{Cauchy}: ($-$39.6\%), \textit{Tukey} ($\delta=1$):($-$32.0\%).
 ]
{\includegraphics[trim={.3cm 0.1cm 1.3cm 0.1cm},clip,width=.3\hsize]{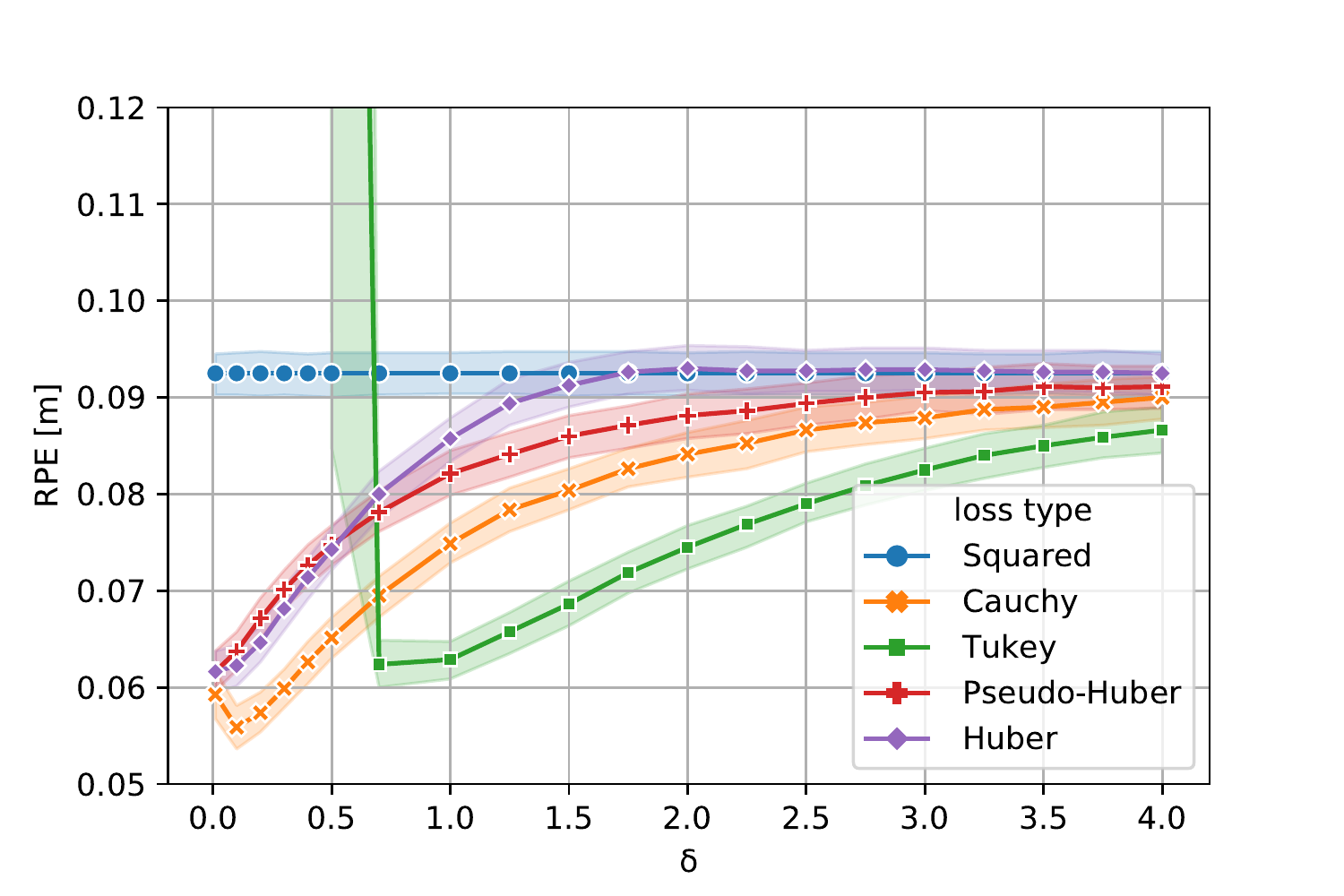}\label{fig:loss_function_rpe}}\hfill
\subfloat[
Drift w.r.t. loss limit $\delta$. For $\delta=0.1$ the improvement compared to a squared loss is: \textit{Huber}: ($-$17.6\%), \textit{Pseudo Huber}: ($-$16.7\%), \textit{Cauchy}: ($-$24.9\%), \textit{Tukey}($\delta=1$):($-$18.5\%)
]
{\includegraphics[trim={.3cm 0.1cm 1.3cm 0.1cm},clip,width=.3\hsize]{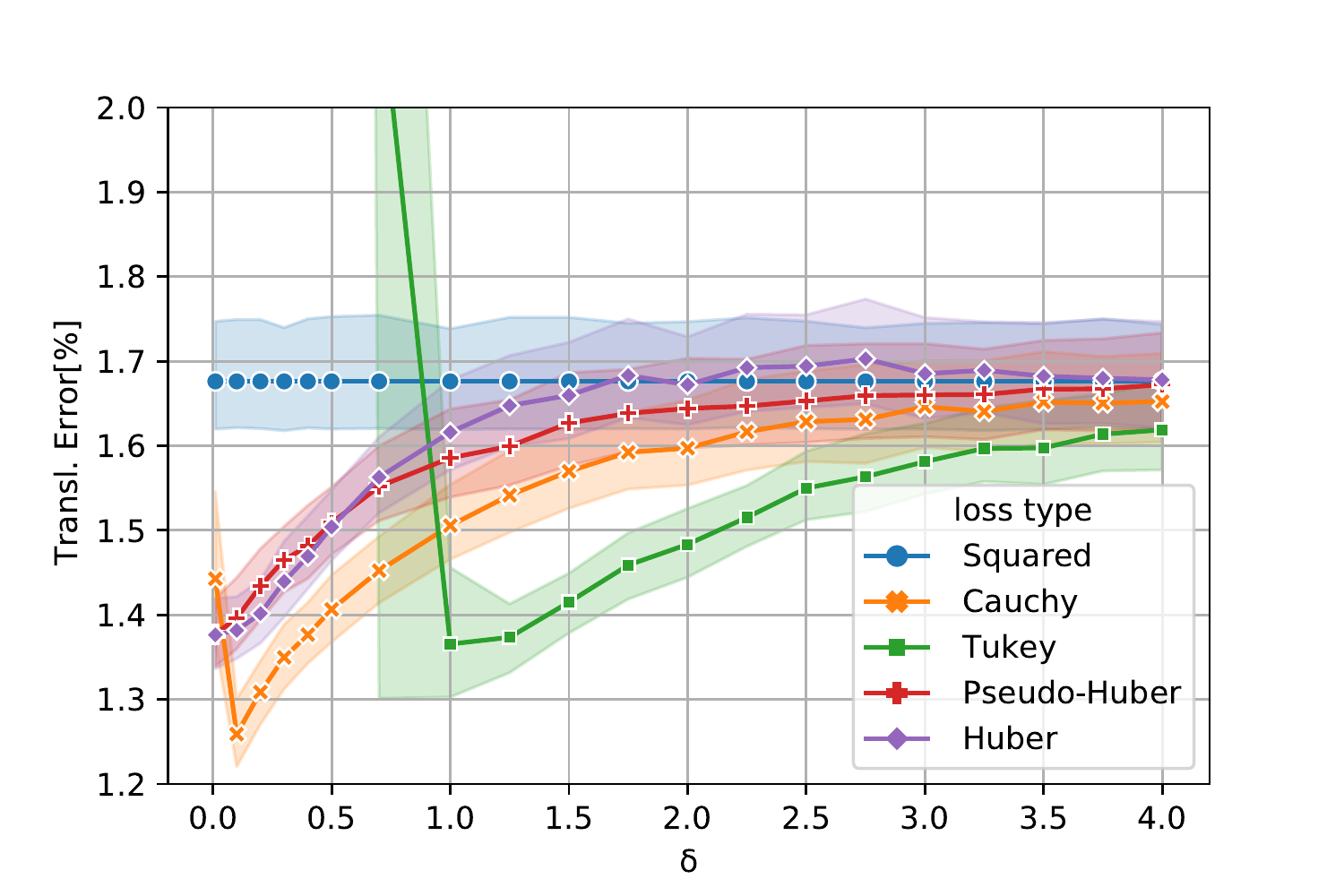}\label{fig:loss_function_odom_error}}\hfill
\caption{Characteristics~(a) and evaluation of RPE~(b) and drift~(c) for the various loss functions.}\label{fig:loss_evaluation}\hfill
\vspace{-0.5cm}
 \end{figure*}
\changed{
Key to our motion compensation is the use of a motion prior, in our case a constant velocity model. If we instead replace the constant velocity model in~\eqref{eqn:solving} with a zero velocity model (no predicted motion), the estimated trajectories are heavily corrupted. Non-zero acceleration, which violates the constant velocity assumption, reduces the correctness of motion compensation, moves the starting point further from the optimum, and decreases odometry performance. However, to understand the significance, we study how acceleration affects the instantaneous pose error ($||transl(\mathbf{E}_i)||$) as depicted in Fig.~\ref{fig:acceleration_and_velocity}(a,c). The instantaneous pose error somewhat increases with acceleration in both datasets. However, the largest errors appear infrequently and do not lead to failures. Instead, low drift is maintained at each of the individual sequences. Additional data acquired at higher acceleration is required to assess the breakdown point within non-artificial conditions. 

We observe that acceleration influence the time efficiency, during the start and stops of the vehicle the predicted starting point for registration is further from the minima, and additional iterations are required until registration convergences. Speed itself had a minor impact as seen in Fig.~\ref{fig:acceleration_and_velocity}(b,d).

 \begin{figure}
 \vspace{-0.5cm}
     \centering
      \subfloat[][Error wrt. acceleration]      {\includegraphics[trim={0 0 0 0},clip,width=.45\hsize]{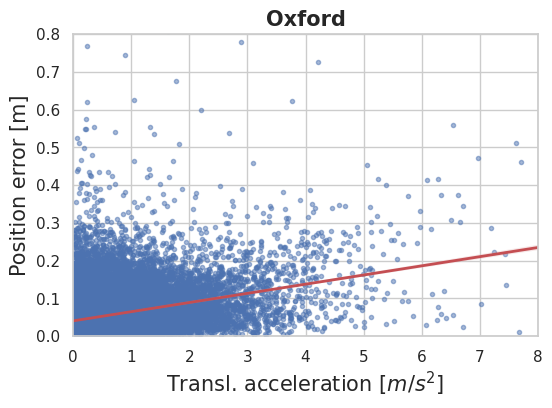}\label{fig:acceleration_oxford}}
       \subfloat[][Position error wrt. speed]
       {\includegraphics[trim={0cm 0cm 0cm 0cm},clip,width=.45\hsize]{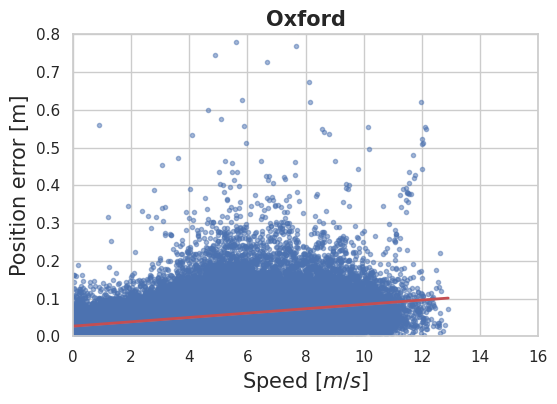}\label{fig:velocity_oxford}}\hfill
       \\
        \subfloat[][Error wrt. acceleration]      {\includegraphics[trim={0 0 0 0},clip,width=.45\hsize]{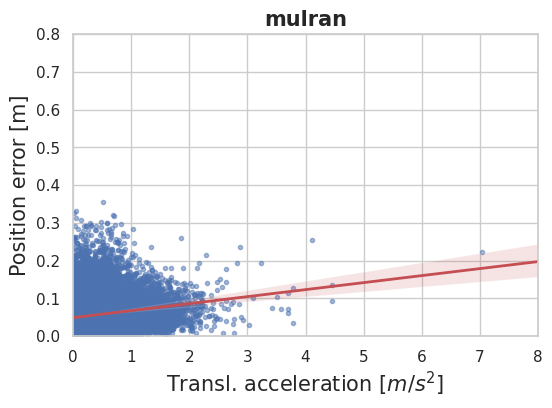}\label{fig:acceleration_MulRan}}
       \subfloat[][Position error wrt. speed]
       {\includegraphics[trim={0cm 0cm 0cm 0cm},clip,width=.45\hsize]{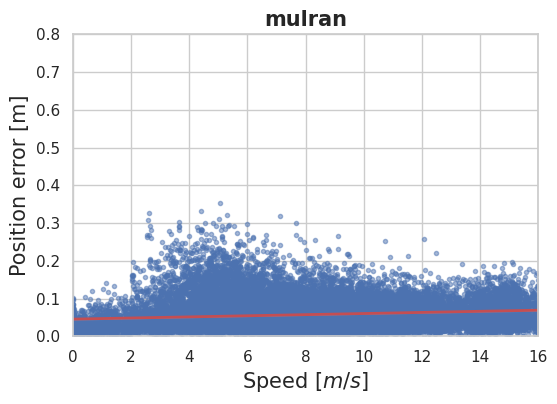}\label{fig:velocity_MulRan}}\hfill
     \caption{Distribution of position error wrt. acceleration (a, c) and speed (b, d), with regression line and $95\%$ confidence interval in red. Acceleration can have a slightly negative influence on motion compensation. Speed has a minor impact.}
     \label{fig:acceleration_and_velocity}
     \vspace{-0.3cm}
 \end{figure}
}
 
\subsubsection{Robust loss function}
When registering the point sets, it is desirable to find reliable surface point correspondences in spite of initial alignment errors. Therefore, the association resolution $r$ needs to be chosen high enough so that true (or close) correspondences are found. However, a large resolution gives rise to outliers that negatively impact pose estimates. For that reason, we have investigated robust loss functions that influence the shape of the cost function to mitigate outliers.
In addition to the Huber loss~\cite{Huber1992} presented in Eq.~\eqref{eq:huber} we investigated the Pseudo Huber~\cite{551699}, Cauchy~\cite{robust_estimation} and Tukey biweight loss function~\cite{ZHANG199759,SINOVA2018219}. Their properties are depicted in Fig.~\ref{fig:loss_example}. 
The Huber and Pseudo-Huber losses reshape the loss from squared to linear outside the boundary $|h|>\delta$, hence avoiding significantly higher impact from large residuals. The Cauchy loss progressively decreases the impact at larger distances, while Tukey's biweight loss function more aggressively suppresses outliers.
\changed{The loss functions are presented in their \textit{objective form} below, each parameterized by $\delta$ that influences their properties.

\begin{align}
&\ln\Big(1 +(\frac{\delta}{h})^2\Big) \tag*{(Cauchy) }\\
&\begin{cases}
    \frac{\delta^2}{6}\big(1-[1-(\frac{h}{\delta})^2 ]^3\big) & \text{if } |h|\leq \delta\\
    \frac{\delta^2}{6} & \text{otherwise.}
\end{cases} 
\eqname{(Tukey's)}\\
    &\delta^2\bigg(\sqrt{1+(\frac{h}{\delta})^2}-1 \bigg)
    \eqname{(Pseudo Huber)}
\end{align}

}

As seen in Fig.~\ref{fig:loss_evaluation}a, all loss functions can be tuned to provide significantly improved RPE and drift compared to a squared loss, and Huber, Pseudo Huber and Cauchy are superior nearly regardless of $\delta$. Cauchy yields the largest improvement (in contrast to Tukey) with up to $24.9\%$ lower drift and $39.6\%$ lower RPE. The Cauchy loss gives higher outlier mitigation compared to the Huber, while (in contrast to Tukey) not fully suppressing outlier influence and hence allowing registration to convergence.

 %
%
\begin{table*}[h!]
\vspace{-0.2cm}
\centering
  \begin{adjustbox}{width=\textwidth}
\begin{tabular}{l|ll|l|llllllll|l|ll}
              & & & & \multicolumn{1}{l}{\textbf{Sequence}}    &  \\
\textbf{Method} & \textbf{Evaluation} & \textbf{Resolution} & \textbf{rate [Hz]}       & 10-12-32 & 16-13-09 & 17-13-26 & 18-14-14 & 18-15-20 & 10-11-46 & 16-11-53 & 18-14-46 & Mean  & \textbf{Mean SCV} & \textbf{Mean Opti.}    \\
\hline \\
Visual Odometry  ~\cite{Churchill2012ExperienceBN} & ~\cite{barnes_masking_2020} & -  & - & -        & -        & -        & -        & -        & -        & -        & -        &    $ 3.980/1.0$ & -  & -         \\
Yoon (Lidar odometry)~\cite{9357964}  & ~\cite{9357964} & - & - & - & $2.56/1.2$ & - & - & $2.41/1.13$  & $2.85/1.29$        & - & - &  $2.65/1.26$  & $3.14/1.53$ & $2.65/1.26$        \\
SuMa (Lidar - SLAM)~\cite{behley2018rss}  & ~\cite{hong2020radarslam }& - & -  & $1.1/0.3^p$ & $1.2$/$0.4^p$ & $1.1/0.3^p$ & $0.9/0.1^p$ & $1.0/0.2^p$  & $1.1/0.3^p$        & $0.9/0.3^p$        & $1.0/0.1^p$        &  $1.03/0.3^p$  & - & -        \\
RadarSLAM-Full~\cite{hong-2022-radarslam} & ~\cite{hong2021radar} &$0.043$ & - & $1.98/0.6$        & $1.48/0.5$        & $1.71/0.5$        & $2.22/0.7$        & $1.77/0.6$        & $1.96/0.7$        & $1.81/0.6$        & $1.68/0.5$        & $1.83/0.6$    & - & -      \\
\hline \\
Cen2018~\cite{8460687}    &   ~\cite{burnett2021radar} & $0.175$& -   & -     & -        & -        & -        & -        & -        & -        & -        & $3.72/0.95$    & - & -       \\
MC-RANSAC~\cite{burnett_we_2021}    &   ~\cite{burnett2021radar} & $0.0438$ & -        & -        & -        & -        & -        & -        & -       & - & -        & $3.31/1.09$    & - & -       \\
CC-means~\cite{what_goes_around_2022}    &   ~\cite{what_goes_around_2022} & - & -   & -     & -        & -        & -        & -        & -        & -        & -        & $2.53/0.82$    & - & -       \\
Robust Keypoints~\cite{barnes_under_2020}    & ~\cite{barnes_under_2020} & $0.346$    & - & -        & -        & -        & -        & -        & -        & -        & -        & $2.05/0.67^*$    & -   & $2.05/0.67$         \\
MByM - Dual Cart~\cite{barnes_masking_2020} & ~\cite{barnes_masking_2020} & $0.2/0.4$ & -       & -  & -        & -        & -        & -        & -        & -        & -        & $1.16/0.3^*$    &   $2.784/0.85 $ & $1.16/0.3$ \\
RadarSLAM-odometry~\cite{hong-2022-radarslam} & ~\cite{hong2021radar} &$0.043$ & - &  $2.32/0.7$        & $2.62/0.7$        & $ 2.27/0.6$        & $2.29/0.7$        & $2.25/0.7$        & $2.16/0.6$        & $2.49/0.7$        & $2.12/0.6$        & $2.32/0.7$    & - & -     \\
Hero~\cite{burnett2021radar} &~\cite{burnett2021radar} & $0.2628$ & - & 1.77/0.62          & 1.75/0.59        & 2.04/0.73        & 1.83/0.61        & 2.20/0.77        & 2.14/0.71        & 2.01/0.61        & 1.97/0.65        & 
1.96/0.66    &  - & -\\
Kung~\cite{kung2021normal} &~\cite{kung2021normal} & $0.125$ & - & -          & -         & -        & -         & 2.20/0.77        & -         & -         & -         & 
1.9584/0.6  &  -  &  -\\
F - MByM~\cite{fmbm} & ~\cite{fmbm} & - & - & -          & -         & -        & -         & -        & -         & -         & -         & 
$2.06/0.63^*$  &  -  &  $2.06/0.63^*$\\
CFEAR-1 (ours) & & $0.0438$  & 160.2 & 1.59/0.57 & 1.84/0.64 & 1.84/0.64 & 1.83/0.60 & 1.71/0.59 & 1.74/0.57 & 2.11/0.63 & 1.69/0.54 &  1.79/0.60 &  1.69/0.60 & 1.63/0.57\\
CFEAR-2 (ours) &  & $0.0438$ & 111.8 & 1.35/0.49 & 1.50/0.51 & 1.52/0.54 & 1.52/0.52 & 1.41/0.50 & 1.33/0.48 & 1.61/0.53 & 1.48/0.50 & 1.46/0.51 & 1.40/0.49 & 1.40/0.49 \\
CFEAR-3 (ours) &  & $0.0438$ & 44.4 & 1.23/0.36 & 1.25/0.39 & 1.25/0.40 & 1.34/0.41 & 1.26/0.41 & 1.26/0.39 & 1.42/0.39 & 1.42/0.44 & 1.31/0.40 &  1.22/0.39 & 1.20/0.39 \\
CFEAR-3-s50 (ours) &  & $0.0438$ & 6.0 & \textbf{1.05/0.34} & \textbf{1.08/0.34} & \textbf{1.07/0.36} & \textbf{1.11/0.37} & \textbf{1.03/0.37} & \textbf{1.05/0.36} & \textbf{1.18/0.36} & \textbf{1.11/0.36} &  \textbf{1.09/0.36} &  - & -  \\

\end{tabular}
\end{adjustbox}

\caption{
\changed{Drift evaluated over 8 sequences from the Oxford Radar RobotCar dataset~\cite{RadarRobotCarDatasetICRA2020}. We compare various methods for visual, lidar and radar odometry, and lidar/radar SLAM that additionally correct the trajectory. Results are given in (\% translation error / deg/$100$~m). 
In the column ``Mean'' we report the best available result are reported, except for CFEAR for which we evaluate with the parameters in Tab.~\ref{tab:Parameter}, optimized for speed and drift. Results obtained via supervised learning within the Oxford environment are marked with $^*$ and copied to ``Mean opti.'' (Optimized) as these cannot be directly compared. Instead, we refer to the ``Mean SCV'' column that contain SCV results. 
Additionally, we provide a SCV for CFEAR (see Sec.~\ref{sec:scv_eval}), reporting results achieved when optimizing parameters for drift only within another environment (``Mean SCV''), parameters optimized for Oxford are reported in (``Mean Opti.'').
Results marked with $^p$ indicate that the numbers are reported from part of the trajectories up to a failure.}
}\label{tab:OxfordTable}
\vspace{-0.9cm}
\end{table*}
\changed{
\subsubsection{Weighted sample covariance and residuals }
In the final part of our ablation study we have investigated: (i): how intensity weighted radar detections can be used to compute more consistent surface points for reduced drift, and (ii): how surface point similarity can be used to measure correspondence quality for weighting residuals.
We start with uniform weights in both cases, and then add intensity weighted surface points (Sec.~\ref{sec:surface_point}). Then, we add one of the four types of residual weighing schemes individually (proposed in Sec.~\ref{sec:registration}). The results are shown in Fig.~\ref{fig:weights}. We found that weighted surface points improve odometry and pose accuracy in all cases, which is expected as the technique improves the location consistency of surface points. The effect is particularly large for CFEAR-3, which is expected as P2P is more reliant on consistently located surface points.

Weighting residuals had a relatively smaller, yet noticeable effect on both drift and pose accuracy. Some level of variation was found among each weighting scheme. The improvement was largest for CFEAR-1, with a  similarly small yet clear effect for CFEAR-(2\&3) that use additional keyframes. We hypothesize that the weights serve a similar purpose to keyframes; to reduce the influence of low-quality correspondences. CFEAR-1 uses fewer correspondences and is hence more likely to benefit from weighting.
 \begin{figure}
     \centering
      \subfloat[][Transl. error vs weighting types.]{\includegraphics[trim={0 0 0.5cm 0},clip,width=0.49\linewidth]{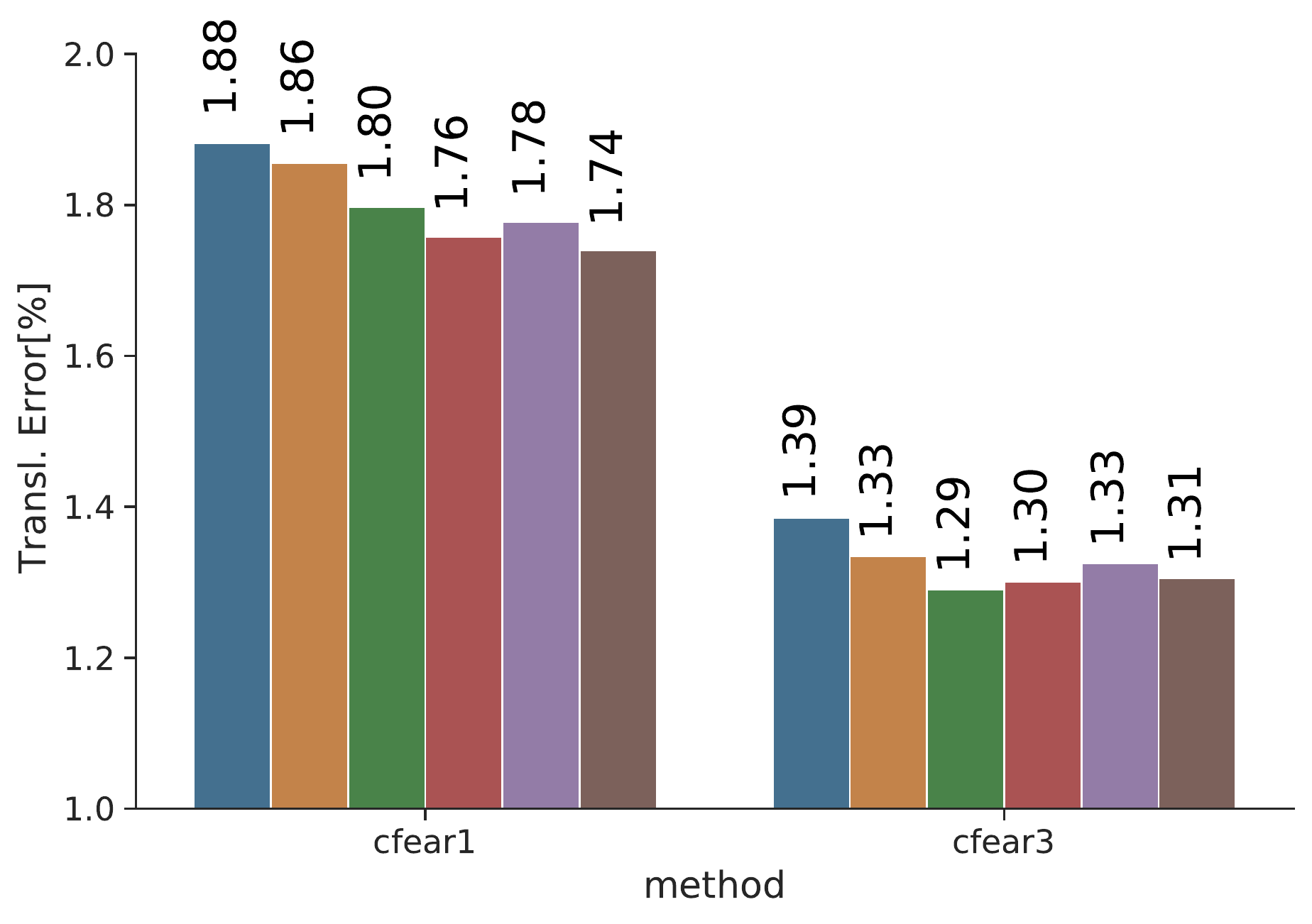}\label{fig:weights_transl}}\hfill
       \subfloat[][RPE vs weighting types.]
       {\includegraphics[trim={0cm 0cm 0.5cm 0cm},clip,width=0.49\linewidth]{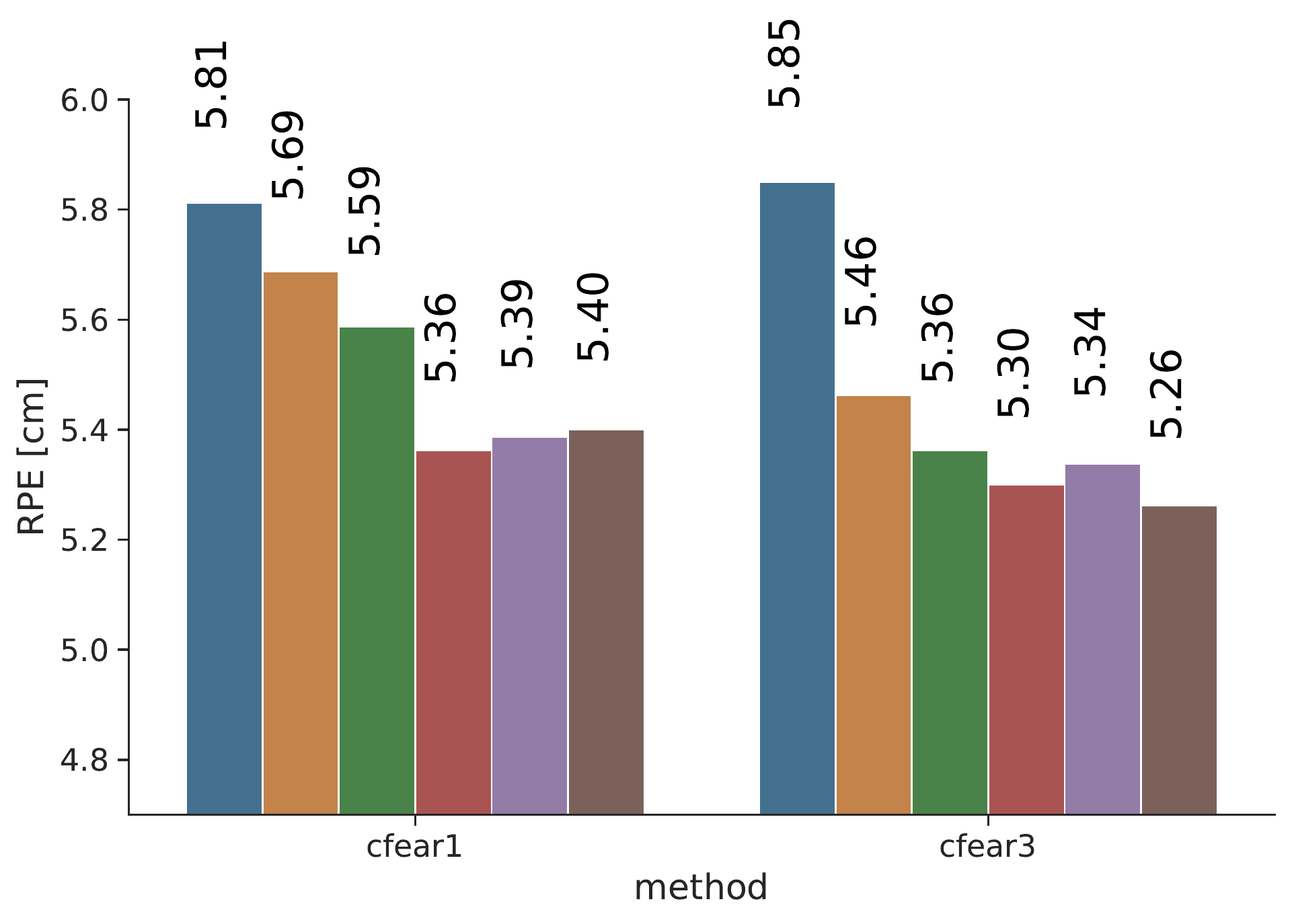}\label{fig:weights_rpe}}\\
       
       \vspace{-1.5cm}
       {\hspace{7cm}\includegraphics[trim={0 5cm 0 0.5cm},clip,width=0.85\linewidth]{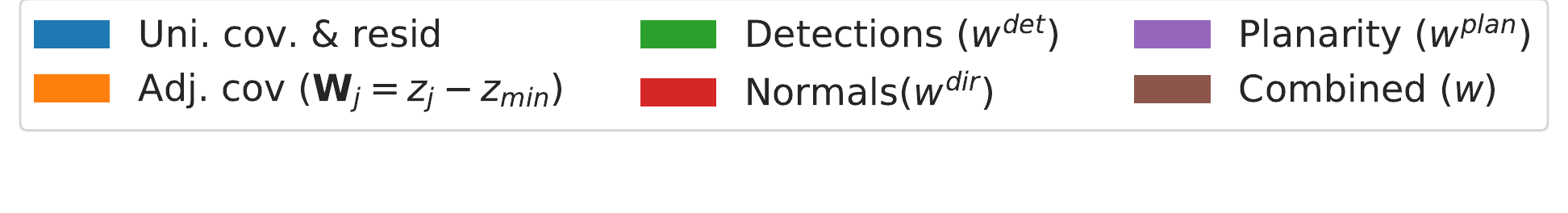}}
       \captionsetup[subfigure]{labelformat=empty}
       \vspace{0.8cm}
          \caption{\changed{
          The effect of weighting on accuracy for CFEAR-(1-3). Blue: uniform weights. Orange: Adding intensity-weighted covariance. Green--purple: Adding individual weighting schemes for residuals. Brown: combining all weights for residuals.
          }}
     \label{fig:weights}
     \vspace{-0.5cm}
 \end{figure}
 }


\begin{figure*}[h!]
	\centering
	\newcommand\figsize{0.24\hsize}
	\begin{center}
		\subfloat[10-12-32]{\includegraphics[trim={0.0cm 0cm 0cm 0cm},clip,width=\figsize]{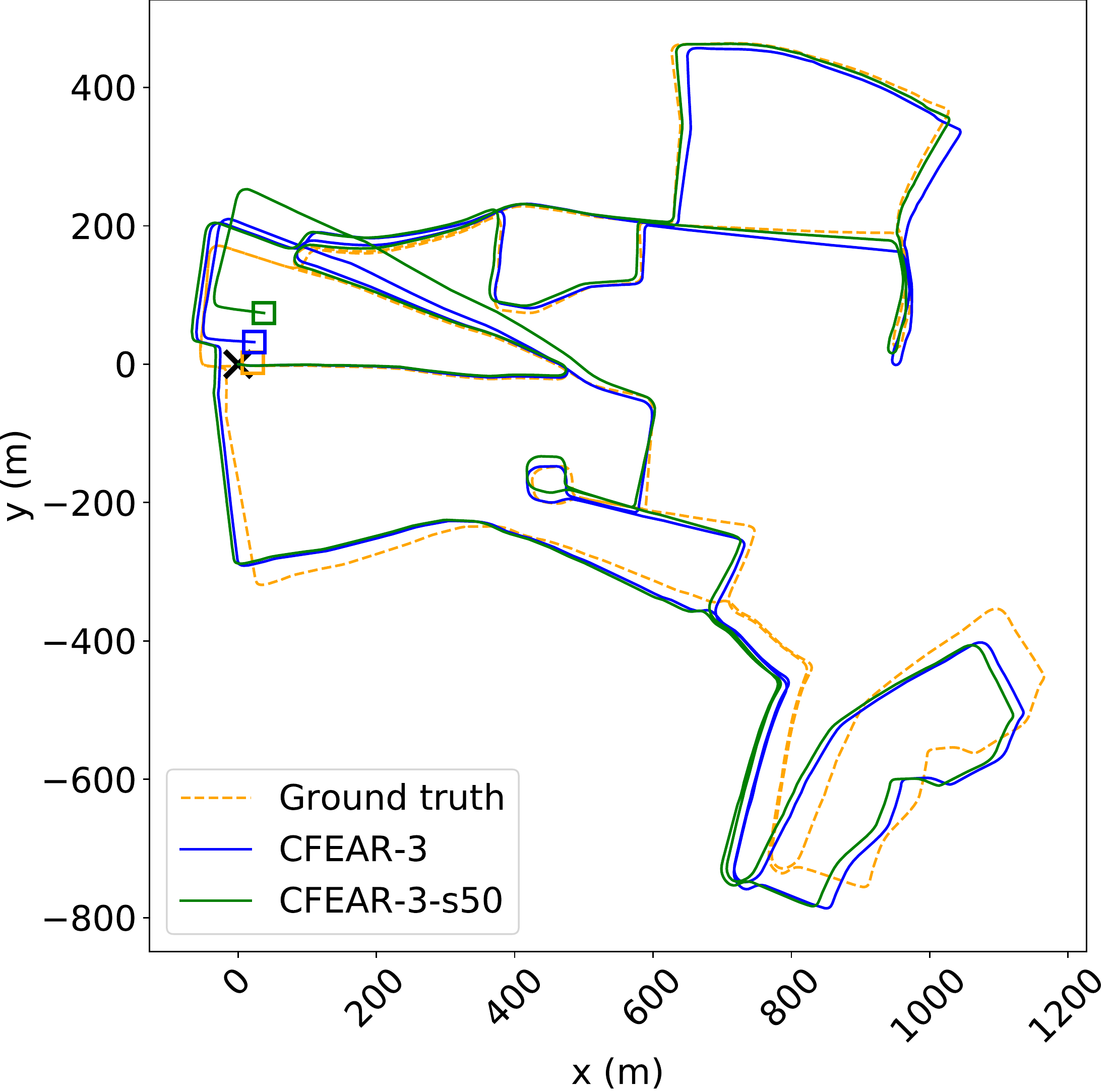}\label{fig:10-12-32}}\hfill
		\subfloat[16-13-09]{\includegraphics[trim={0.0cm 0cm 0cm 0cm},clip,width=\figsize]{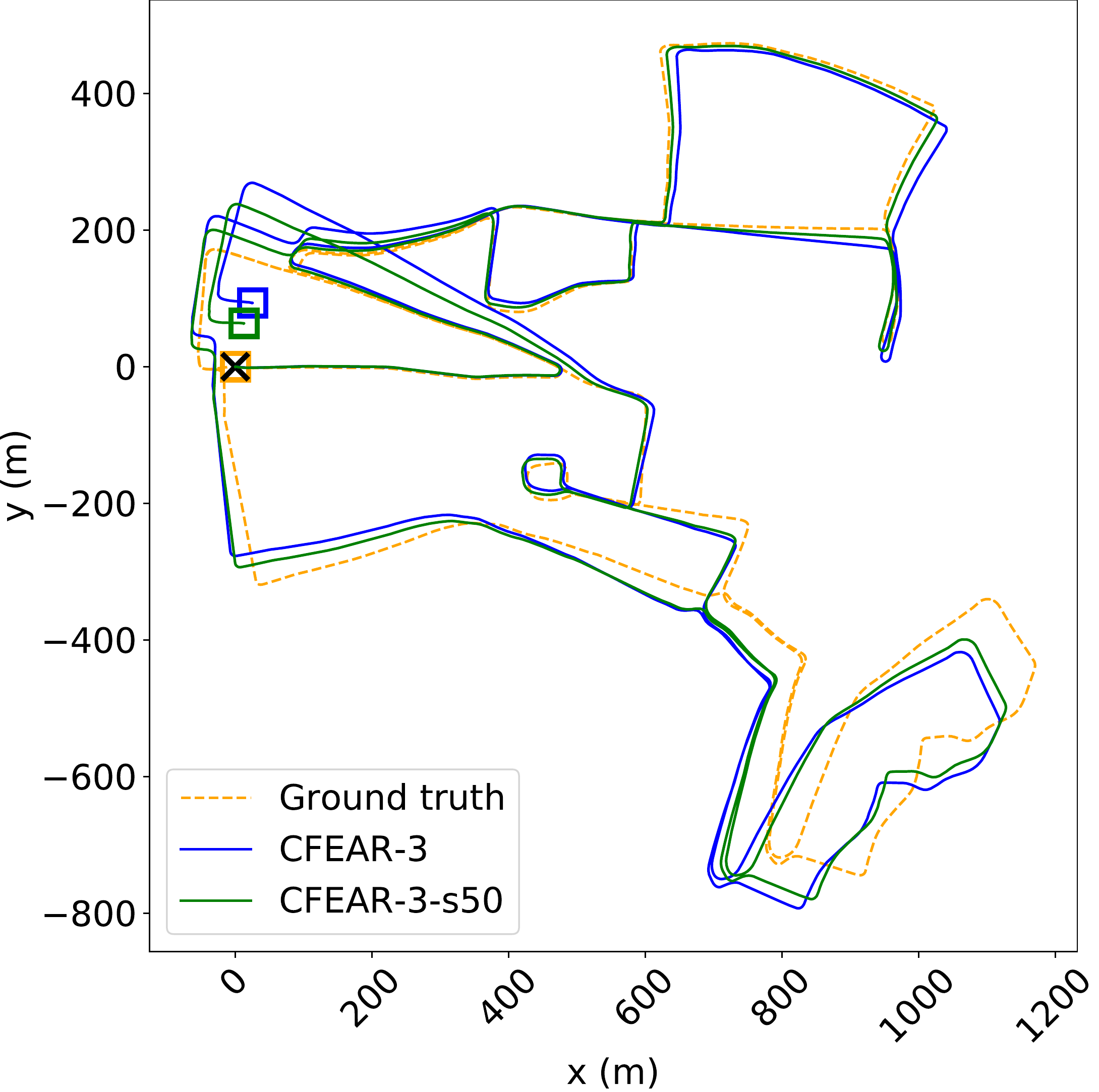}\label{fig:16-13-09}}\hfill
		\subfloat[17-13-26]{\includegraphics[trim={0.0cm 0cm 0cm 0cm},clip,width=\figsize]{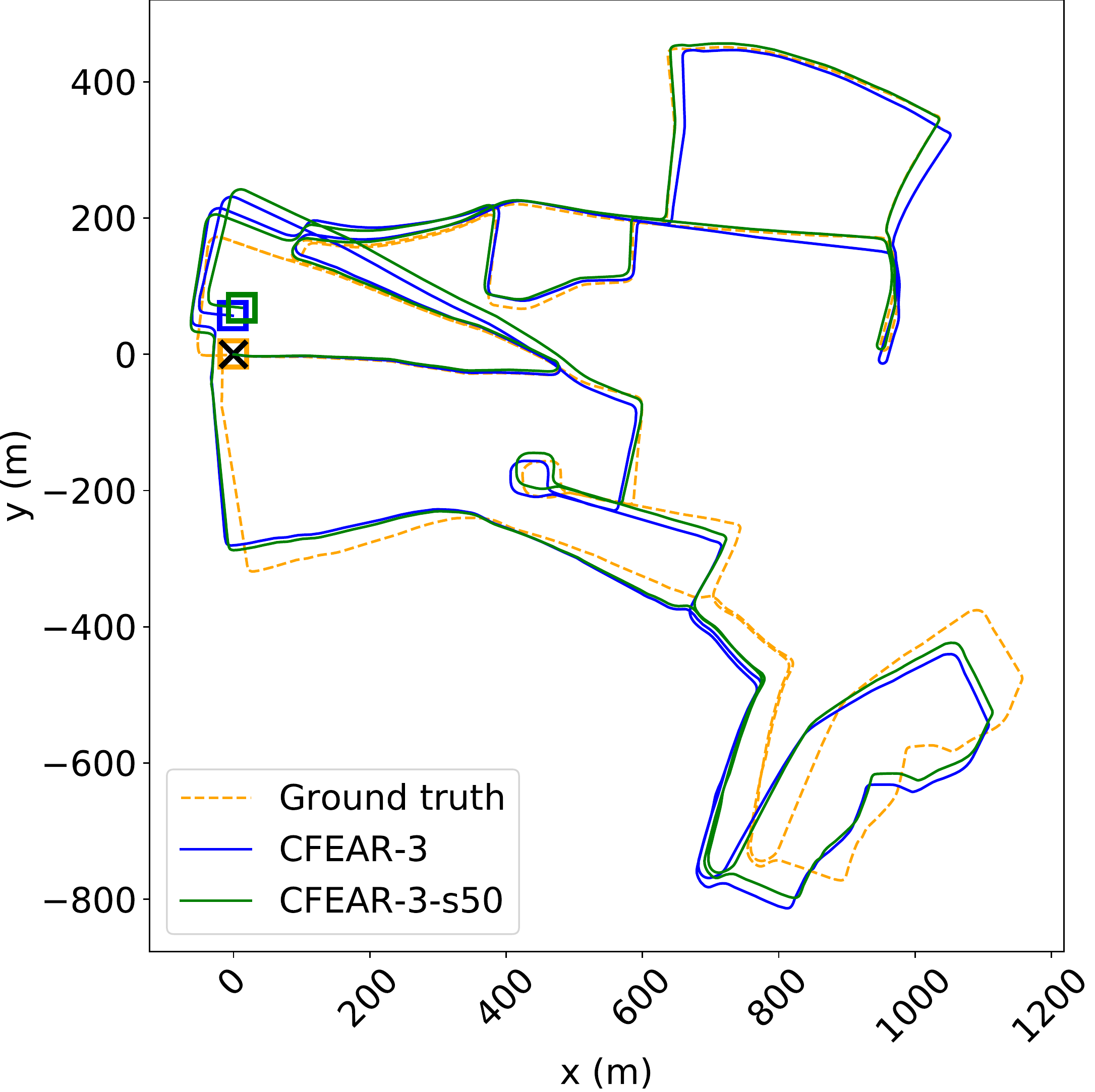}\label{fig:17-13-26}}\hfill
		\subfloat[10-11-46]{\includegraphics[trim={0.0cm 0cm 0cm 0cm},clip,width=\figsize]{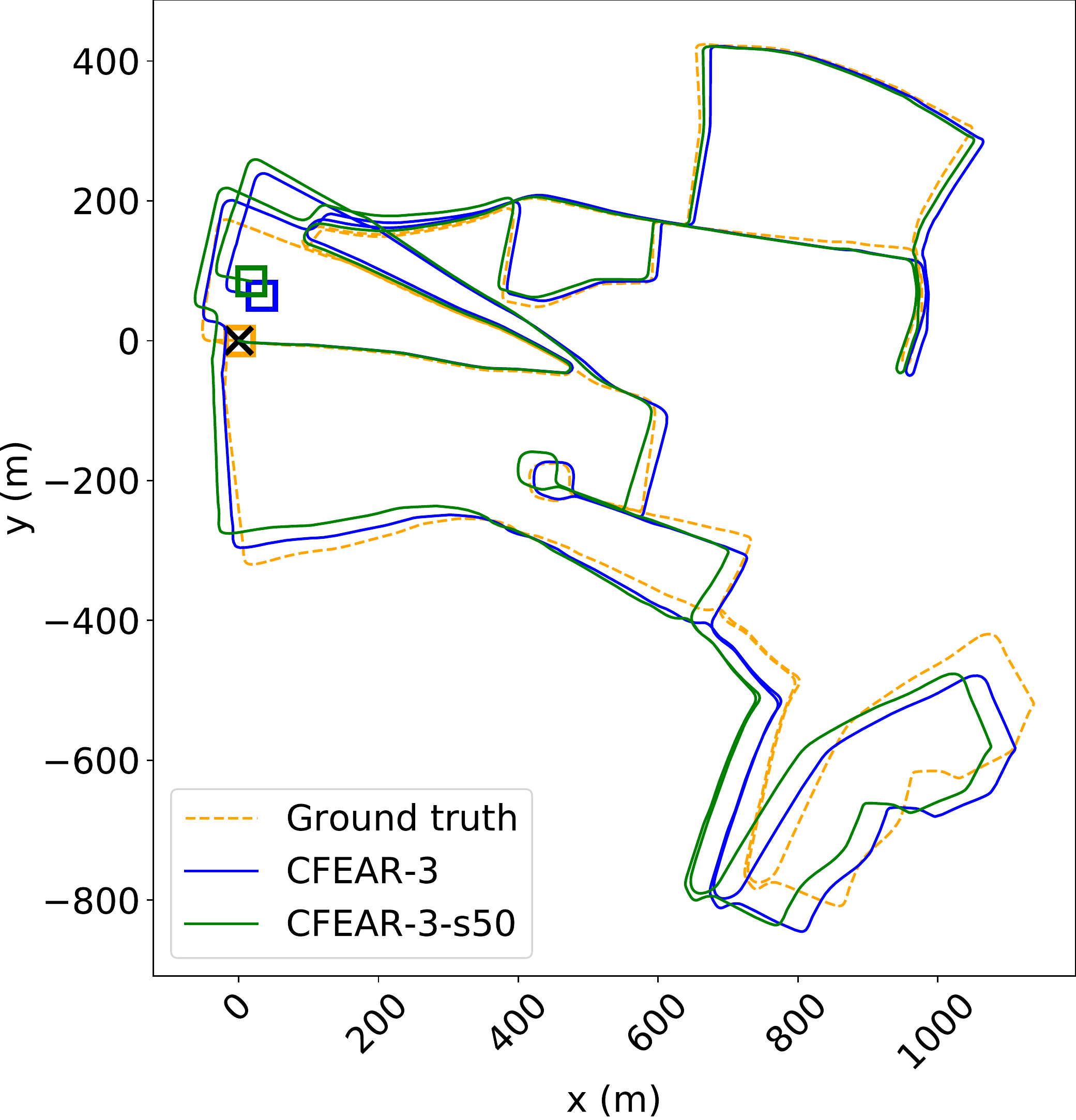}\label{fig:110-11-46}}\vspace{-0.3cm}\\
		\subfloat[18-14-14]{\includegraphics[trim={0.0cm 0.0cm 0.0cm 0cm},clip,width=\figsize]{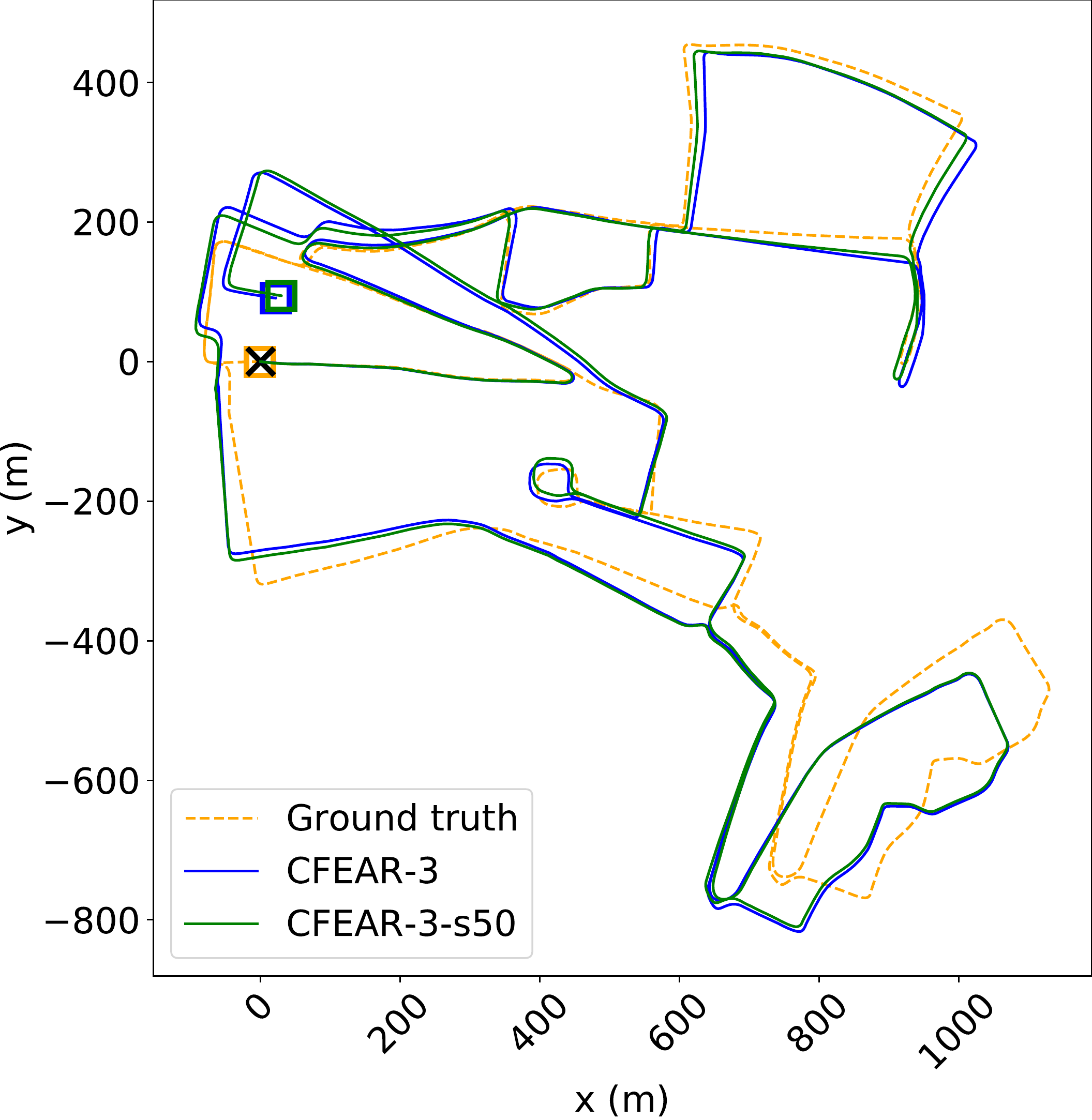}\label{fig:18-14-14}}\hfill
		\subfloat[18-15-20]{\includegraphics[trim={0.0cm 0cm 0cm 0cm},clip,width=\figsize]{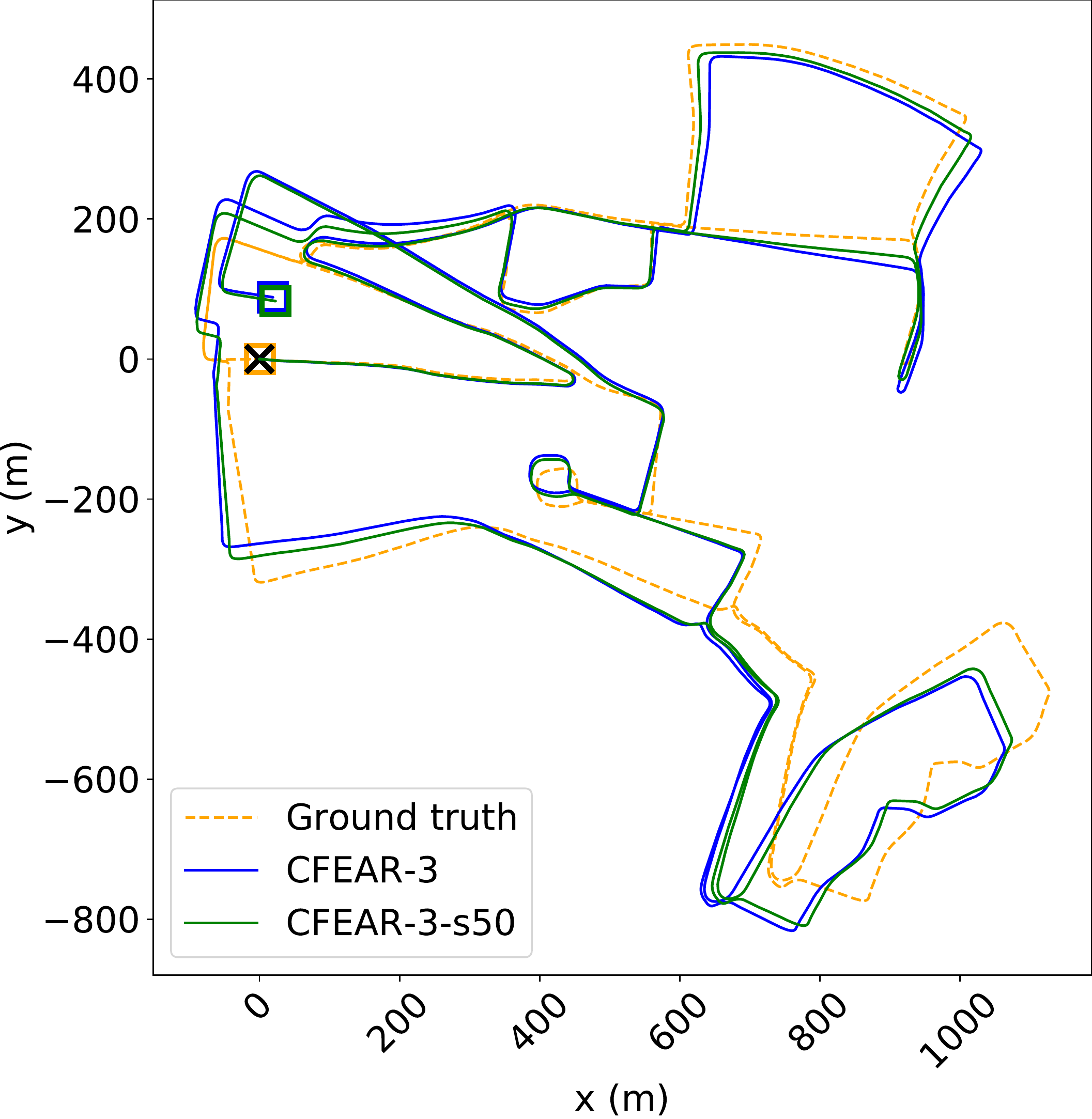}\label{fig:18-15-20}}\hfill
		\subfloat[16-11-53]{\includegraphics[trim={0.0cm 0cm 0cm 0cm},clip,width=\figsize]{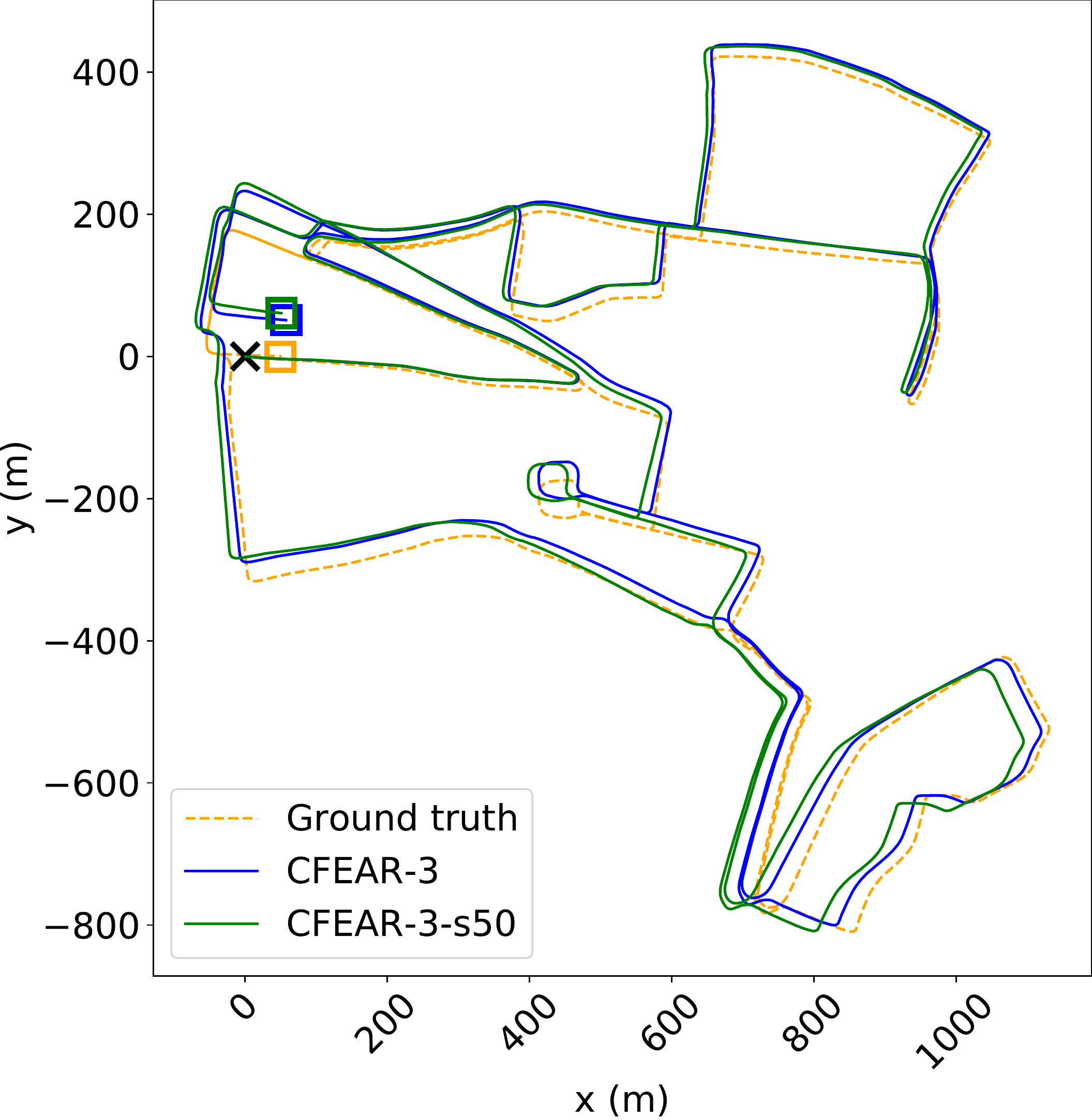}\label{fig:16-11-53}}
		\subfloat[18-14-46]{\includegraphics[trim={0.0cm 0cm 0cm 0cm},clip,width=\figsize]{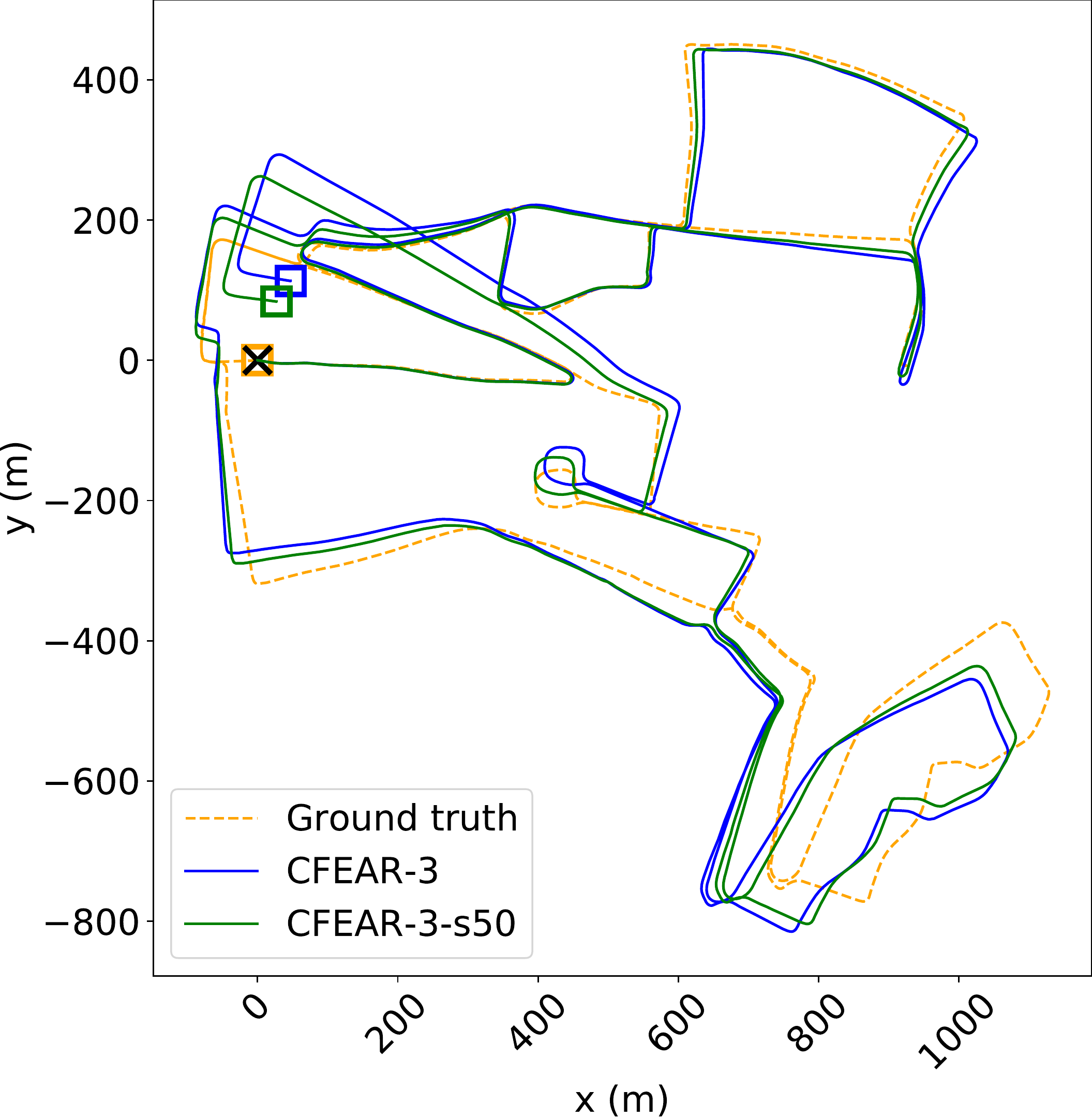}\label{fig:18-14-46}}
		\caption{\changed{Radar odometry estimation for the Oxford sequences obtained with CFEAR-3 (blue) and CFEAR-3-s50 (green) compared to ground truth (orange). Trajectories are aligned by their initial pose, marked with $\mathbf{\times}$. The final pose is marked with $\square$. The same sequences are evaluated in~\cite{hong2020radarslam,hong2021radar,adolfsson2021cfear} and can be compared. The odometry quality is consistently high and trajectory errors are systematic.} \label{fig:sequences_oxford}}
	\end{center}
	\vspace{-0.5cm}
\end{figure*}

\subsection{Comparative evaluation within the Oxford dataset}

 \begin{figure}
  \begin{center}
    \subfloat[Translation error vs path length.]{\includegraphics[trim={0.0cm 0cm 0cm 0cm},clip,width=0.49\hsize]{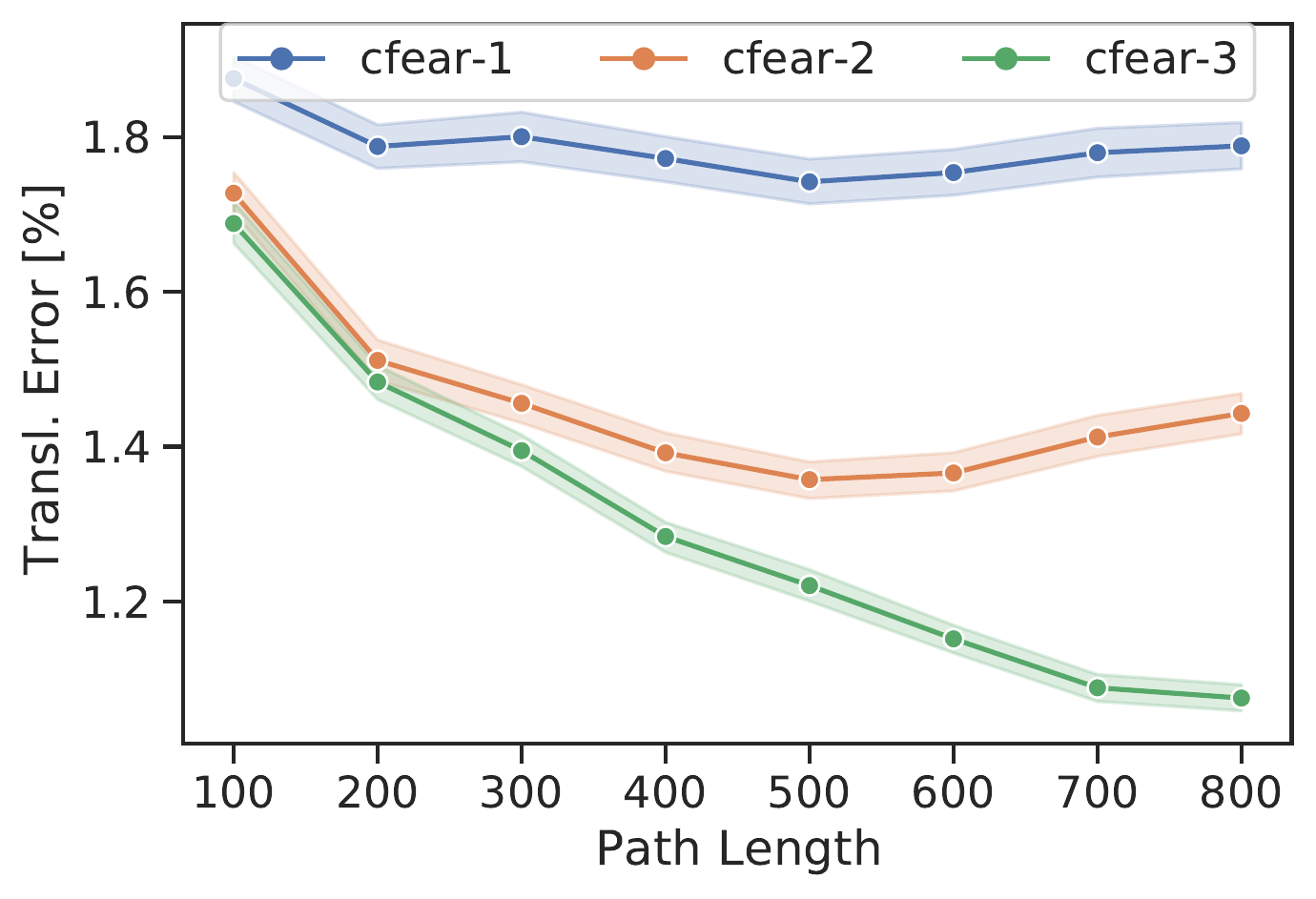}\label{fig:oxford_error_trans}}
    \subfloat[Rotation error vs path length.]{\includegraphics[trim={0.0cm 0cm 0cm 0cm},clip,width=0.49\hsize]{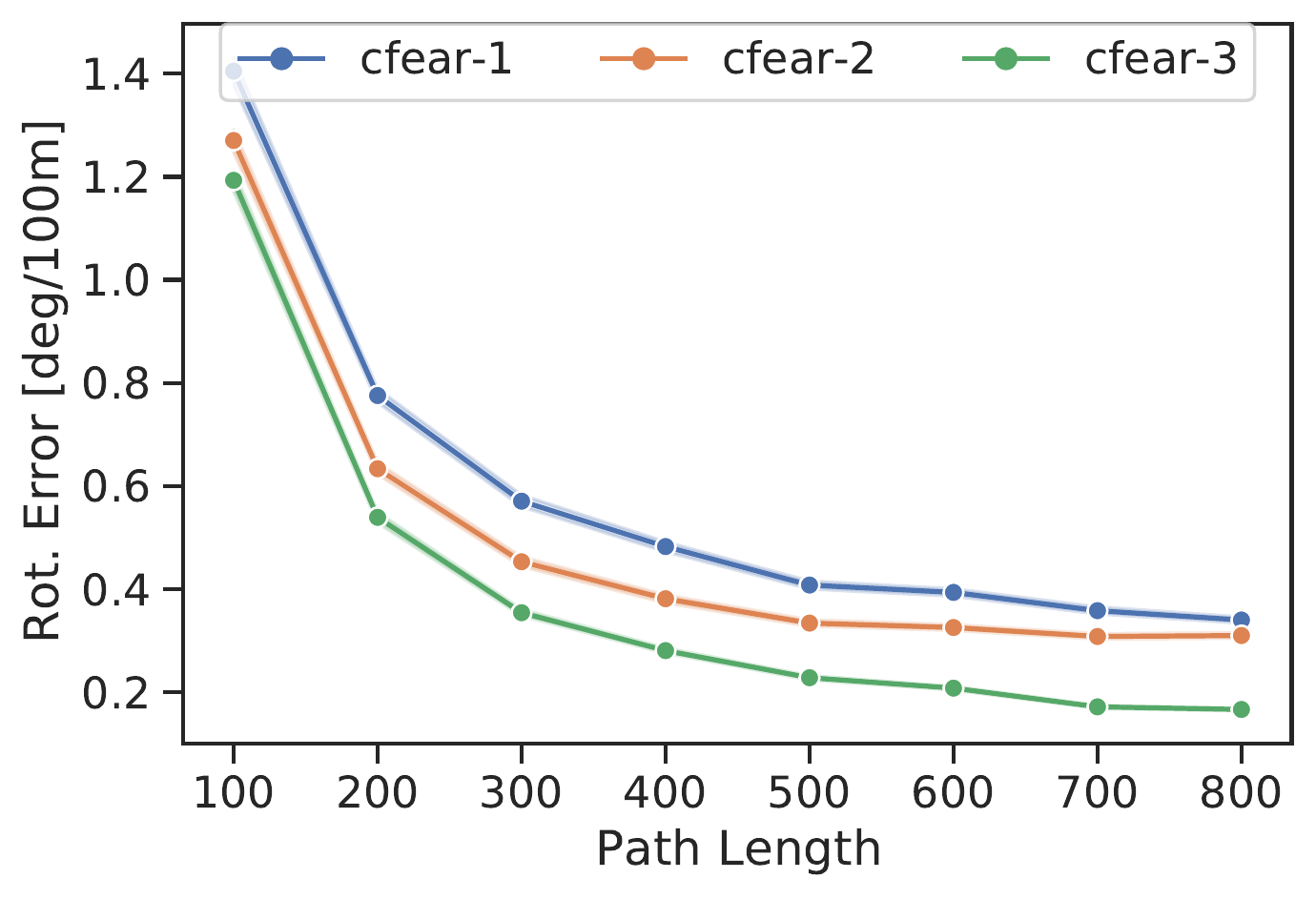}\label{fig:oxford_error_rot}}
  \end{center}
  \caption{Drift vs path distance on the Oxford dataset.
  }\label{fig:OxfordErrorPlot}
  \vspace{-0.5cm}
\end{figure}
\changed{
We compare our method to previously published odometry and SLAM methods that have been evaluated on the Oxford dataset.
All experiments in the comparative evaluation were conducted on an Intel i7-8700k CPU using a single thread. Aiming to assess the current state of spinning radar odometry compared to other modalities, we include baselines from vision and lidar. We use the evaluation of SuMa~\cite{behley-2018-suma} (lidar SLAM) that was carried out Hong et al.~\cite{hong2021radar}, using default parameters without tuning. As the method failed during their evaluation, drift was measured from the part of the trajectory until the first failure occurred. 
For the method \textit{RadarSLAM-odometry} and \textit{RadarSLAM-full} by Hong et al, we import results from their preprint~\cite{hong2021radar} as the article published to IJRR~\cite{hong-2022-radarslam} contain a partly different selection of sequences.
Our last baseline is the unsupervised lidar odometry method by Yoon et al.~\cite{9357964}. In their evaluation, they achieve results comparable to state of the art in lidar odometry within the KITTI dataset. However, both SuMa and Yoon's lidar odometry 
achieves lower performance on the Oxford dataset compared to the KITTI benchmark. 
\changedd{We believe the reduced performance can be attributed to the following factors: (i) KITTI benchmark provides motion-compensated lidar scans while no compensation was carried out in the Oxford or MulRan dataset. (ii) Compared to KITTI, the laser scanners in Oxford and MulRan have reduced field of view due to the mounting of the radars. (iii) In Oxford and Mulran, a 32-diode laser was used rather than the 64-diode laser in the KITTI dataset.} Hence, the reported performances of lidar results in the Oxford and MulRan datasets are likely pessimistic.
}

Our comparison is presented in the (``Sequence'') and (``Mean'') columns of Tab.~\ref{tab:OxfordTable} with corresponding trajectories of our method in Fig.~\ref{fig:sequences_oxford} and path error in Fig.~\ref{fig:OxfordErrorPlot}.
The learning-based method ``masking by moving''~\cite{barnes_masking_2020} yields the second-lowest error when trained and evaluated on the exact same route, yet with different traversals. However, when performing spatial cross-validation (to separate between training and evaluation environment) the error increases by 106\% for their setting \textit{Dual Cart}. To make a fair comparison, we evaluate our method to other supervised learning methods that have been spatially cross-validated (column ``Mean SCV'') to make sure these methods are not overfitted to the environment surrounding the route of the evaluation sequences~\cite{lovelace_geocomputation_2019}.
We found that \textit{CFEAR-s50} yields the lowest translation and rotation error in every sequence among comparable methods with an average translation error of $1.09\%$~(@5Hz)  Among our efficient methods, CFEAR-3 outperforms all odometry baselines over all sequences with an overall translation error of $1.31\%$~(@44Hz), followed by CFEAR-2 at $1.48\%$~(@111Hz) and CFEAR-1 ($1.79\%)$~(@160Hz).

Surprisingly, we found the (open loop, incremental) odometry error of all our configurations to be lower than state-of-the-art in radar SLAM by Hong et al.~\cite{hong2021radar} (RadarSLAM-full), where their trajectory is additionally corrected by loop closure and pose graph optimization. \changedd{On the other hand, the KITTI odometry metric does not fairly quantify the impact of SLAM, given that most loops in the Oxford and MulRan data sets stretches over lengths which exceed the upper KITTI range of 800~m.}
Moreover, \textit{CFEAR-3-s50} challenges the lidar SLAM method \textit{SuMa}~\cite{behley2018rss} with a slightly higher translation error ($1.09\%$) compared to $1.03\%$, and outperforms \textit{SuMa} in some of the sequences. At a slightly higher mean drift ($1.31\%$), our efficient method CFEAR-3 challenges \textit{SuMa} within some sequences.

One of our best results was obtained by tailoring the parameters ($k$, $s$ and resolution $r$) for a single sequence, we present the trajectory, after rigid alignment with ground truth, in Fig.~\ref{fig:oxford_best_sequence}. In this particular case, the odometry is accurate to the point where it is visually hard to detect the odometry errors in a larger part of the trajectory, although odometry has been estimated open-loop over $10$~km. 


%
 \begin{figure*}[h!]
  \begin{center}
    \subfloat[KAIST01]{\includegraphics[trim={0.0cm 0cm 0cm 0cm},clip,height=0.28\hsize]{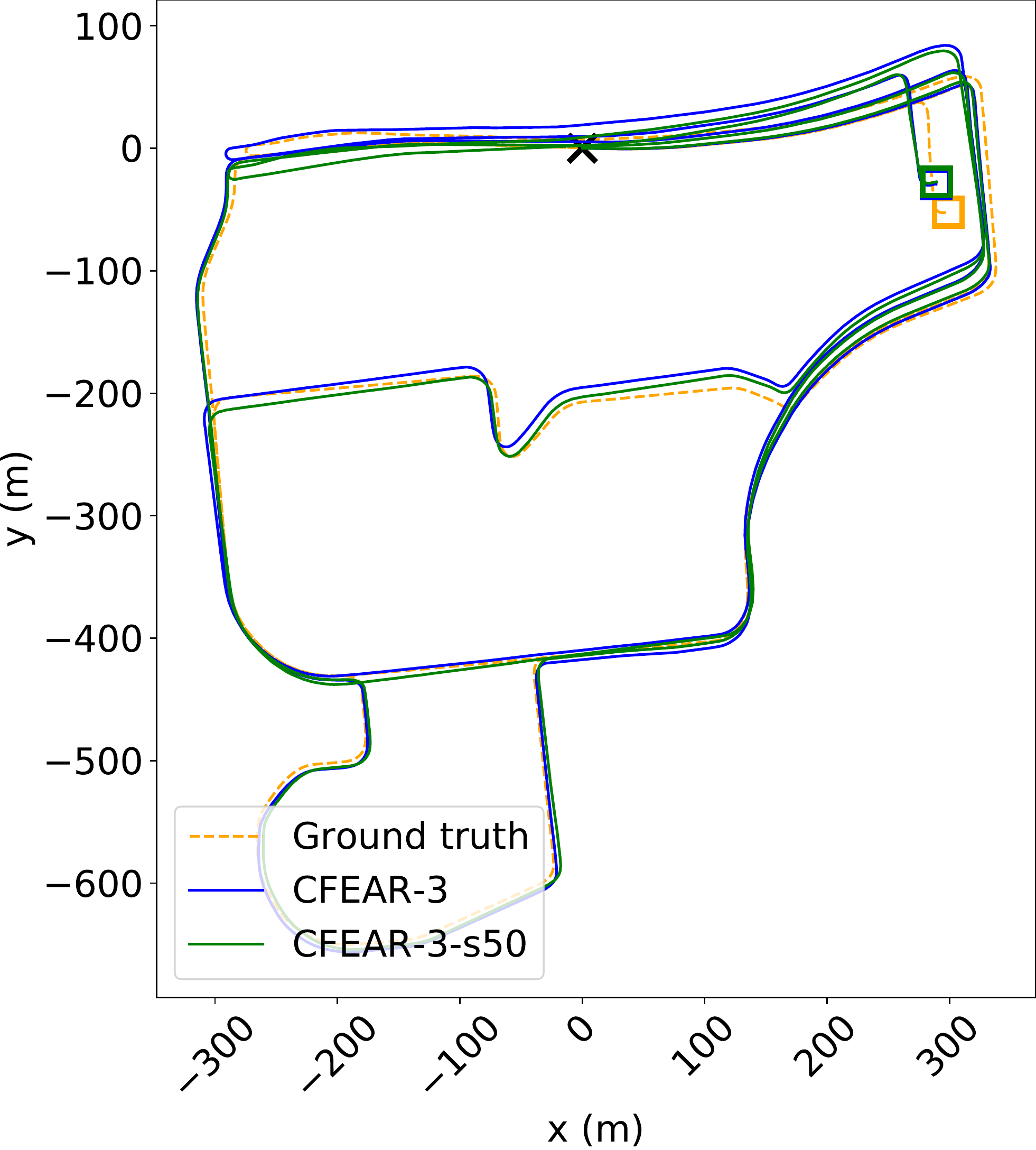}\label{fig:KAIST01}}
    \subfloat[KAIST02]{\includegraphics[trim={0.0cm 0cm 0cm 0cm},clip,height=0.28\hsize]{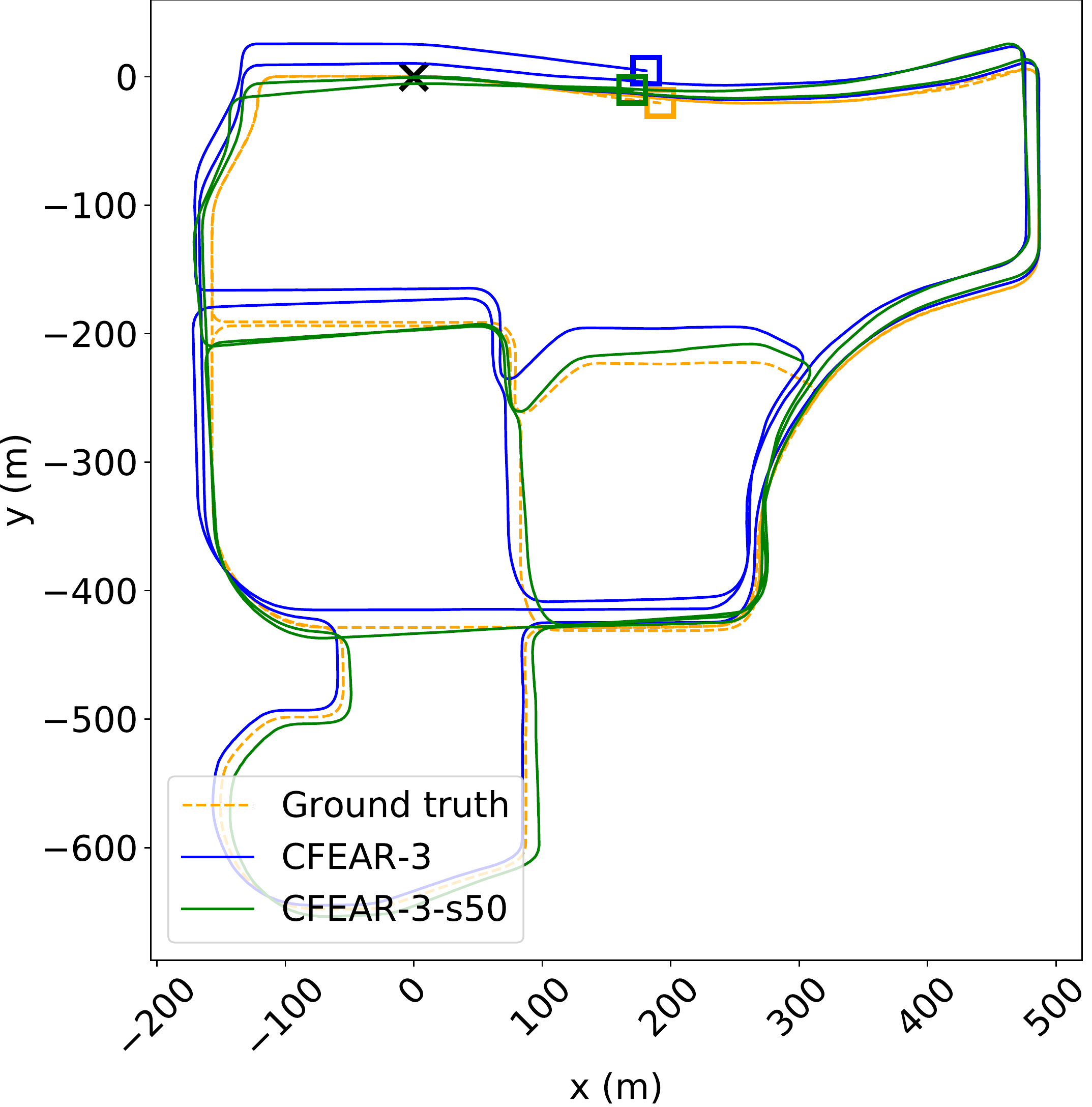}\label{fig:KAIST02}}
    \subfloat[KAIST03]{\includegraphics[trim={0.0cm 0cm 0cm 0cm},clip,height=0.28\hsize]{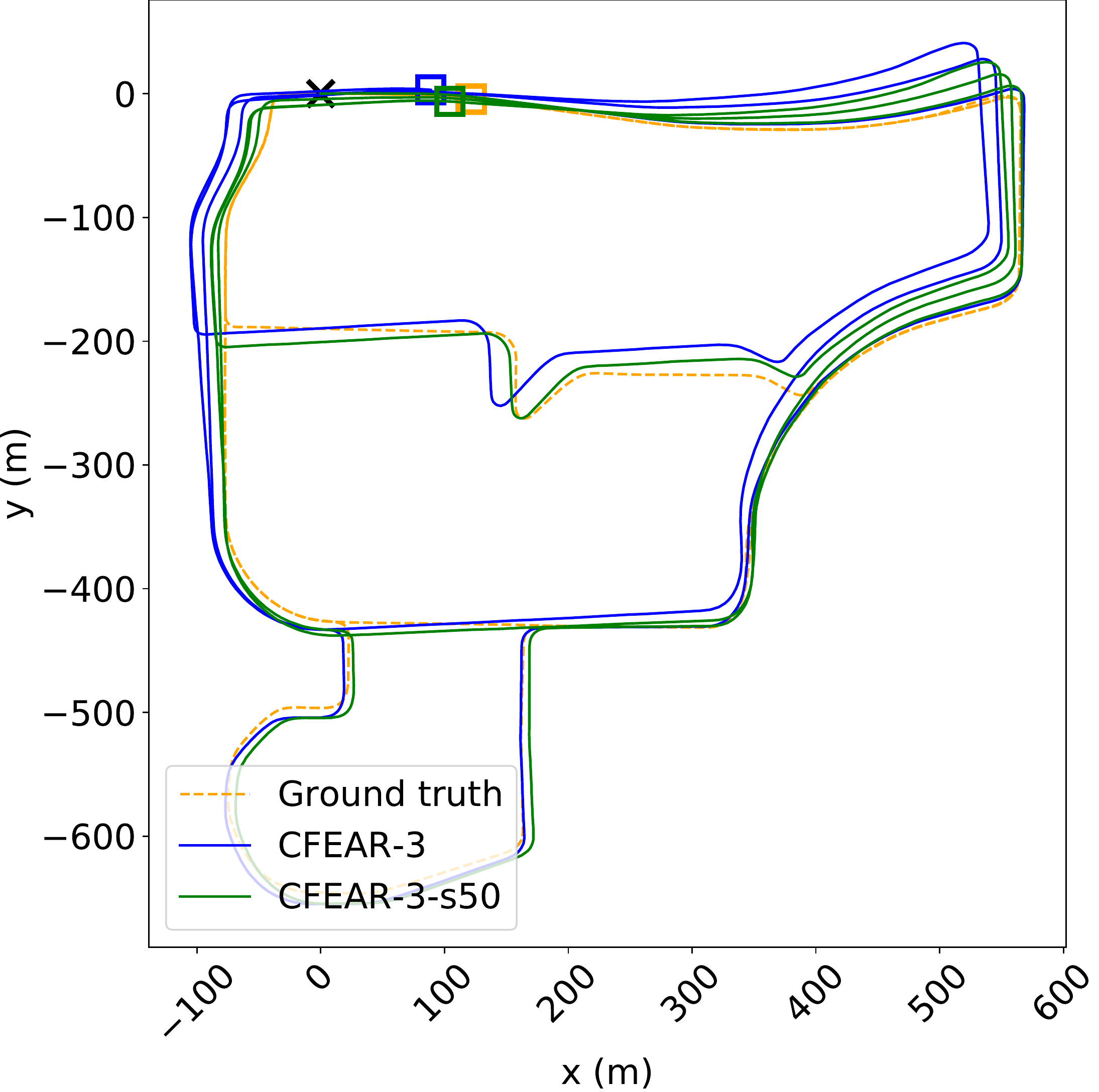}\label{fig:KAIST03}}
    \subfloat[DCC01]{\includegraphics[trim={0.0cm 0cm 0cm 0cm},clip,height=0.28\hsize]{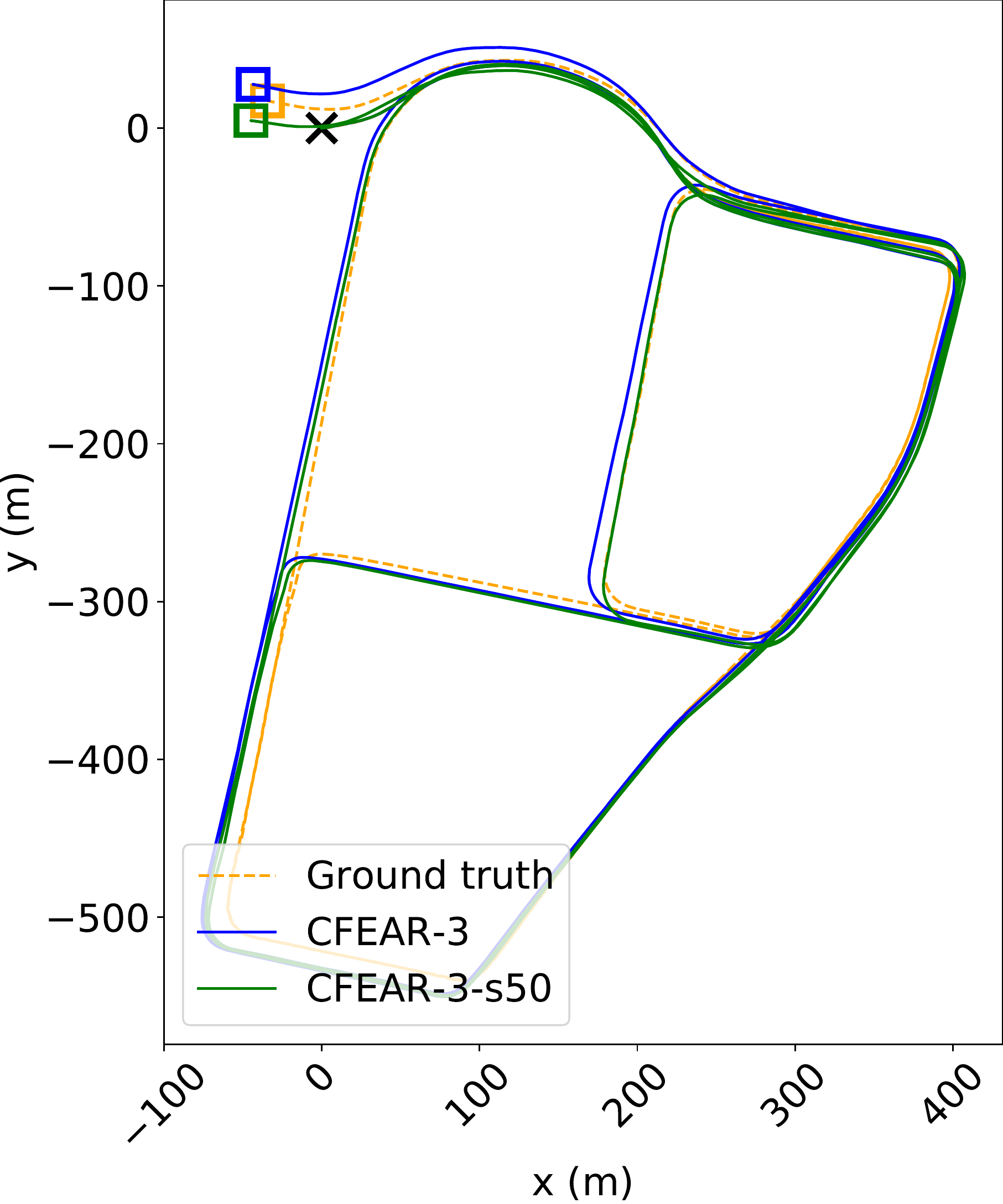}\label{fig:DCC01}}\\
    \subfloat[DCC02]{\includegraphics[trim={0.0cm 0cm 0cm 0cm},clip,height=0.25\hsize]{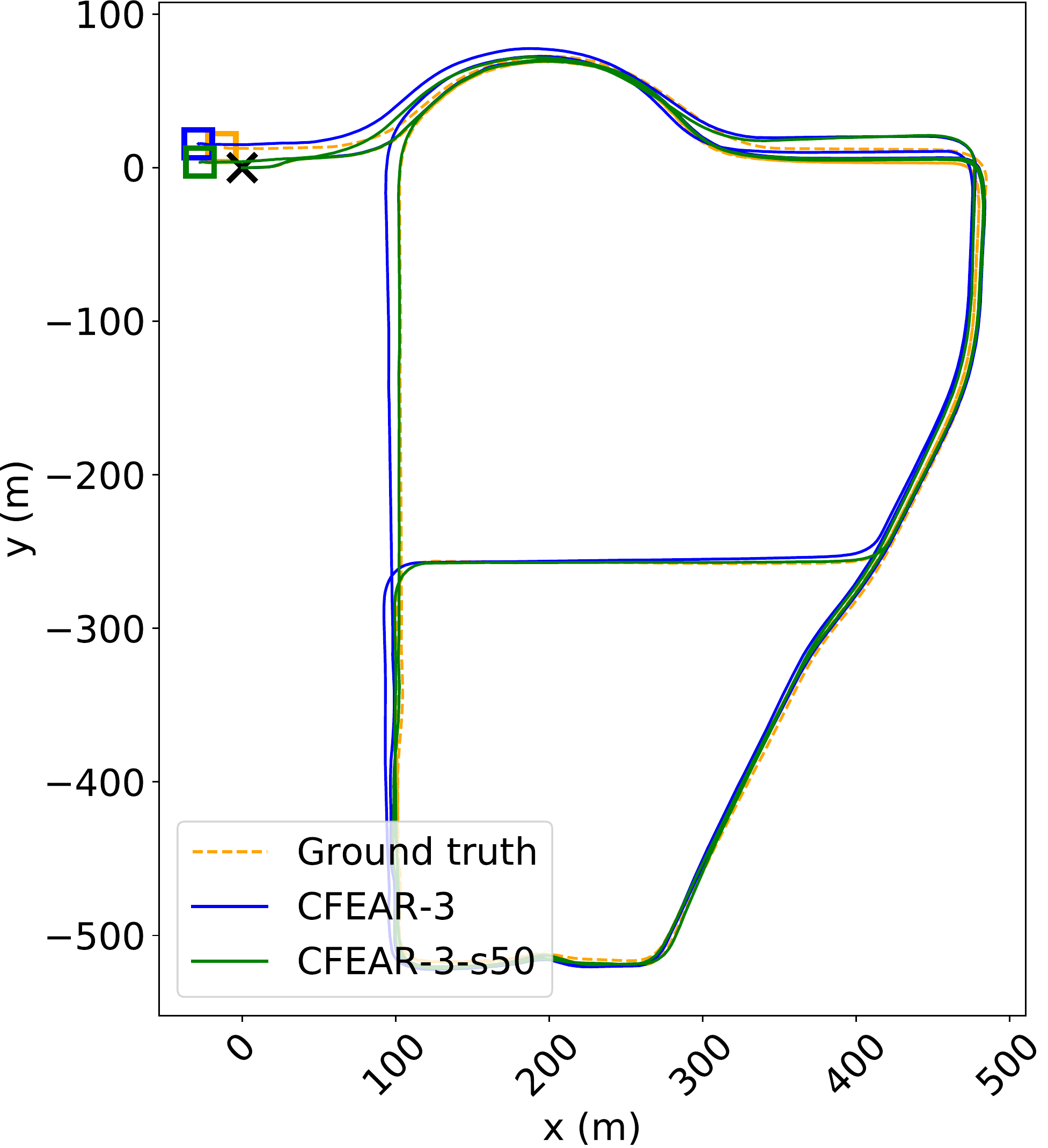}\label{fig:DCC02}}
    \subfloat[DCC03]{\includegraphics[trim={0.0cm 0cm 0cm 0cm},clip,height=0.25\hsize]{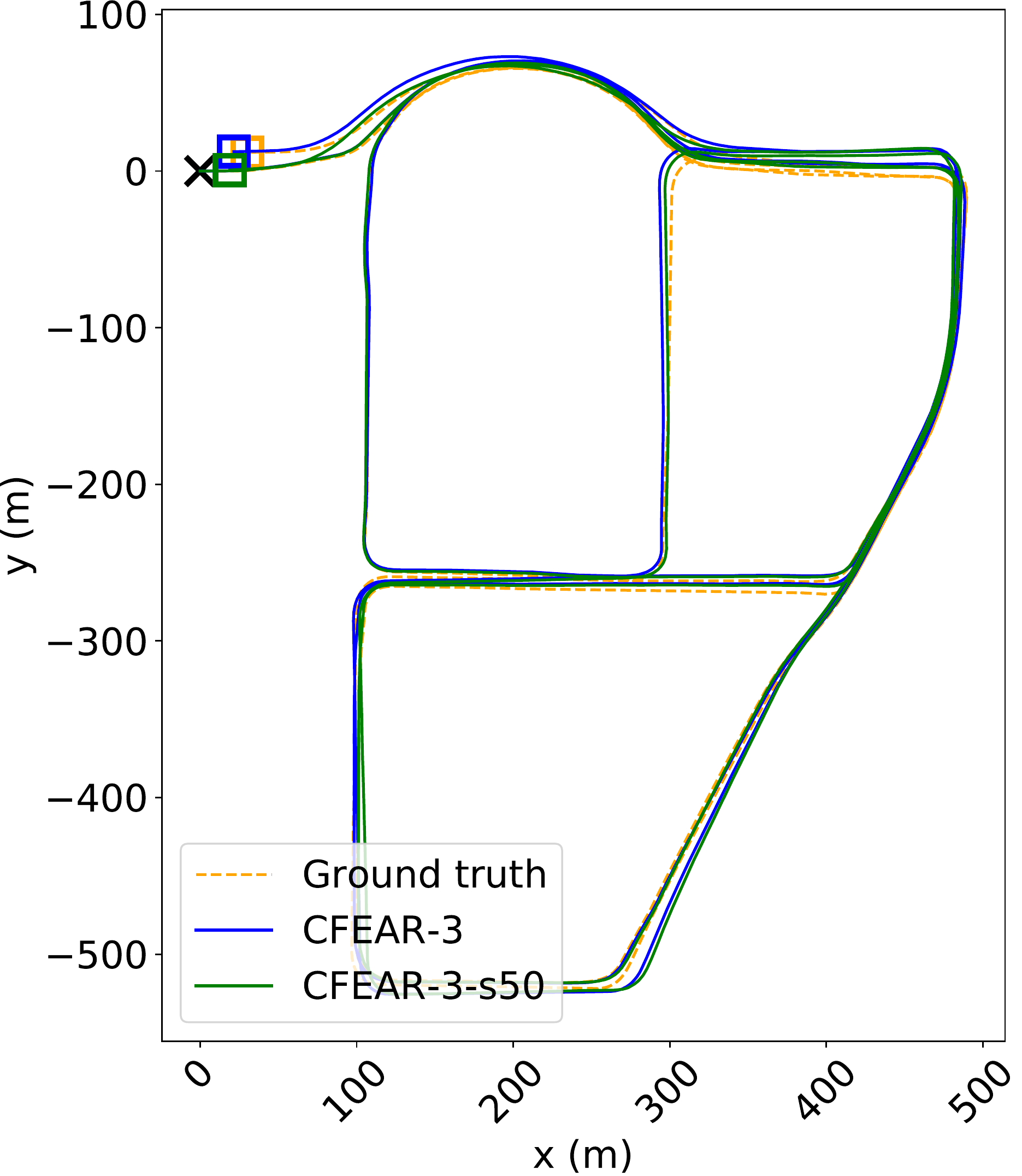}\label{fig:DCC03}}
    \hspace{1cm}
    \subfloat[][RIV01]{\includegraphics[trim={0.0cm 0cm 0cm 0cm},clip,height=0.25\hsize]{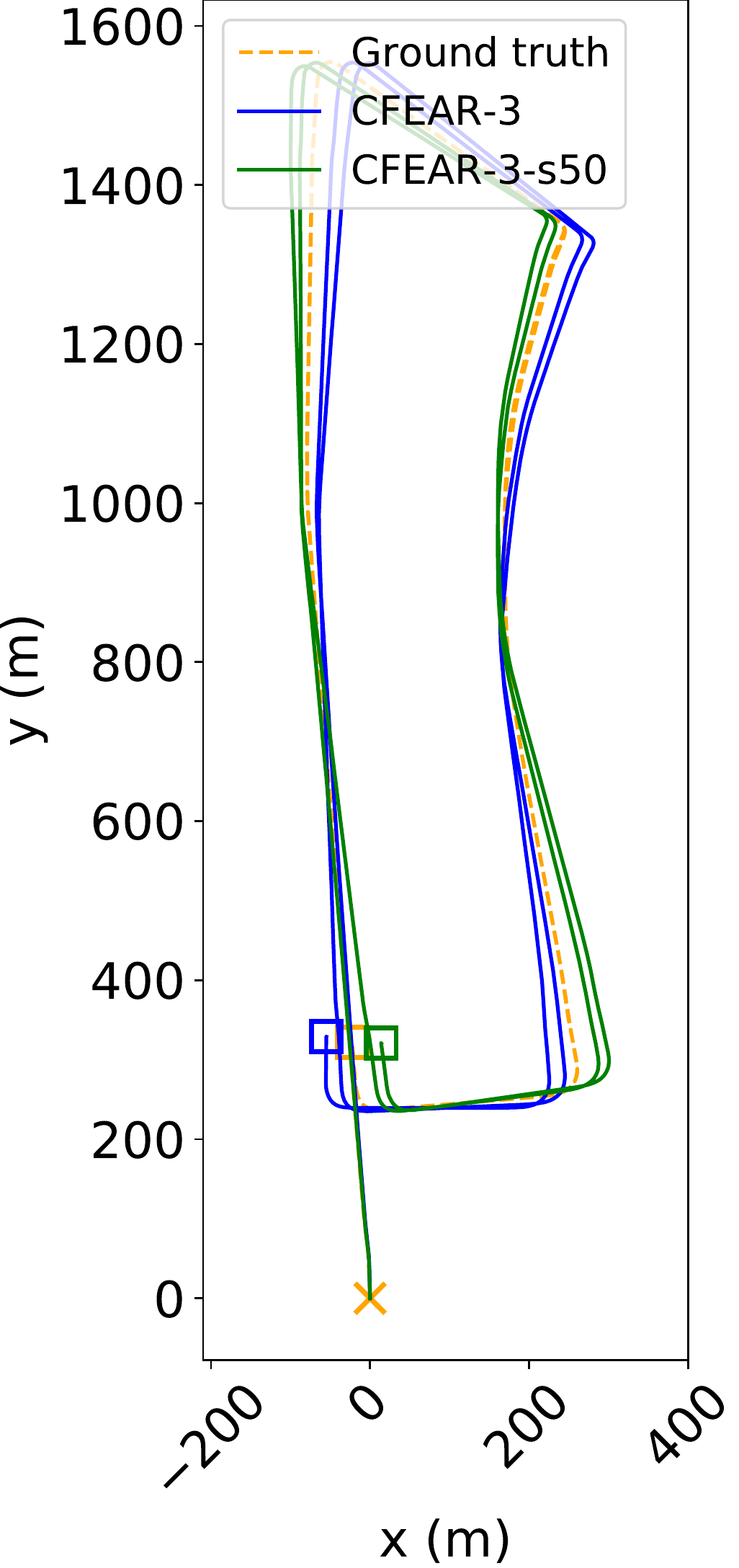}\label{fig:Riverside01}}
    \hspace{1cm}
    \subfloat[RIV02]{\includegraphics[trim={0.0cm 0cm 0cm 0cm},clip,height=0.25\hsize]{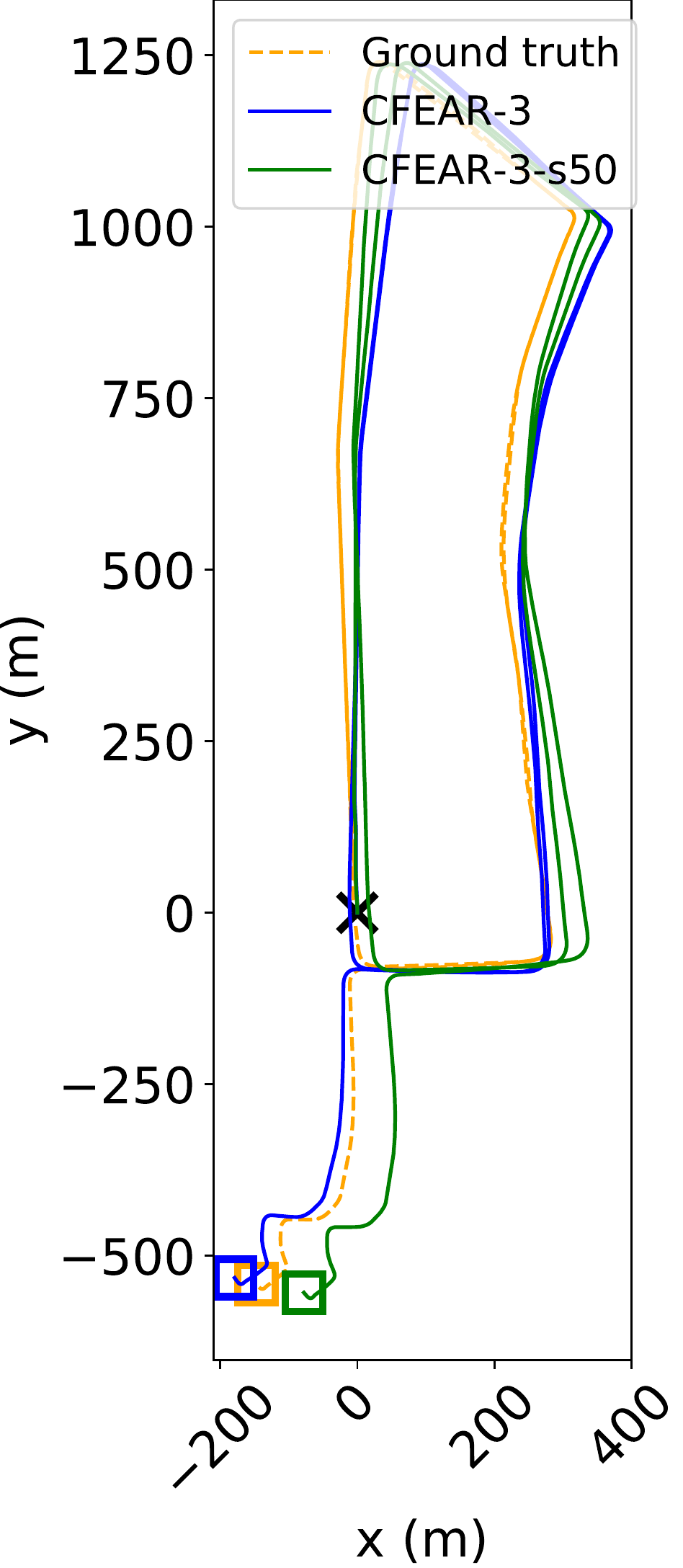}\label{fig:Riverside02}}
    \hspace{1cm}
    \subfloat[RIV03]{\includegraphics[trim={0.0cm 0cm 0cm 0cm},clip,height=0.25\hsize]{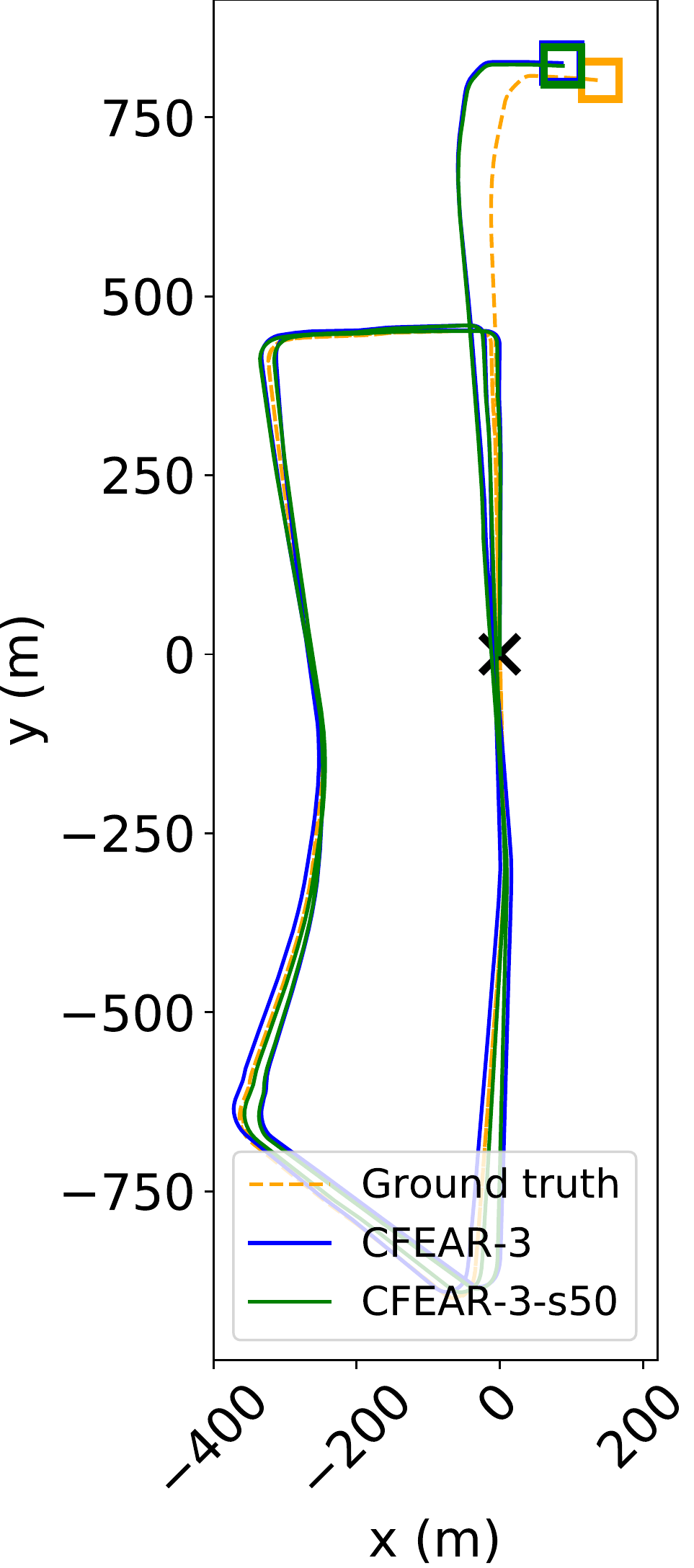}\label{fig:Riverside03}}
  \end{center}
  \caption{Radar odometry estimation for the MulRan sequences obtained with CFEAR-3 (blue) and CFEAR-3-s50 (green) compared to ground truth (orange). For comparison, we selected the same 9 sequences as in~\cite{hong-2022-radarslam}}\label{fig:MulRanTrajectories}
  \vspace{-0.3cm}
\end{figure*}
\begin{table*}[h!]
\centering
 \begin{adjustbox}{width=\textwidth}
\begin{tabular}{l|l|lllllllll|l|ll}
              & & & \multicolumn{1}{l}{\textbf{Sequence}}\\
\textbf{Method}  & \textbf{resolution} & KAIST01 & KAIST02 & KAIST03 & DCC01 & DCC02 & DCC03 & RIV01 & RIV02 & RIV03 & Mean & Mean SCV & Mean Opti.\\
\hline \\
SuMa Full~\cite{behley2018rss}  & 0.0583   & 2.9/0.8 & 2.64/0.6 & 2.17/0.6 & 2.71/0.4 & 4.07/0.9 & 2.14/0.6 & 1.66/0.6\changedd{$^P$} & 1.49/0.5\changedd{$^P$} & 1.65/0.4\changedd{$^P$} & 2.38/0.5 & {-} & {-}\\
\hline\\
LOAM~\cite{loam}  & -   & 2.70/0.82 & 2.80/0.84 & 7.54/0.79 & 3.16/0.86 & 2.64/0.74 & 2.23/0.74 & 4.25/0.88 & 4.14/0.90 & 4.21/1.02 & 3.74/0.84 & {-} & {-}
\\
RadarSLAM-Full~\cite{hong2021radar}  & 0.0583   &  1.75/0.5 & 1.76/0.4 & 1.72/0.4 & 2.39/0.4 & 1.90/0.4 & 1.56/0.2 &  3.40/0.9 & 1.79/0.3 & 1.95/0.5 & 2.02/0.4 & {-} & {-}\\
\hline\\
RadarSLAM-odometry~\cite{hong2021radar}  & 0.0583   &   2.13/0.7 & 2.07/\textbf{0.6} & 1.99/\textbf{0.5} & 2.70/\textbf{0.5} & 1.90/\textbf{0.4} & 1.64/\textbf{0.4} &  2.04/\textbf{0.5} & 1.51/\textbf{0.5} & 1.71/0.5 & 1.97/\textbf{0.5} & {-} & {-}
\\
CFEAR-1 & 0.0595 &  2.62/0.97 & 2.45/0.90 & 2.85/1.08 & 2.73/0.73 & 1.82/0.60 & 1.77/0.62 & 2.55/0.90 & 2.71/0.82 & 3.56/0.82 &  2.56/0.83 &  2.34/0.79 & 2.34/0.79\\
CFEAR-2  & 0.0595 &  2.12/0.81 & 1.93/0.74 & 2.08/0.87 & 2.44/0.63 & 1.65/0.54 & 1.41/0.50 & 2.30/0.80 & 2.07/0.66 & 2.60/0.59 &  2.07/0.68 & 1.93/0.66 & 1.93/0.66\\
CFEAR-3  & 0.0595 & 1.59/0.66 & 1.62/0.66 & 1.73/0.78 & 2.28/0.54 & 1.49/0.46 & 1.47/0.48 & 1.59/0.63 & 1.39/0.51 & 1.41/0.40 &  1.62/0.57 & 1.60/0.57 & 1.55/0.56
\\
CFEAR-3-s50  & 0.0595 & \textbf{1.48/0.65} & \textbf{1.51}/0.63 & \textbf{1.59}/0.75 & \textbf{2.09}/0.55 & \textbf{1.38}/0.47 & \textbf{1.26}/0.47 & \textbf{1.62}/0.62 & \textbf{1.35}/0.52 & \textbf{1.19/0.37} & \textbf{1.50}/0.56 & - & -
\\

\end{tabular}
  \end{adjustbox}
\caption{
\changed{Drift evaluated over 9 sequences from the MulRan dataset~\cite{gskim-2020-mulran}. We compare various methods for lidar and radar odometry, and lidar/radar SLAM that additionally correct the trajectory. Results are given in (\% translation error / deg/$100$~m). In the column ``Mean'' we report the best  available result of all methods, except for CFEAR for which we evaluate the configurations presented in Tab.~\ref{tab:Parameter}, optimized for speed and drift jointly.
Additionally, we provide a SCV for CFEAR (see Sec.~\ref{sec:scv_eval}), reporting results achieved when optimizing parameters for drift only within another environment (``Mean SCV''), parameters optimized for MulRan are reported in (``Mean Opti.'').
Results marked with $^p$ indicate that the numbers are reported from part of the trajectories up to a failure.}
}
\label{tab:MulRanTable}
\vspace{-0.2cm}
\end{table*}
\begin{table*}[h!]
\centering
 \begin{adjustbox}{width=\textwidth}
\begin{tabular}{l|l|lllllllll|l|ll}
              & & & \multicolumn{1}{l}{\textbf{Sequence}}\\
\textbf{Method}  & \textbf{resolution} & KAIST01 & KAIST02 & KAIST03 & DCC01 & DCC02 & DCC03 & RIV01 & RIV02 & RIV03 & Mean & Mean SCV & Mean Opti.\\
\hline \\
SuMa Full~\cite{behley2018rss}  & 0.0583 &  { -/38.7} & { -/31.9} & { -/46.0} & {-/13.5} & { -/17.8} & { -/29.6} & { -/-} & { -/-} & { - / -} & { -/22.9} & {-} & {-}  \\
\hline\\
RadarSLAM-Full~\cite{hong2021radar}  & 0.0583 & { -/6.9} & { -/{6.0}} & { -/{4.2}} & { -/12.9} & { -/9.9} & { -/{3.9}} & { -/9.0} & { -/{7.0}} & { -/{10.7}} & { -/{7.8}} & {-} & {-}  \\
\hline\\
PhaRaO-Full~\cite{9197231} & 0.0583 &  -/12.8 & -/12.8 & -/12.8 & -/13.26 & -/13.26 & -/13.26 & -/31.8 & -/31.8 & -/31.8 & -/19.3 & {-} & {-} \\
 CFEAR-1  & 0.0595 & 7.53/37.06 & 7.22/23.28 & 7.47/38.79 & 8.22/21.64 & 5.00/15.83 & 5.77/20.77 & 7.98/30.81 & 7.36/49.25 & 6.39/129.23 &  6.99/40.74 & {-} & {-} \\
CFEAR-2  & 0.0595 &7.18/17.26 & 6.70/16.58 & 6.96/23.80 & 7.99/17.32 & 4.70/7.95 & 5.46/6.81 & 7.39/22.06 & 6.86/47.28 & 5.83/80.52 &  6.56/26.62 & {-} & {-} \\
CFEAR-3  & 0.0595 & 6.37/8.72 & \textbf{6.01}/9.89 & 6.21/13.44 & 7.83/6.82 & 4.54/5.13 & 5.16/4.88 & \textbf{5.90}/10.98 & \textbf{5.38/3.26} & 4.50/17.83 &  \textbf{5.77}/8.99 & {-} & {-} \\
CFEAR-3-50  & 0.0595 & \textbf{6.34/7.31} & 6.04/\textbf{6.72} & \textbf{6.19/6.45} & \textbf{7.56/6.09} & \textbf{4.45/4.90} & \textbf{5.04/4.65} & 6.36/\textbf{10.24} & 5.65/16.74 & \textbf{4.63/12.99} & 5.81/\textbf{8.46} & {-} & {-} \\
\end{tabular}
  \end{adjustbox}
\caption{Complementary pose accuracy (RPE) and full Absolute Trajectory Error error (ATE). The trajectories have been aligned with ground truth to make a fair comparison between the methods. For each sequence we report (RPE [cm]/ATE - RMSE [m]). 
\textit{CFEAR-3-s50} achieves an overall ATE of 8.46m, which is 56\% lower compared to PhaRaO and only slightly higher compared to RadarSLAM-full.
}
\label{tab:MulRanATE}
\vspace{-0.2cm}
\end{table*}

\subsection{Comparative evaluation within the MulRan dataset}
\label{sec:Mulran_eval}
\begin{figure}
  \begin{center}
  \vspace{-0.2cm}
    \subfloat[Translation error]{\includegraphics[trim={0.0cm 0cm 0cm 0.2cm},clip,width=0.49\hsize]{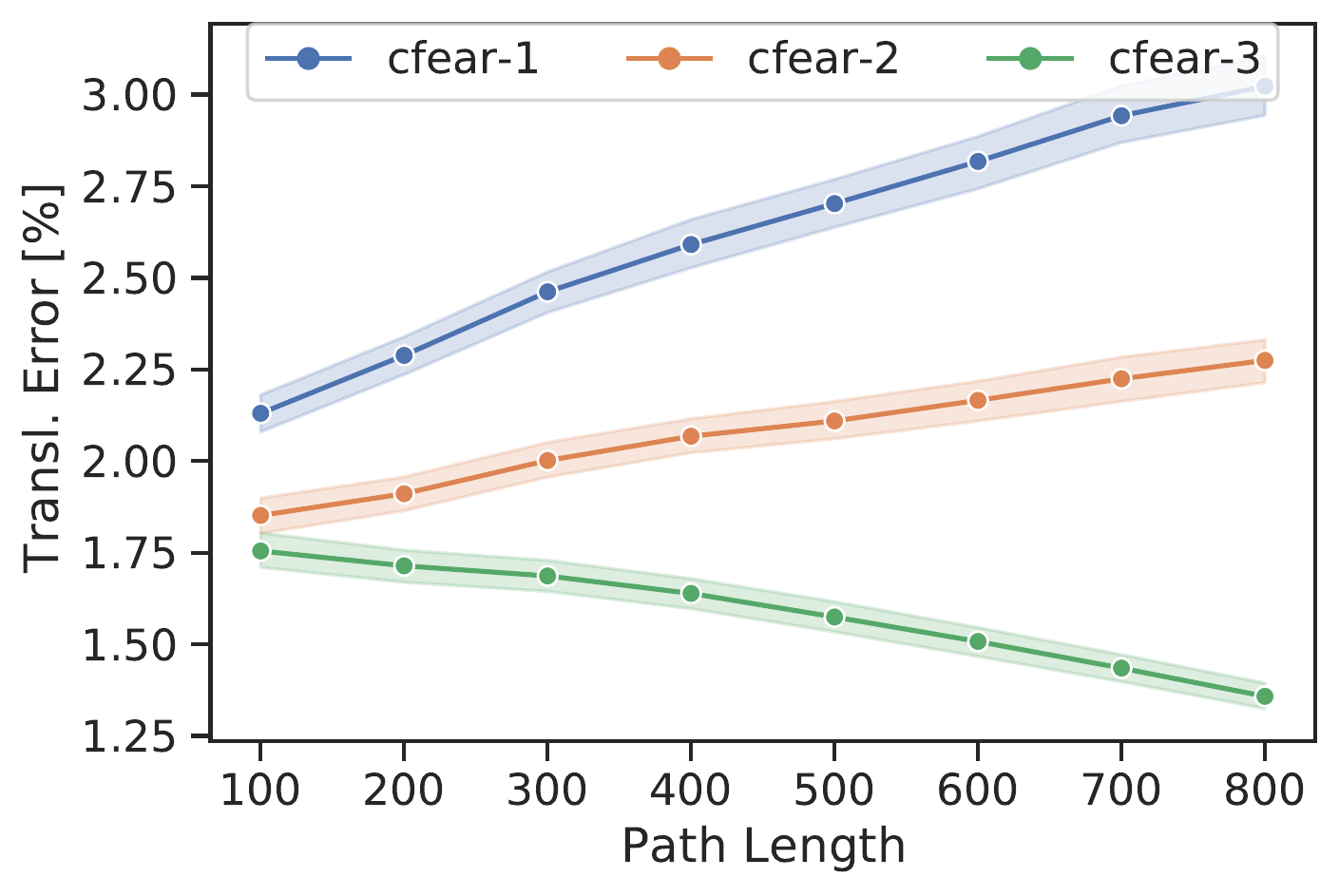}\label{fig:mulran_error_trans}}
    \subfloat[Rotation error]{\includegraphics[trim={0.0cm 0cm 0cm 0.2cm},clip,width=0.49\hsize]{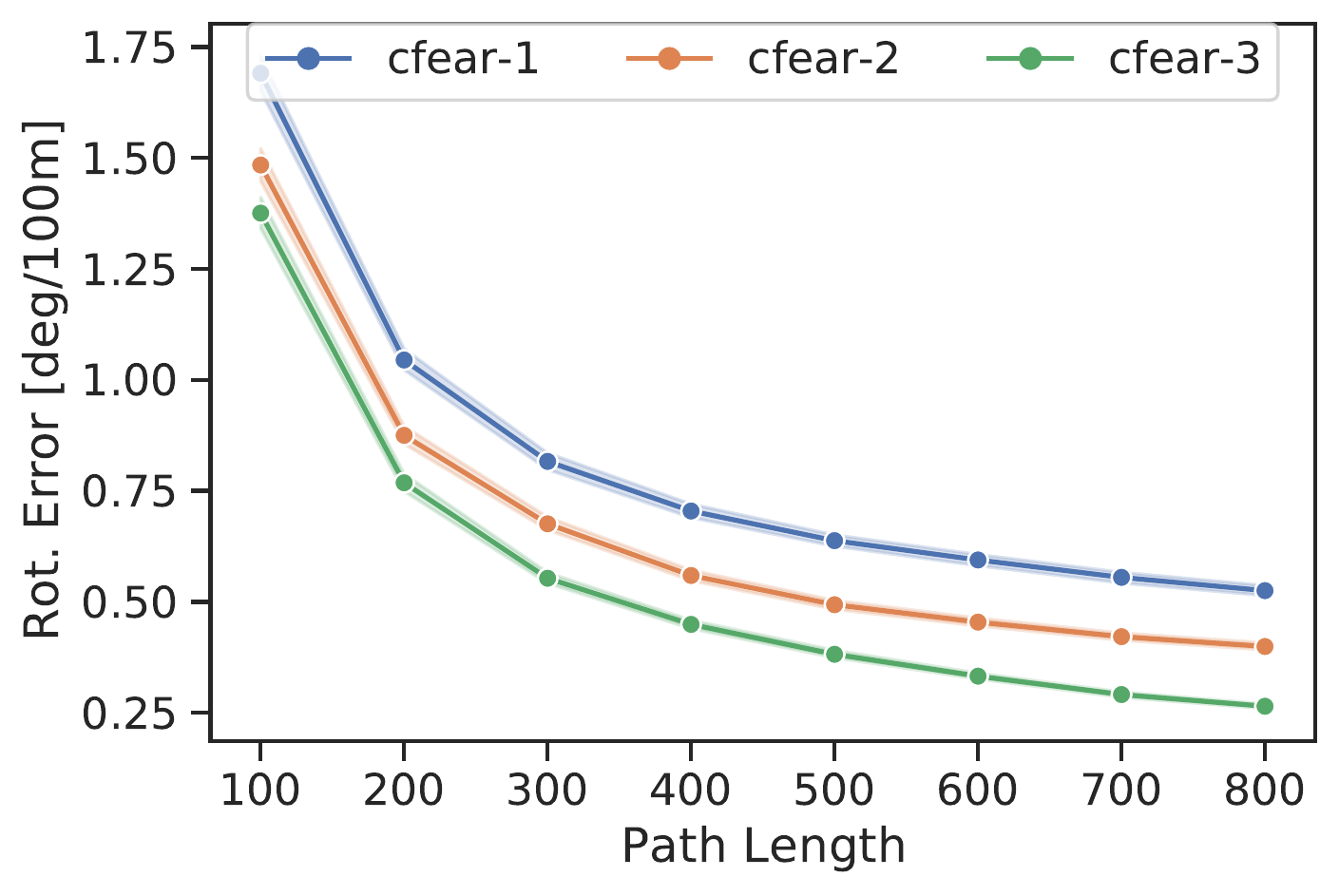}\label{fig:mulran_error_rot}}
    \vspace{-0.2cm}
  \end{center}
  \caption{Drift vs path distance on the MulRan dataset.}\label{fig:MulranErrorPlot}
  \vspace{-0.5cm}
\end{figure}

Without changing any parameter from the Oxford dataset, we also evaluated our method on 9 sequences from the MulRan dataset, 3 traversals for each of the sequences DCC (average 4.9km), KAIST (6.1km), and Riverside (6.8km). These sequences were selected to complement the urban Oxford dataset in terms of structural diversity, to quantitatively evaluate generalization across environments, and to compare our method with RadarSLAM~\cite{hong2021radar} and PhaRaO~\cite{9197231}.
We include results of the lidar odometry method LOAM~\cite{loam} without motion compensation. Note that the lidar has reduced field-of-view by roughly 70deg as the sensor is partly blocked by the radar. Hence, this dataset slightly favors the radar.
As the method ``PhaRaO'' by Park et al.~\cite{9197231} is missing the KITTI odometry metrics in their original publication, we complement the odometry performance with RPE and Absolute Trajectory Error (ATE).
The exact sensor range resolution used to record the dataset is not specified in the dataset documentation. Hence, we calculated the resolution from the max range and the number of range bins: $\gamma=\frac{\text{max distance}}{n}=0.0595$~m.
We present the odometry error in Tab.~\ref{tab:MulRanTable}, the estimated trajectories in ~\ref{fig:MulRanTrajectories}, and the path error in Fig.~\ref{fig:MulranErrorPlot}.
We found that both CFEAR-3-s50 ($1.5\%)$) and the more efficient CFEAR-3 ($1.62\%)$) outperforms RadarSLAM-Full, LOAM and SuMa in translation error. The translation error in these sequences is slightly larger compared to  the Oxford dataset, which we believe reflects that the environment is slightly more diverse and less structured. In Tab.~\ref{tab:MulRanATE} we provide complementary results on pose accuracy and ATE.
ATE is generally not well suited for quantifying odometry performance and largely favors methods for SLAM that can correct for drift at scale. However, we found our CFEAR-3 and CFEAR-3-s50 produce accurate odometry even over the full trajectories beyond 800m. Surprisingly, over the full trajectory, our method achieved an ATE only slightly higher compared to RadarSLAM-full, and 56\% lower compared to PhaRaO.

\changed{
\subsection{Spatial Cross-Validation}
\label{sec:scv_eval}
The previous sections demonstrated that parameters can be tuned for multiple environments \changedd{and run-time performance simultaneously. These results are presented in the ``Mean'' column of Tab.~\ref{tab:OxfordTable} and Tab.~\ref{tab:MulRanTable}.}
However, it is desirable to know how environment-specific parameter tuning affects the performance when presented to new unseen environments. For that reason, we carried out spatial cross-validation (with over $98$ million pose estimates) in the Oxford and MulRan datasets using the efficient methods CFEAR-(1-3).
The parameters were first optimized for the Oxford sequences and then evaluated on the MulRan sequences and vice versa. Specifically, we use exhaustive grid search to optimize the feature extraction parameters: $z_{min}$, $k$ and resolution $r$, and the matching parameters: resolution $r$, keyframes $s$, $\mathcal{L}_{\delta}$ and residual weight $w$. 
The following parameter space was explored: $z_{min}=\{50, 60, 70, 80 \}$, $k=\{10,12,15,30,40,50\}$, $r=\{2.5, 2.75, 3.0, 3.5\}$, $s=\{1,2,3,4,5,6\}$, $\mathcal{L}_{\delta}=\{0.1, 0.2, 0.3, 0.4\}$ and $w=\{0,4\}$. To retain the characteristics of the methods (Efficient, Balanced, Low Drift), we constrain the search space of CFEAR-(1/2) by $k\leq15$ (conservative filtering) and keep the keyframes to ($s=1$ and $s=3$) respectively as presented in Tab.~\ref{tab:Parameter}. The cross-validated and dataset-optimized performance is reported in the columns ``Mean SCV'' and  ``Mean Opti.'' in Tab.~\ref{tab:OxfordTable} and Tab.~\ref{tab:MulRanTable}. \changedd{
Note that in this case, parameters are optimized for drift only. Consequently, both ``Mean Opti.'' and  ``Mean SCV'' may yield lower drift compared to ``Mean'' which was tuned for run-time performance as an additional objective.} 
Interestingly, the mean SCV (when averaged over both datasets) is only mariginally higher compared to the optimized drift: CFEAR-1 ($2.01\%$ up from $1.98\%$), CFEAR-2 ($1.66\%$ no increase from $1.66\%$) and CFEAR-3 ($1.41\%$ up from $1.37\%$). These results clearly demonstrate that our parameters are not overfitted to each of the evaluated datasets. 
}
\subsection{Qualitative evaluation of generalization}
\label{sec:qualitative_eval}
As previous experiments were carried out in large urban environments, the following experiments intend to evaluate how our method generalizes to substantially different environment types. Specifically, we present a qualitative evaluation using the three datasets: \emph{Kvarntorp} (underground mine), \emph{Volvo-CE} (outdoor, woods and open field) and \emph{Orkla} (indoor warehouse). These three datasets represent a range of relevant robotics use cases.
 \changed{Since no ground truth is available in these datasets, the odometry error in Kvarntorp and Volvo-CE is estimated by comparing the map of points as projected by the odometry before and after revisiting a location and making a visual estimation of the error. Both absolute and percentual final errors are reported. These estimated errors have a larger variance compared to the drift reported in the previous sections and should be interpreted as rough measurements.}
 For Orkla, the trajectory is much shorter and we instead qualitatively inspect trajectory smoothness and map blur. 

As in Sec.~\ref{sec:Mulran_eval}, we used the same parameter setting in all experiments to test how well our method generalizes to different environments. In the Orkla dataset, we additionally evaluate our method with a smaller resolution to demonstrate how \papertitle{} robustly estimates odometry even with low surface point quality. In each of these datasets, we equipped a vehicle with a Navtech CIR154XH radar, configured with a range resolution of $\gamma=0.15$~m in \changedd{Volvo-CE and Kvarntorp, and $\gamma=0.0438$~m in the Orkla dataset.} 

 \begin{figure}[h!] 
  \begin{center}
    \subfloat[Overview of the VolvoCE vehicle test track. A wheel loader with radar started inside the white tent (right) and was driven (red) in a large loop on a gravel road and then in two smaller loops on an uneven path in the forest. The locations of the photos in Figures (b) and (c) are indicated along the trajectory. The following final errors was measured over the large loop, CFEAR-2 ($11.5$~m, $1~\%$), followed by CFEAR-3 ($16.5$~m, $1.4~\%$) and CFEAR-1 ($27.5$~m, $2.4~\%$). 
    ]%
    {\begin{tikzpicture}
    \node[anchor=south west,inner sep=0] (image) at (0,0) {\includegraphics[trim={0.0cm 0cm 0cm 0cm},clip,width=0.99\hsize]{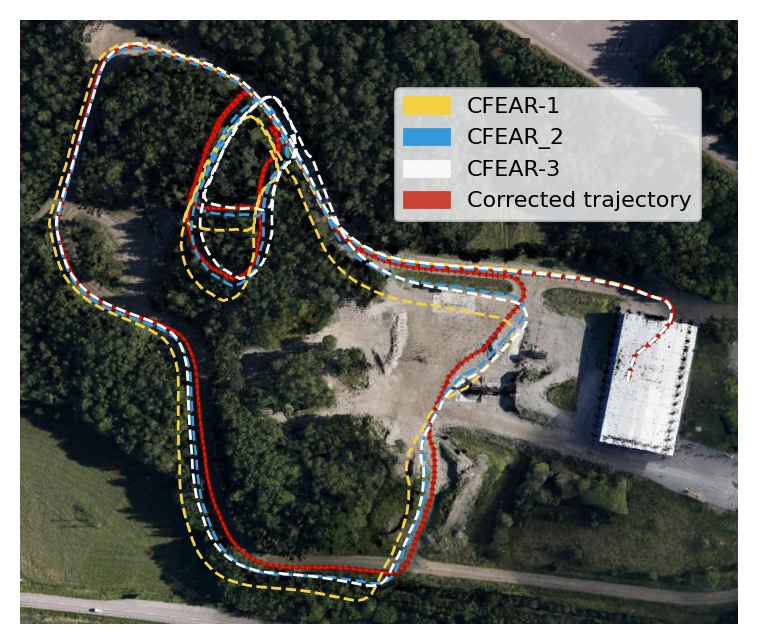}\label{fig:volvo_trajectory}};
    \begin{scope}[x={(image.south east)},y={(image.north west)}]
    \node[white!70!white] at(0.32,0.72) {(\textbf{b})};
    \node[white!70!white] at(0.45,0.72) {(\textbf{c})};
    \end{scope}
    \end{tikzpicture}}\\
        \subfloat[Driving on a narrow uneven path in the forest. ]{\includegraphics[trim={0.0cm 0cm 0cm 0cm},clip,height=40mm]{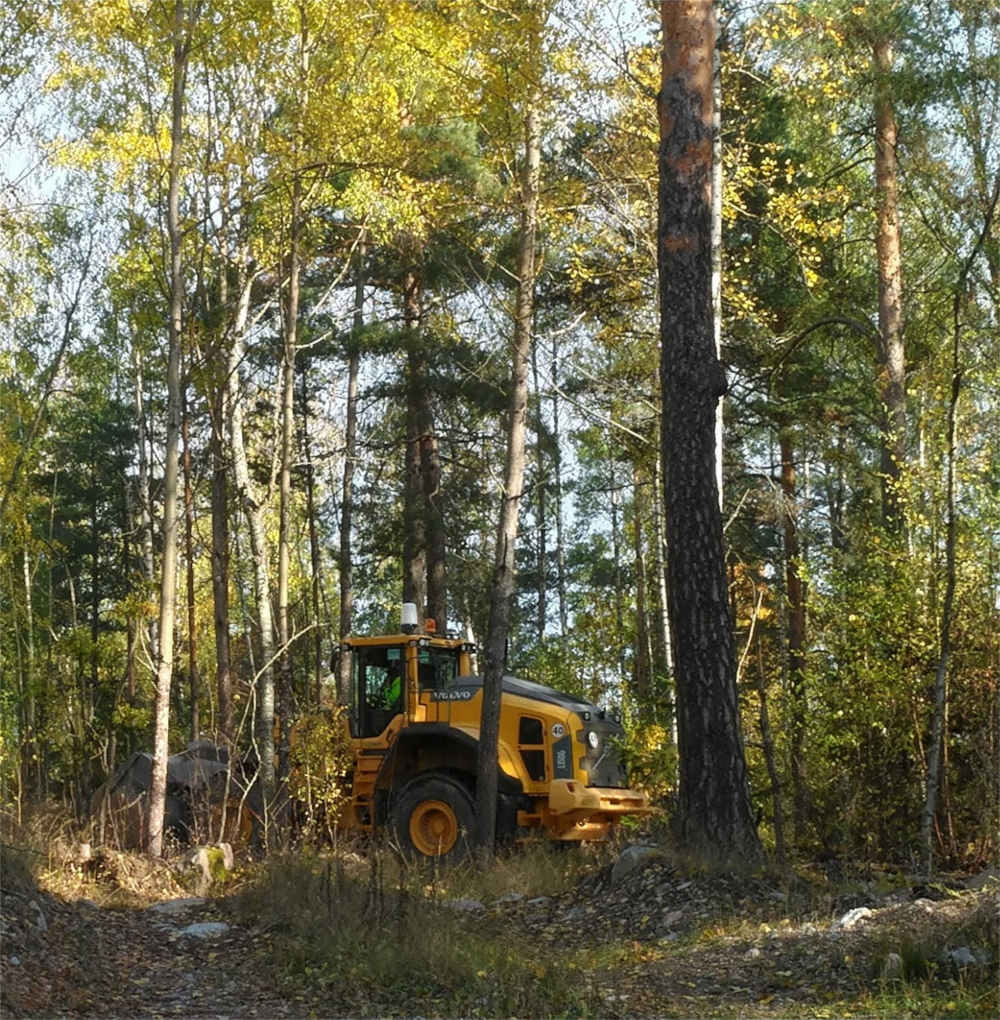}\label{fig:volvo_forest}}
        \qquad\subfloat[Driving on a gravel road uphill.]{\includegraphics[trim={0.0cm 0cm 0cm 0cm},clip,height=40mm]{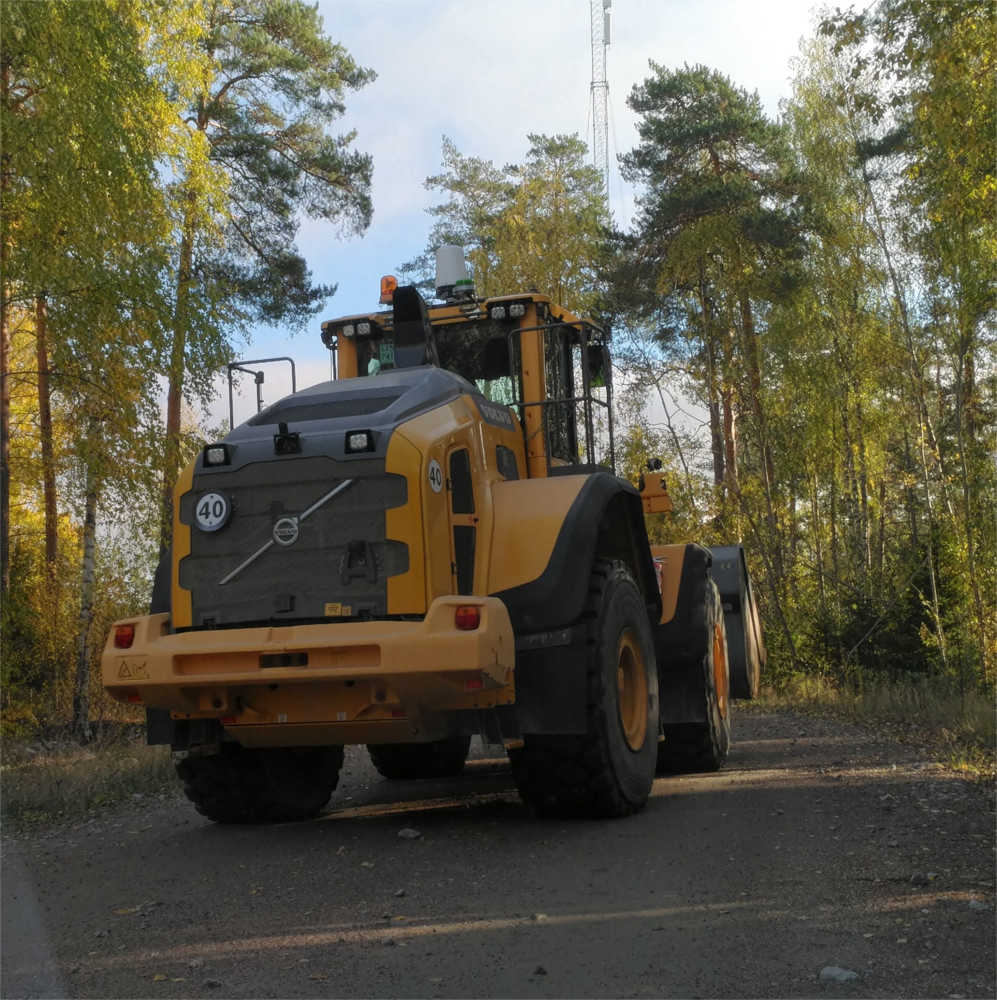}\label{fig:volvo_road}}
    \caption{Overview of semi-structured VolvoCE dataset.\label{fig:volvo_test_track_and_vehicle}}
  \end{center}
  \vspace{-0.6cm}
  \end{figure}

\subsubsection{Volvo CE -- outdoor, woods and open field}
In this dataset, the wheel-loader in Fig.~\ref{fig:volvo_road} drove through a forest environment over a distance of $1605$~m with an average speed of $10$~km/h. The actual trajectory of the wheel-loader, obtained by correcting radar odometry using pose-graph optimization, is visualized Fig.~\ref{fig:volvo_trajectory} together with the estimated trajectories using CFEAR-(1-3)
In the first part of the trajectory (large loop), the wheel-loader started within a tent and drove $1150$~m over non-planar gravel roads and slopes in an adjacent forest. The final segment of the trajectory was driven off-road within the forest, closing two smaller loops that partly overlap. This data set is difficult as the combination of clutter with few well-defined structures and the coarse sensor resolution ($\gamma=0.175$~m) can make it challenging to compute consistent features.
We have visually estimated the error in the final pose as described above.
\changed{
In the large loop, CFEAR-2 ($11.5$~m, $1~\%$) is most accurate, followed by CFEAR-3 ($16.5$~m, $1.4~\%$) and CFEAR-1 ($27.5$~m, $2.4~\%$). 
During the final segment of the trajectory, we expect part of the estimated small-loop trajectories to overlap as the wheel-loader was repeatedly driven on the same trail. CFEAR-3 is more accurate in this region (lower drift than we can measure, $<1$~m), which can best be seen by the trajectory being more consistent with the corrected trajectory in Fig.~\ref{fig:volvo_test_track_and_vehicle}a. In the same segment, the final error is $2$~m for CFEAR-1 and $3$~m for CFEAR-2.
}

\subsubsection{Kvarntorp -- underground mine}
\label{sec:kvarntorp_sec}
\begin{figure}[h!]
    \subfloat[Corrected map of Kvarntorp mine and estimated trajectories.]{\includegraphics[trim={0cm 0cm 0cm 0cm},clip,width=0.95\hsize]{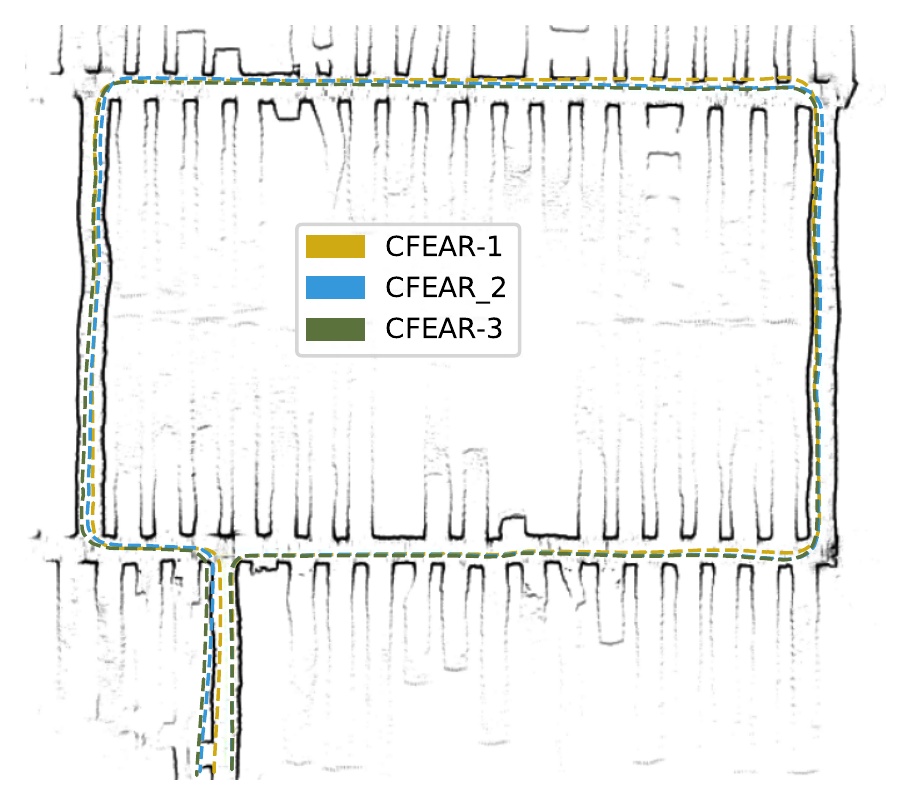}\label{fig:kvarntorp_map}}\\
    \includegraphics[trim={0.0cm 0cm 0cm 0cm},clip,height=0.29\hsize,angle=0]{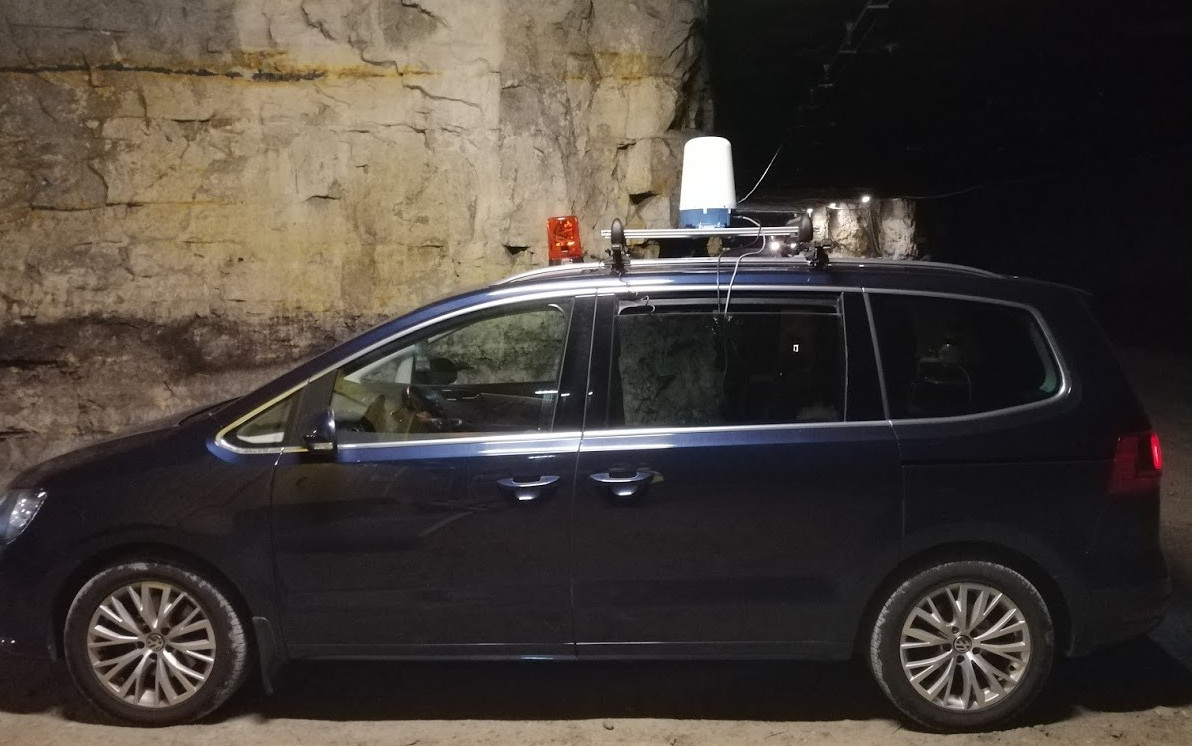}\label{fig:kvarntorp_vehicle}
      \includegraphics[trim={0.0cm 1cm 0cm 2cm},clip,height=0.29\hsize,angle=0]{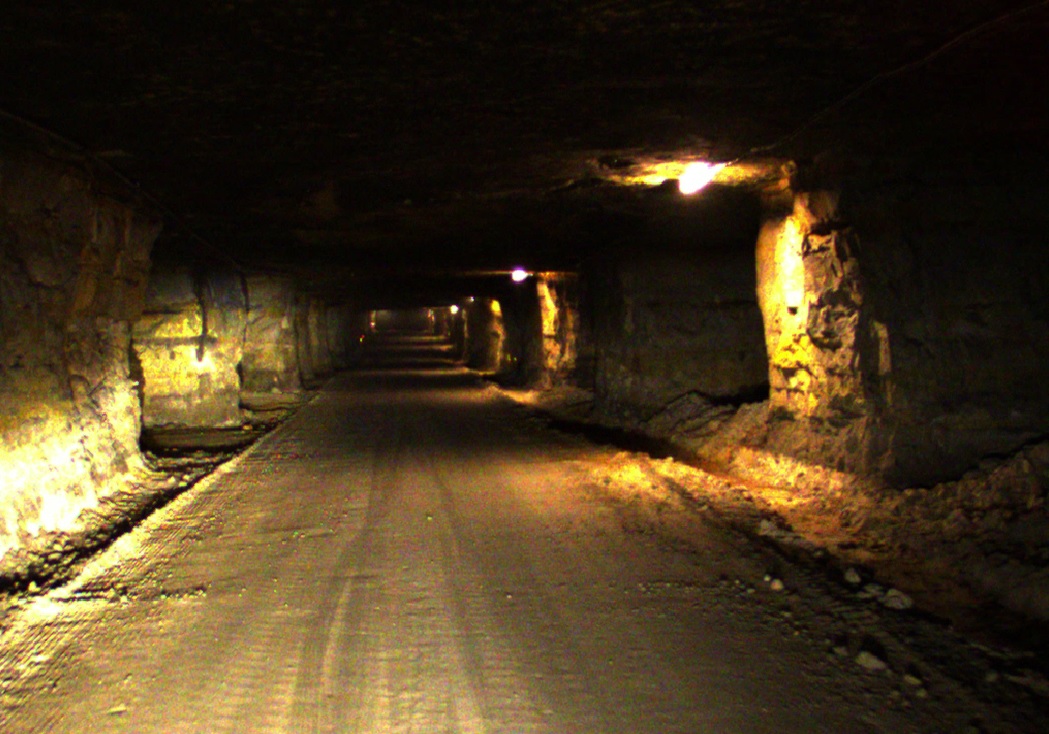}\label{fig:kvarntorp_mine}
    \caption{Kvarntorp dataset collected within an underground mine. A car with a roof-mounted radar was driven $1235$~m at a speed of $10$km/h in a loop to an endpoint close to the starting location. The vertical passages are feature-poor and challenging to localize within.
    }\label{fig:kvarntorp_example}
  \vspace{-0.4cm}
\end{figure}

\begin{figure}[h!]
  \begin{center}
    \subfloat[][CFEAR-1]{\includegraphics[trim={0.3cm 0.5cm 0.6cm 0cm},clip,width=0.5\hsize]{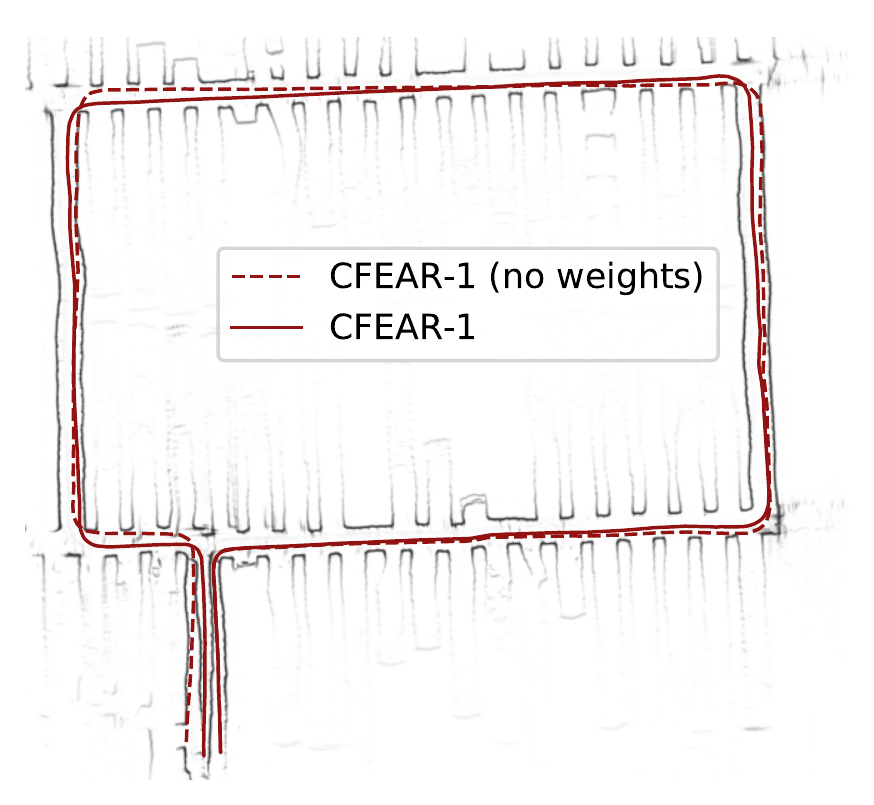}\label{fig:kvarntorp_cfear1}} 
    \subfloat[CFEAR-2]{\includegraphics[trim={15.0cm 1.5cm 11.5cm 0cm},clip,width=0.5\hsize]{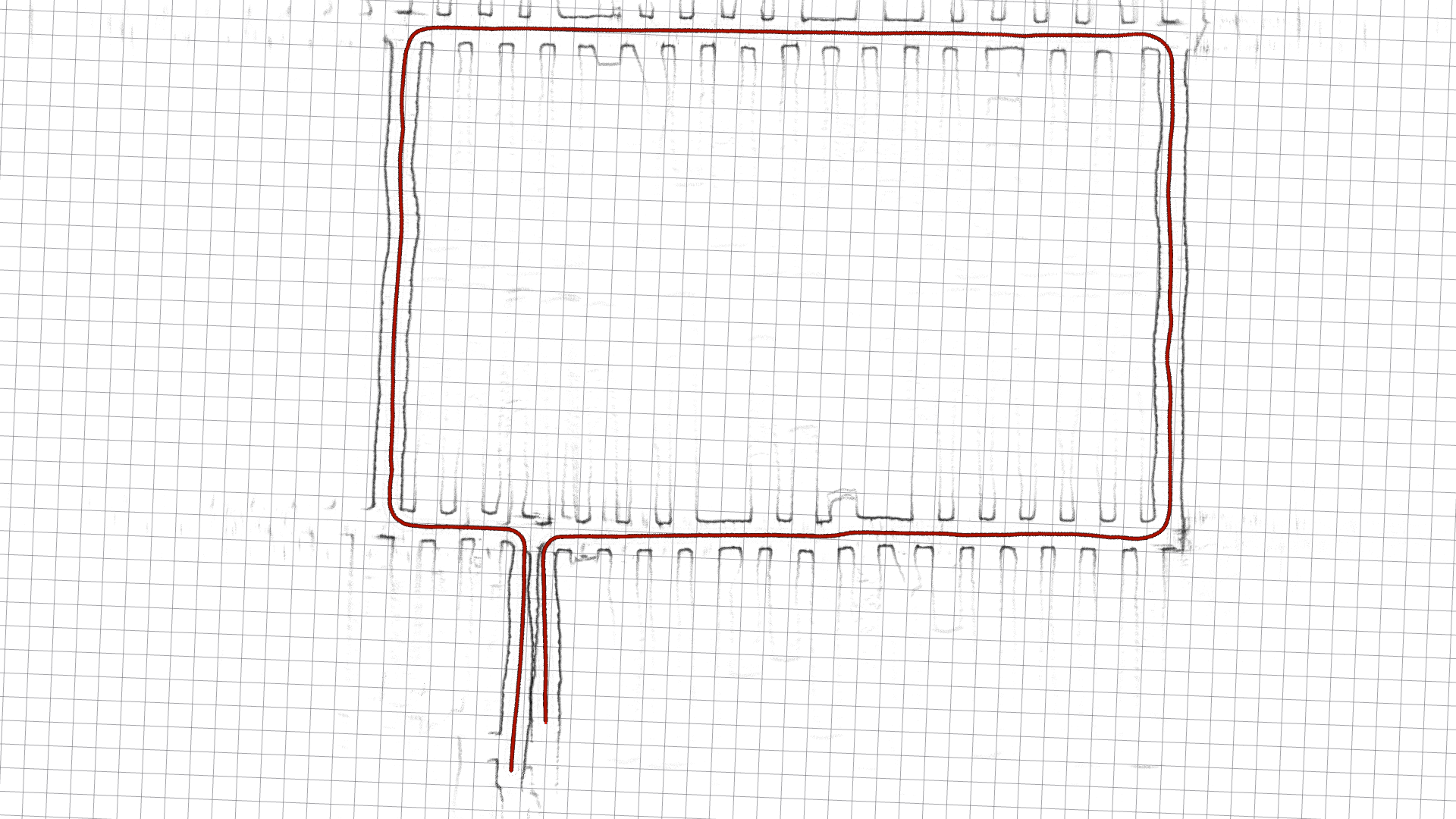}\label{fig:kvarntorp_cfear2}}\\
    \subfloat[CFEAR-3]{\includegraphics[trim={15.0cm 1cm 11.5cm 0cm},clip,width=0.5\hsize]{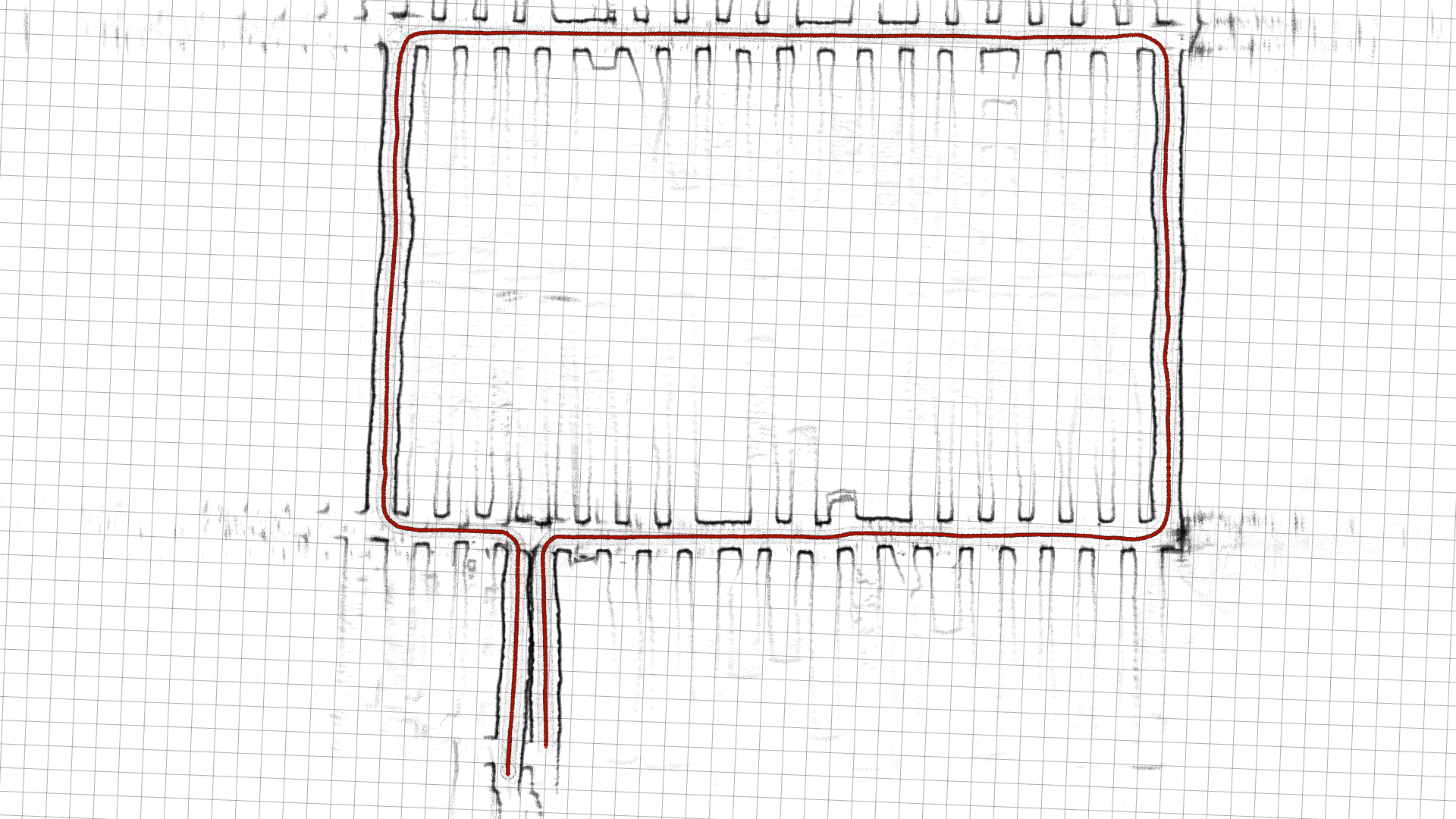}\label{fig:kvarntorp_cfear3}}
     \subfloat[\textit{Baseline odometry} ]{\includegraphics[trim={12.0cm 1cm 10cm 20cm},clip,width=0.5\hsize]{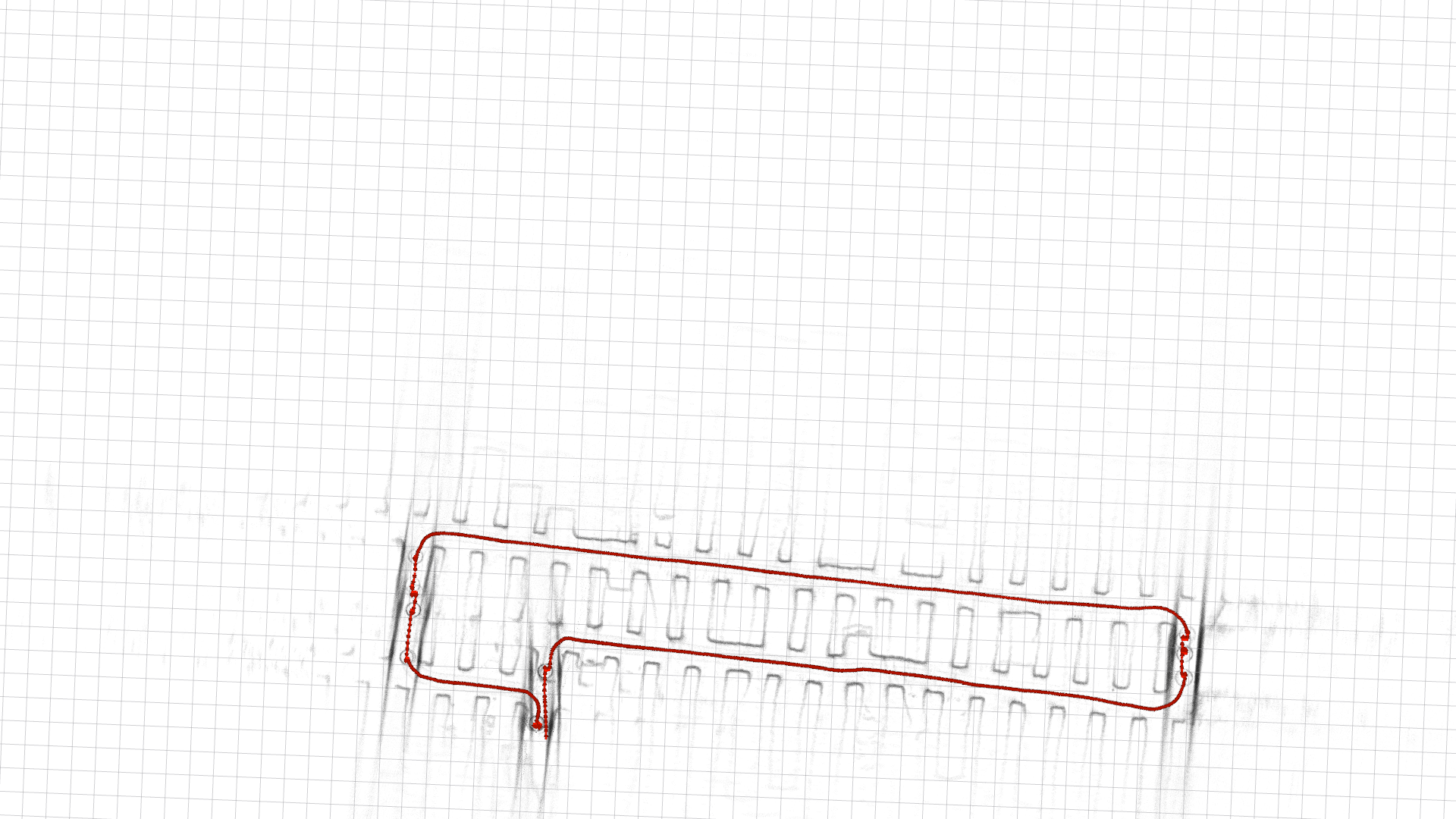}\label{fig:kvarntorp_vanilla}}\\
  \end{center}
  \caption{Estimated odometry using CFEAR-1 (with and without weights), CFEAR-(2/3) and \textit{Baseline odometry} in the Kvarntorp dataset. \textit{Baseline odometry} fails in this scenario and estimates hardly any motion along the long vertical corridors where features are scarce.}\label{fig:kvarntorp_qualitative}
  \vspace{-0.3cm}
\end{figure}

In this dataset, a car with a roof-mounted radar was driven $1235$~m at a speed of $10$km/h through a loop in an underground mine. The walls in the vertical passages in Fig.~\ref{fig:kvarntorp_map} contain few subtle features and higher odometry uncertainty is expected. As seen in Fig.~\ref{fig:kvarntorp_vanilla}, ~\textit{Baseline odometry}, which uses P2P matching with a single keyframe, fails in these cases. In contrast, we found that all configurations of CFEAR achieved smooth and reliable odometry without any failures, trajectories are presented in Fig.~\ref{fig:kvarntorp_qualitative}(a,b,c). 
\changed{The following final position errors was measured: CFEAR-1 ($7$~m, $0.5~\%$) followed by CFEAR-3 ($13$~m,  $1.0~\%$) and CFEAR-2 ($14$~m, $1.1~\%$). A clear improvement can be attributed to the weighting scheme and the residual similarity weights in particular. As found in past experiments, the effect is highest for CFEAR-1, for which we plot the difference Fig.~\ref{fig:kvarntorp_cfear1}. Without weights, the final errors increase by: $186\%$ for CFEAR-1 (up to $20$~m, $1.6~\%$), by $15\%$ for CFEAR-3 (up to $15$~m, $1.2~\%$), and by $64\%$ for CFEAR-2 (up to $23$~m, $1.8~\%$).  
}
In these experiments, we see that the environment can be accurately modeled via the surface points and that one-to-multiple correspondences, local surface geometries, and correspondence similarity are key elements to overcoming uncertainty within feature-poor environments.
\\
\subsubsection{Orkla -- Indoor intra-logistics}
\label{sec:orkla_sec}
\begin{figure}[h!]
  \begin{center}
    \subfloat[][Top down view of Orkla facilities mapped with lidar odometry.]{\includegraphics[trim={0.0cm 0cm 0cm 0cm},clip,height=0.56\hsize]{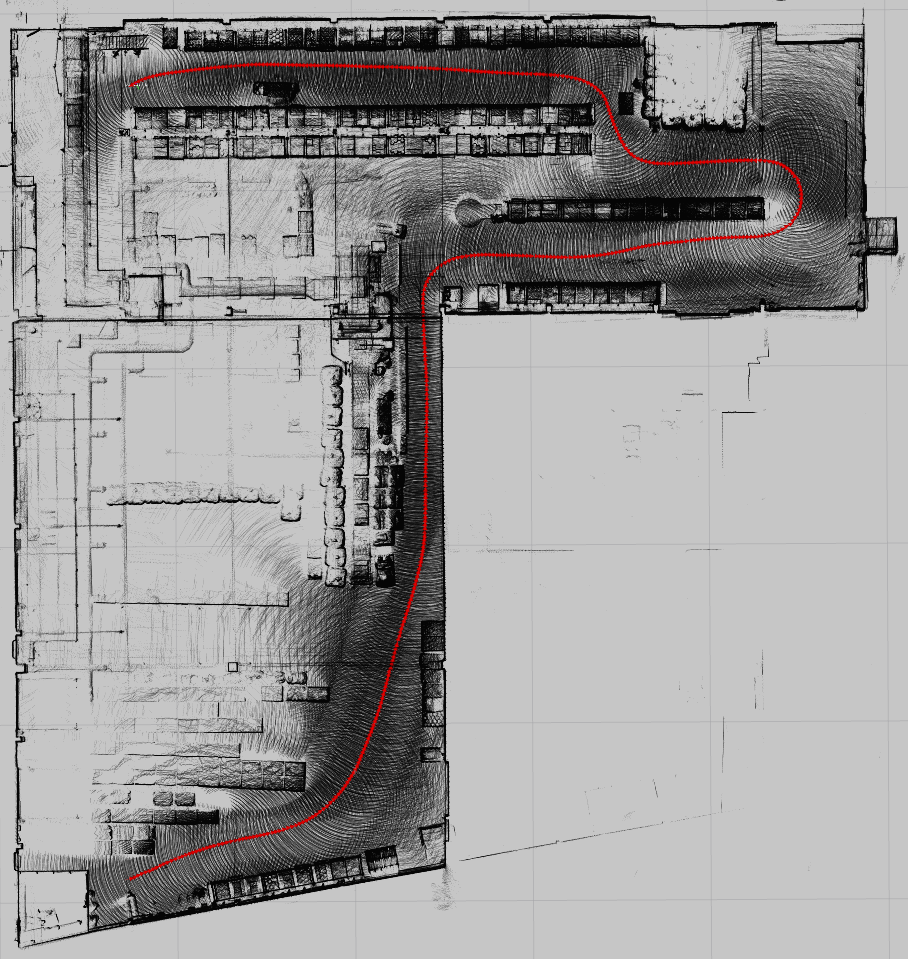}\label{fig:orkla_lidar_odom}}\hspace{0.05cm}
     \subfloat[][Fork lift equipped with top mounted radar.]{\includegraphics[trim={.0cm 0cm 0cm 0cm},clip,height=0.56\hsize]{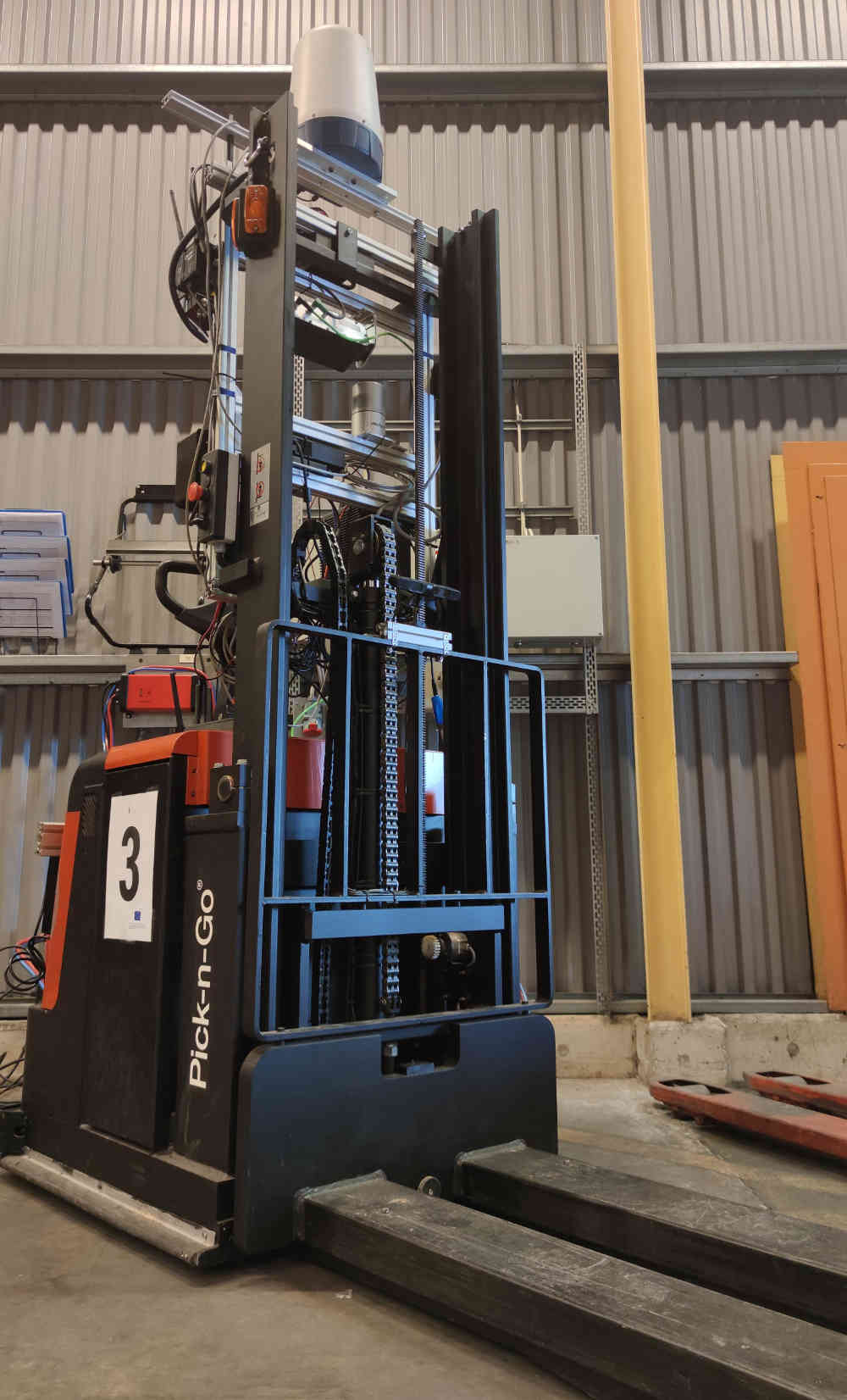}\label{fig:orkla_forklift}}
    \\
    \subfloat[][Overview of Orkla facilities.]{\includegraphics[trim={0.0cm 5cm 0cm 0cm},clip,width=0.9\hsize]{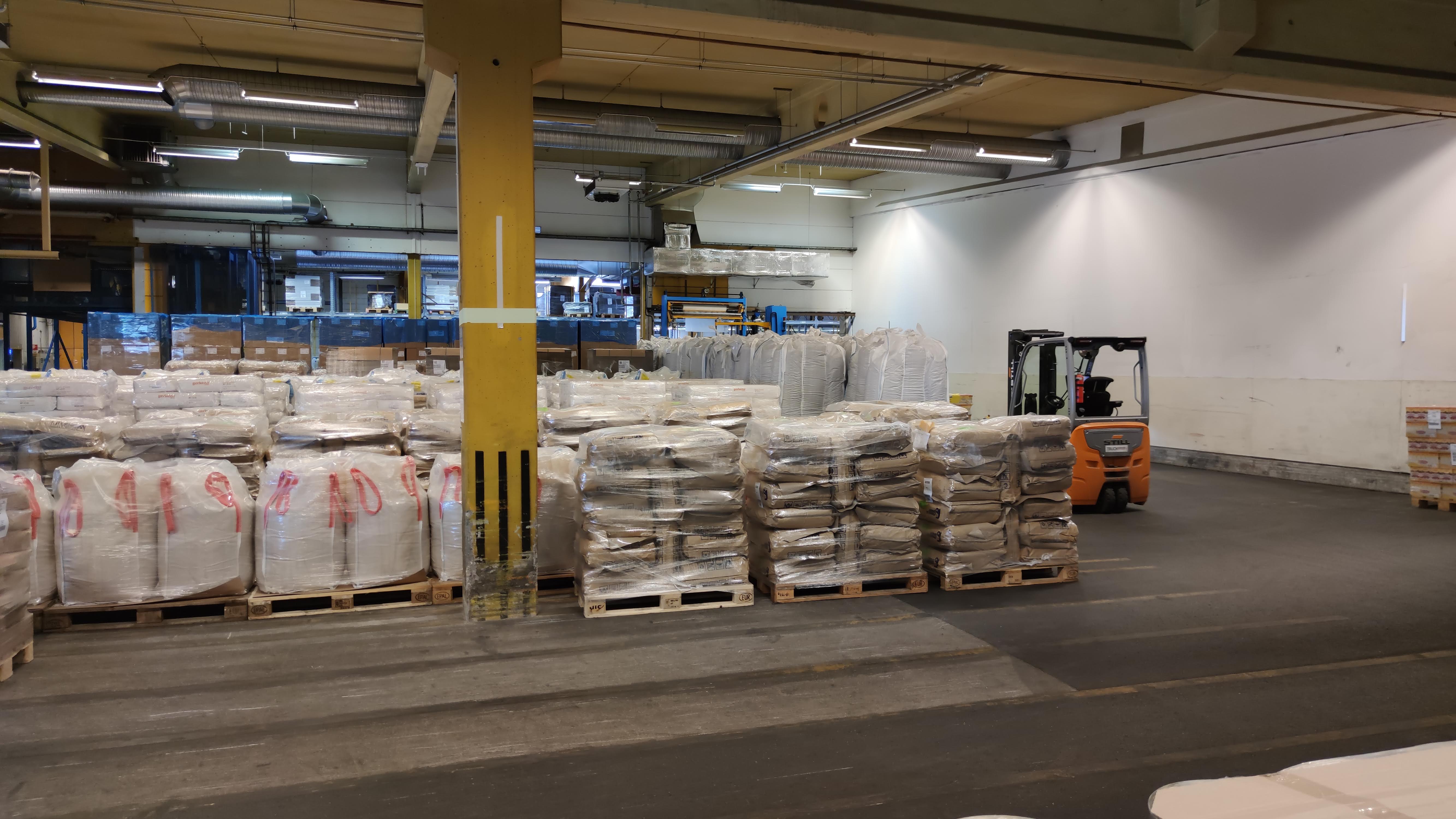}\label{fig:orkla_photo}}
  \end{center}
  \caption{Orkla intra-logistics dataset. The forklift was driven $216$~m at a speed of $2$~km/h within an environment containing pallets, shelves and assembly lines. The size of the environment is roughly $60\times 60$~m. }\label{fig:orkla_overview}
  \vspace{-0.4cm}
\end{figure}

\begin{figure}[h!]
  \begin{center}
       \subfloat[][baseline odometry with $r=3$~m.]{\frame{\includegraphics[trim={15.0cm 2cm 15cm 0cm},clip,width=0.48\hsize]{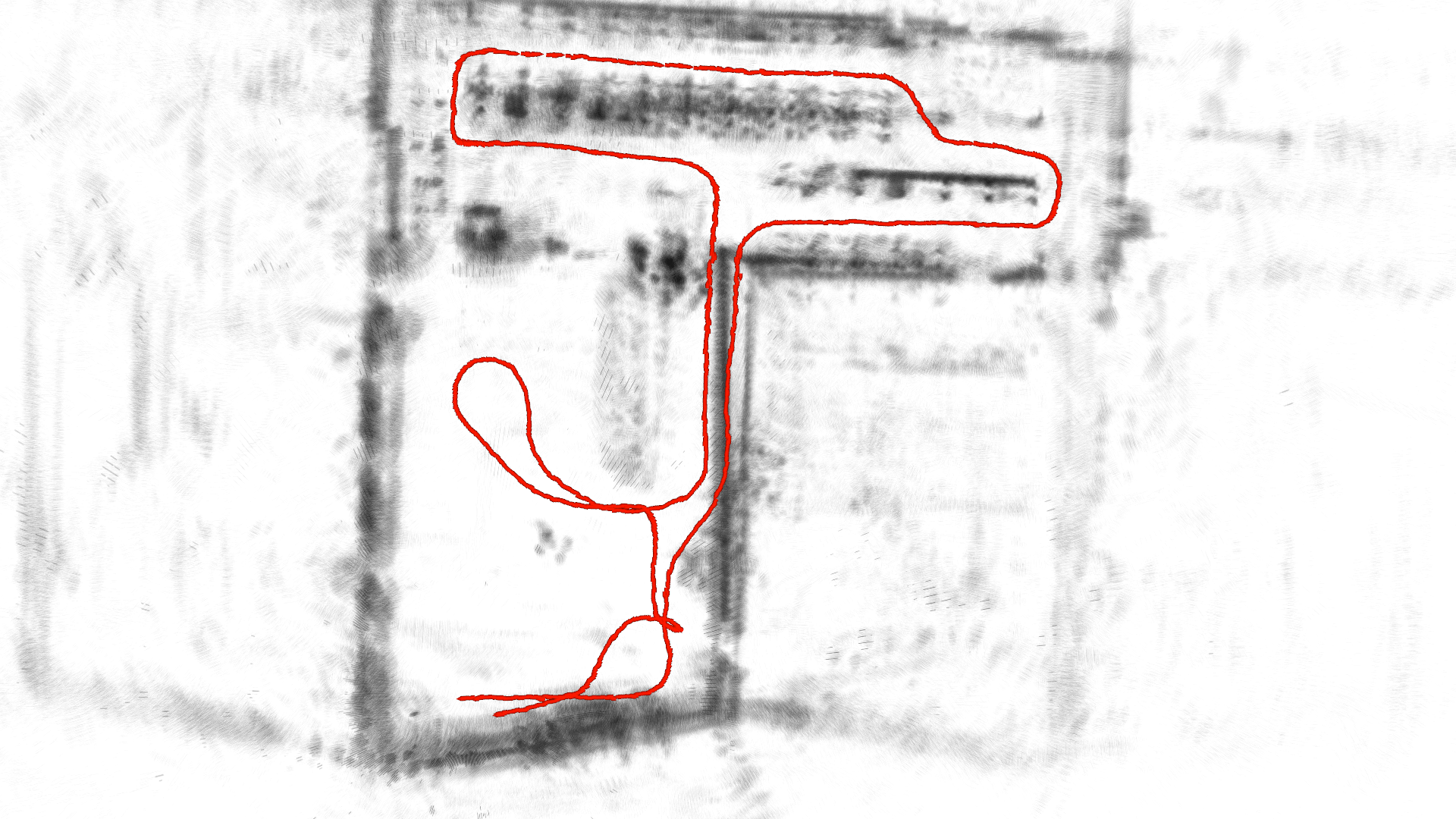}\label{fig:orkla_baseline_r3}}}\hfill
       \subfloat[baseline odometry with $r=1$~m.]{\frame{\includegraphics[trim={15.0cm 2cm 15cm 0cm},clip,width=0.48\hsize]{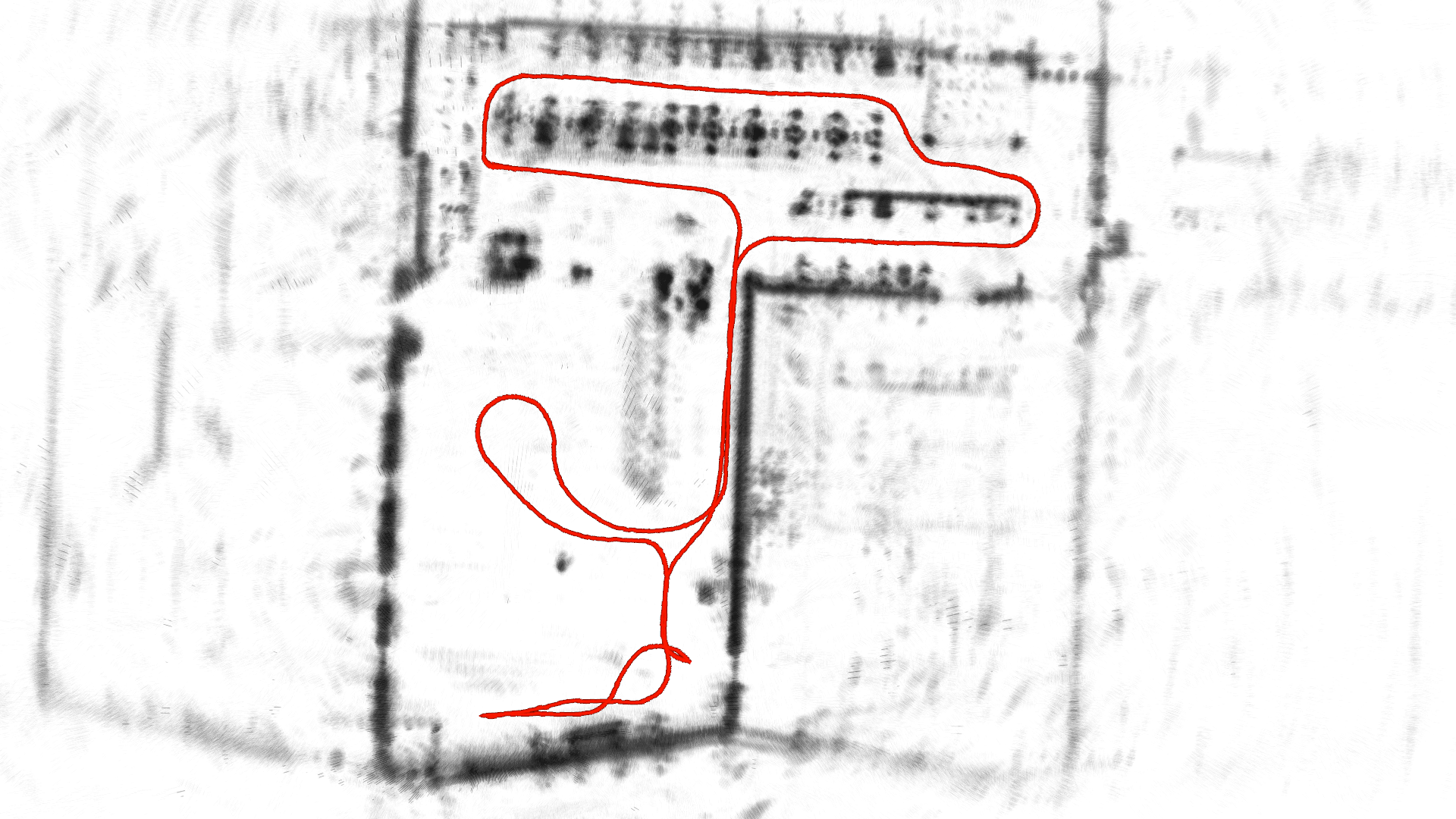}\label{fig:orkla_baseline_r1}}}
  \end{center}
  \caption{Odometry  estimated with \textit{baseline odometry} using CFEAR-features with unchanged settings (a) and with resolution tuned to the environment (b). The baseline method is sensitive to parameter tuning, a too large resolution $r=3$~m produces noisy pose estimates and high drift.}\label{fig:orkla_baseline}
\vspace{-0.4cm}
\end{figure}

In the final dataset, a fork-lift with a top-mounted radar was driven $216$~m through the indoor intra-logistic environment depicted in Fig.~\ref{fig:orkla_overview} at an average speed of $2$~km/h.
A large number of metal shelves and walls are highly radar-reflective and give rise to strong multi-path reflections, 
\changedd{especially when nearby walls are observed with a high incident angle.}
The reflections can be seen e.g. in Fig.~\ref{fig:orkla_baseline_r1} where a part of the map of the environment appears mirrored around the vertical wall in the figure center.

\begin{figure}[h!]
  \begin{center}
    \subfloat[With resolution $r=3$~m, the trajectory has low drift with higher noise in pose estimates.]{\frame{\includegraphics[trim={20.0cm 0cm 13cm 0cm},clip,width=0.48\hsize]{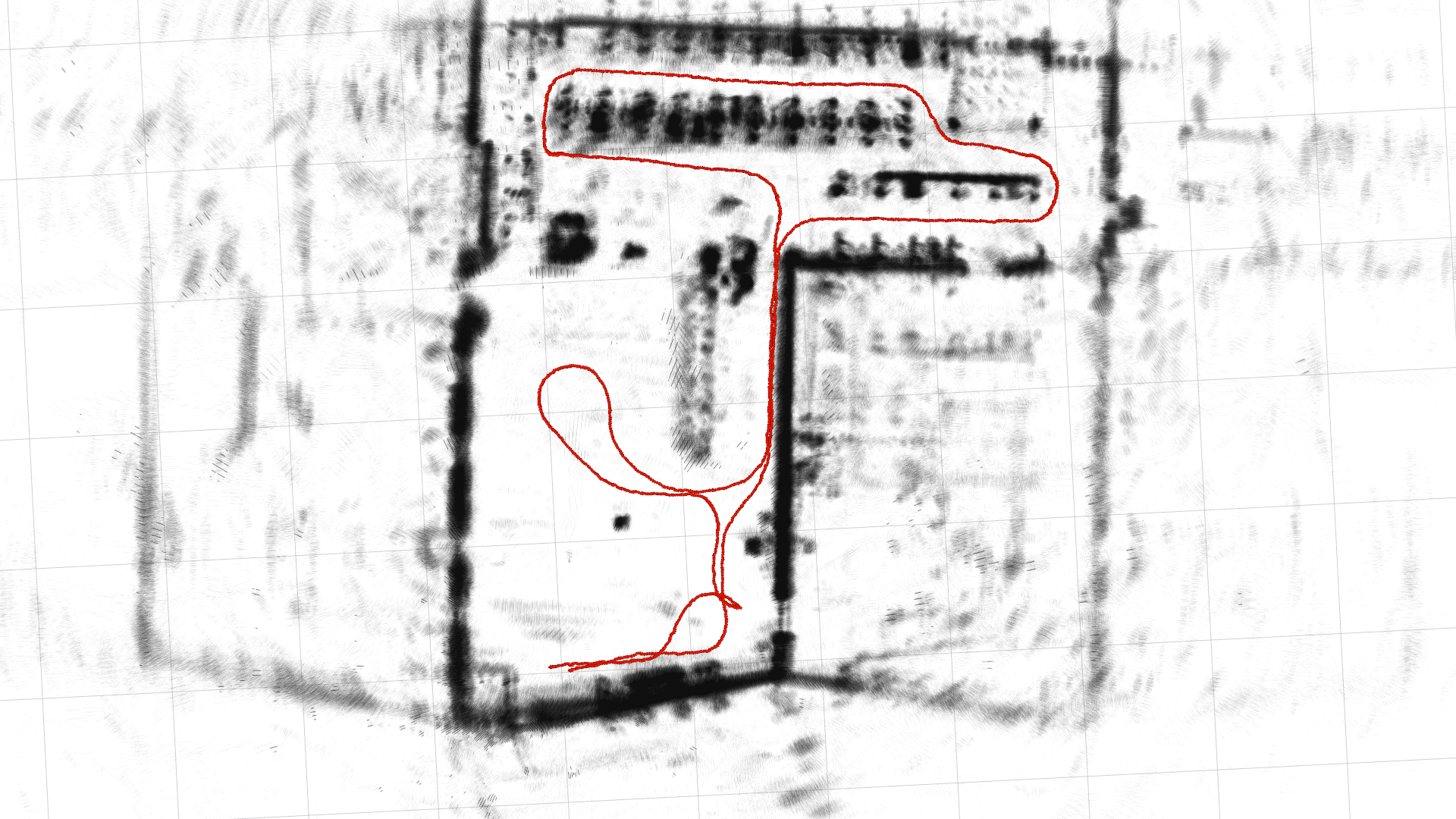}}\label{fig:orkla_cfear3r3_full}}\hfill
    \subfloat[With resolution $r=1$~m (adapted for environment scale), the  trajectory is smoother.]{\frame{\includegraphics[trim={20.0cm 0cm 13cm 0cm},clip,width=0.48\hsize]{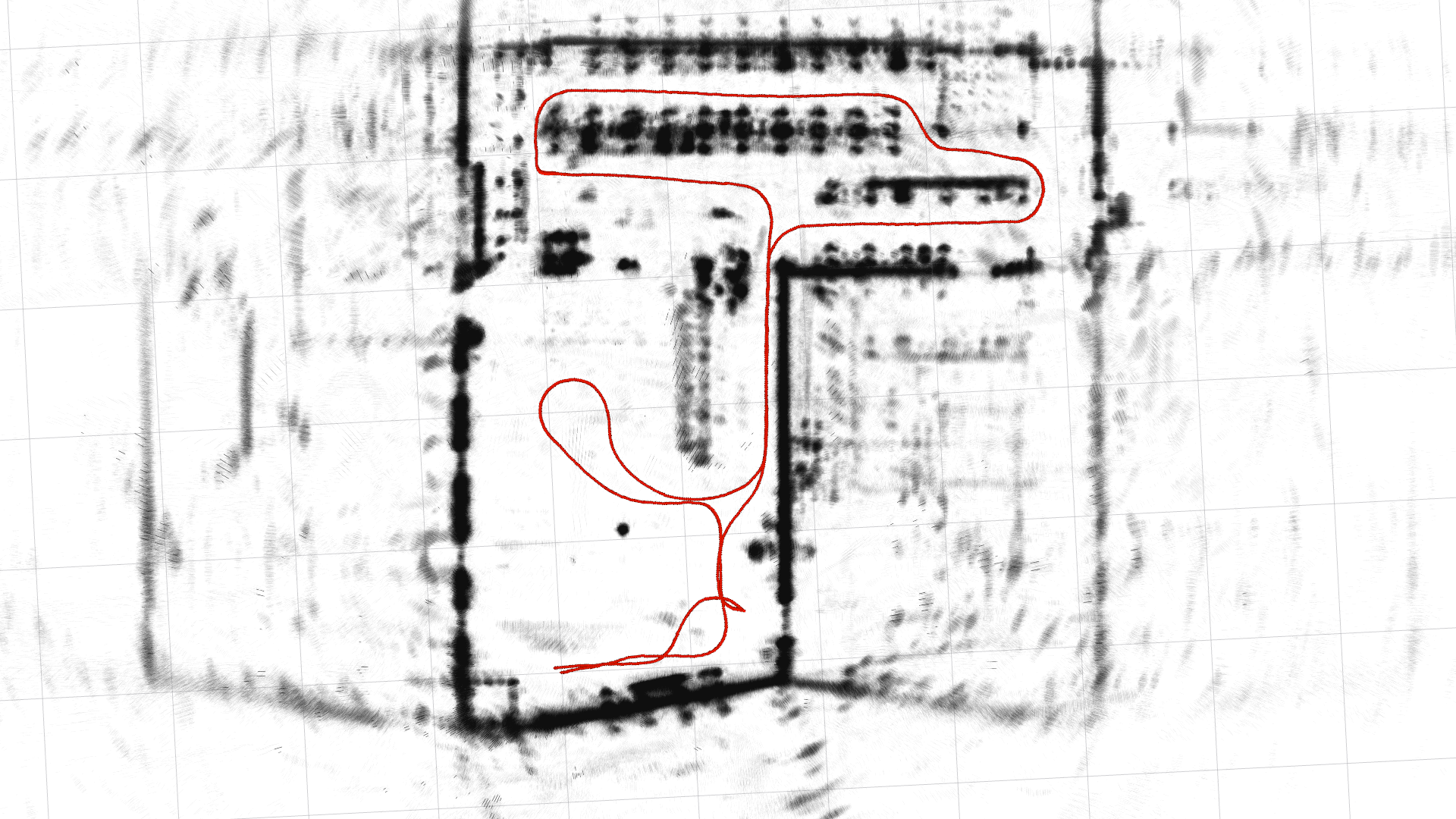}}\label{fig:orkla_cfear3r1_full}}

  \end{center}
  \caption{Estimated odometry in the Orkla dataset using CFEAR-3 with (a) unchanged settings and (b) grid resolution adjusted to environment scale  $r=1$~m. A smaller resolution yields a slightly smoother trajectory and higher map quality. However, the odometry is robust in both cases and maintains low drift over the full trajectory. \changedd{The resilience to low feature quality is achieved with one-to-multiple scan registration.}
  }\label{fig:orkla_qualitative}
  \vspace{-0.3cm}
\end{figure}

The scale of the environment is significantly smaller compared to the previous datasets and important landmarks such as walls, shelves and pallets are generally located close to the radar. Hence, the resolution $r$ of the oriented surface points should ideally be smaller compared to the default resolution.
We compare how \textit{Baseline odometry} and CFEAR-3 are affected by a resolution that is ill-tuned with respect to spatial scale.
The estimated trajectory of \textit{baseline odometry} is visualized with default resolution ($r=3$~m) in Fig.~\ref{fig:orkla_baseline_r3}, and with adapted resolution ($r=1$~m) in Fig.~\ref{fig:orkla_baseline_r1}. The trajectory with default resolution is noisy (which can be seen by closely inspecting the trajectory) and quickly accumulates drift (which can be observed as map blur).
As expected, when the resolution is lowered according to the environment scale, both pose noise and drift are greatly reduced.
On the other hand, CFEAR-3 achieves a smooth trajectory with low drift for both resolutions as seen in Fig.~\ref{fig:orkla_qualitative}(a,b), although the map created with the adapted resolution is crisper and the trajectory is slightly smoother (lower pose noise).
Hence, our presented registration approach enables robust odometry in spite of poor feature quality and ill-tuned resolution.



\section{Conclusions}
In this paper, we presented CFEAR Radar odometry, a method for efficient and accurate spinning radar odometry.

Our pipeline was quantitatively evaluated on two public benchmarks, including the Oxford Radar RobotCar Dataset~\cite{RadarRobotCarDatasetICRA2020,RobotCarDatasetIJRR} and MulRan~\cite{gskim-2020-mulran}.
We achieve an overall translation error of $1.09\%$ in the Oxford dataset, thus improving our previous state-of-the-art in radar \emph{odometry} with $38\%$ lower drift. The performance was confirmed by experiments in MulRan, for which we present state-of-the-art results with $1.50\%$ translation error.
Surprisingly, without any refinement or loop closure, 
our online incremental method outperforms the currently best radar-based method for \emph{SLAM} with a drift that is lower by \changedd{$40.4\%$ in Oxford and $26.7\%$ in MulRan.} 

We proposed four predefined parameter configurations that can be chosen depending on efficiency preference. In Oxford, our three efficient settings achieve between $1.31\%$ and \changedd{$1.79\%$} translation error, running between $44$ and $160$~Hz on a desktop CPU. Our most accurate setting achieves $1.09\%$ error at $5$~Hz. 

In order to evaluate how our proposed radar odometry method CFEAR achieves accuracy and robustness, we carried out an extensive ablation study from which we derive important insights on odometry estimation. The study found that numerous aspects, including filtering, intensity weighted computation of surface points, motion compensation, residual weighting, and robust loss functions had a high impact on drift and pose accuracy. 
However, the key to the substantial improvement of the state-of-the-art is the combination of point-to-point matching with multiple keyframes. This allows us to overcome sparsity, reduce bias and noise in pose estimates and improve robustness to low-quality features and scene changes, and reduce parameter sensitivity.  

By quantitative and qualitative evaluation, we found our method to robustly generalize across  sensor resolution and different environments scales and types: from outdoor woodland and urban driving in traffic, to indoor warehouses, and underground mines, without changing a single parameter.

We believe that this paper is a significant step in the community's voyage to develop methods for efficient, accurate, robust and environment-independent localization methods based on radar as a highly resilient sensing modality. We believe so for two reasons: because of the consequential insights from our ablation study about what is important to do radar localization ``right'' and because we are able to present an odometry pipeline that outperforms the current state-of-the-art in radar SLAM and starts to challenge lidar-based localization.

\bibliographystyle{IEEEtran}
\bibliography{references}
\iftrue

\begin{IEEEbiography}
    [{\includegraphics[width=1in,height=1.25in,clip,keepaspectratio]{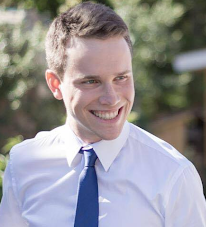}}]{Daniel Adolfsson} 
is a PhD student at the mobile robotics and olfaction (MRO) lab at Örebro University, active within the field of robotic perception. He received his B.S./M.S. degree at Malardalens University, Sweden in 2016 with a specialization in robotics. His previous work targets the creation of high-quality geometric maps and large-scale global localization. His current research interest targets introspective perception and robust methods for localization using spinning lidar and radar.
\end{IEEEbiography}
\vspace{-1cm}
\begin{IEEEbiography}
    [{\includegraphics[width=1in,height=1.25in,clip,keepaspectratio]{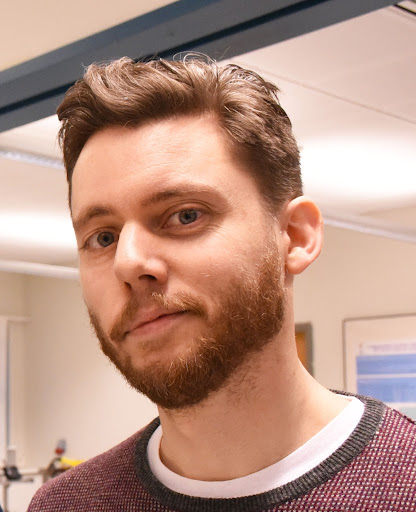}}]{Martin Magnusson} is an associate professor (docent) in computer science at the MRO lab, Örebro University. He received his MSc degree in computer science from Uppsala University in 2004 and PhD degree from Örebro University in 2009. His research interests include 3D perception (including efficient and versatile 3D surface representations), creation and usage of flow-aware and reliability-aware robot maps that go beyond mere geometry, and methods for making use of heterogeneous maps with high uncertainty. 
\end{IEEEbiography}
\vspace{-1cm}
\begin{IEEEbiography}
    [{\includegraphics[width=1in,height=1.25in,clip,keepaspectratio]{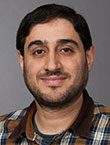}}]{Anas Alhashimi} is a post-doctoral researcher at MRO at Örebro University. He finished his PhD at the department of computer science, electrical and space engineering, Luleå university of technology (2018). He completed his Bachelors's and MSc degrees in electronic and communications engineering (1999, 2002). Anas research interests include: FMCW radar and Laser range finder (Lidar) for robotic applications, Statistical sensor modeling and calibration, Parameter estimation and model order selection under variable noise variance (heteroscedasticity), Joint parameters and state estimation (linear and non-linear systems), Monte Carlo methods, Robot Localization and Mapping, Robot Navigation and path-planning
\end{IEEEbiography}
\vspace{-1cm}
\begin{IEEEbiography}
    [{\includegraphics[width=1in,height=1.25in,clip,keepaspectratio]{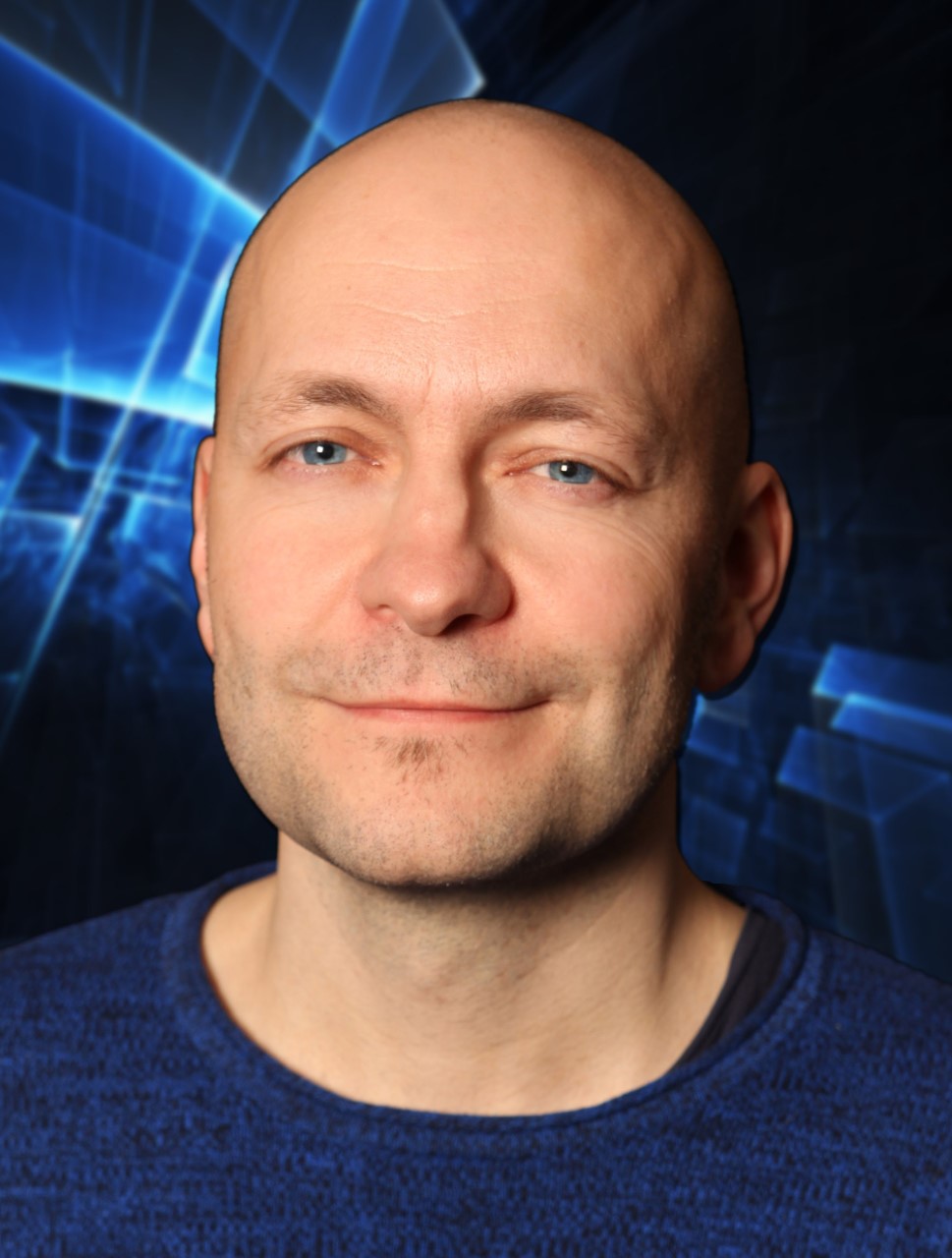}}]{Achim J. Lilienthal,}
is professor of Computer Science and head of the MRO lab at Örebro University. His core research interest is in perception for intelligent systems. Typically based on approaches that leverage domain knowledge and AI, his research work addresses mobile robot olfaction, rich 3D perception, navigation of autonomous transport robots, human-robot interaction and mathematics education research. Achim J. Lilienthal obtained his PhD in computer science from Tübingen University. He has published more than 250 refereed conference papers and journal articles, is senior member of IEEE and evaluator for several national funding agencies and the EU.
\end{IEEEbiography}
\vspace{-1cm}
\begin{IEEEbiography}
    [{\includegraphics[width=1in,height=1.25in,clip,keepaspectratio]{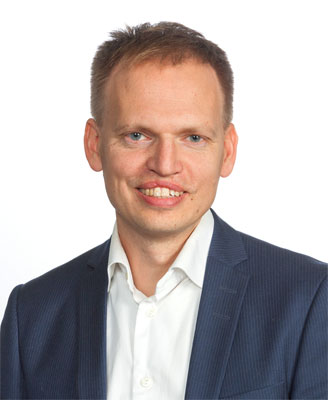}}]{Henrik Andreasson} 's research interest lies mainly in perception aspects ranging from people detection in harsh industrial environments to mapping and localization with a focus on industrial vehicles. He also has an interest in navigation and fleet coordination, more specifically in combining motion planning and control for single and multiple vehicles. He completed his undergraduate studies at the Royal Institute of Technology (KTH) and obtained his PhD from Örebro University.
\end{IEEEbiography}
 \fi

\end{document}